\documentclass[runningheads]{llncs}
\usepackage[T1]{fontenc}
\usepackage{graphicx}
\usepackage{booktabs}
\usepackage{amsmath}
\usepackage{amssymb}
\usepackage{amsfonts}
\usepackage{algorithm}
\usepackage{algorithmic}
\usepackage{multirow}
\usepackage{subcaption}
\usepackage{xcolor}
\usepackage{array}

\usepackage{url}
\usepackage[hidelinks]{hyperref}
\usepackage{caption}

\begin{document}

\title{QSplitFL: Capability Aware Deep Q-Learning for Optimal Split Point Selection in Split Federated Learning}

\titlerunning{QSplitFL: Deep Q-Learning for Split Point Selection in SFL}

\author{Nazmus Shakib Shadin\inst{1}\orcidID{0000-0002-0999-924X} \and
Xinyue Zhang\inst{1}\orcidID{0000-0002-4243-083X} \and
Jingyi Wang\inst{2}\orcidID{0000-0002-4988-5626} \and
Miao Pan\inst{3}\orcidID{0000-0003-2138-4413}}

\institute{Department of Computer Science, Kennesaw State University, Marietta, GA, 30060 USA\\
\email{nshadin@students.kennesaw.edu, xzhang48@kennesaw.edu} \and
Department of Computer Science, San Francisco State University, San Francisco, CA, 94132 USA\\
\email{jingyiwang@sfsu.edu} \and
Department of Electrical and Computer Engineering, University of Houston, Houston, TX, 77204 USA\\
\email{mpan2@uh.edu}}

\maketitle

\begin{abstract}
Federated Learning (FL) combined with Split Learning (SL) is a privacy preserving paradigm that enables training deep neural networks (DNNs) on resource constrained devices while reducing overall training cost. However, determining the optimal split point, meaning the layer where the model is divided still remains a critical challenge, especially when clients have heterogeneous hardware capabilities. Fixed split points can overload weak devices and increase the communication and server load, which slows convergence and reduces stability. This paper introduces QSplitFL, a novel capability-aware Deep Q-Network (DQN) framework for optimal split point selection in Split learning based Federated Learning (SFL) environments. Unlike existing approaches that rely on high-dimensional model weight representations, QSplitFL employs a lightweight state representation derived directly from client hardware metrics, including CPU utilization, memory, battery level, and network latency. The proposed framework incorporates a decayed loss-drop reward function that prioritizes early convergence, and a committee-based DQN architecture with majority voting to mitigate reward hacking. Extensive experiments on MNIST, Fashion-MNIST, CIFAR-10, and CIFAR-100 datasets using CNN, ResNet50, MobileNetV4, and ConvNeXt architectures demonstrate that our approach achieves better convergence and higher accuracy compared to existing methods, while effectively adapting to heterogeneous device resources. The source code is publicly available at \url{https://github.com/AIPO-Lab/QSplitFL}.
\end{abstract}

\section{Introduction}
The rapid adoption of edge devices along with growing concerns about data privacy and security have driven interest in Federated Learning (FL), which enables collaborative model training without centralizing sensitive raw data~\cite{li2020federated}. FL is particularly relevant in Internet of Things (IoT), Internet of Medical Things (IoMT), and edge computing environments, where data are naturally distributed across smartphones, IoT sensors, and medical devices deployed in remote and resource limited settings~\cite{khan2021federated,abreha2022federated}. Despite its benefits, FL relies on client side computation and often assumes that participating clients can train complete neural networks locally, which can fail in real-world scenarios with resource constrained client devices~\cite{thapa2022splitfed}. Considering a practical healthcare scenario where rural hospitals and community health centers aim to collaboratively develop and train deep learning models for a particular disease detection using patient data distributed across multiple facilities~\cite{khan2021federated}. While collaboration is essential for achieving robust models that generalize across diverse patient populations, these facilities often operate with limited computational infrastructure, including legacy hardware, constrained power supplies, and unreliable network connectivity~\cite{beutel2020flower}. For such environments, the computational cost associated with training deep neural networks (DNNs) locally becomes the dominant bottleneck, which limits the applicability of conventional FL.

To address this limitation, Split Learning (SL) offers a compelling alternative by partitioning the model between clients and the server, which enables resource constrained devices to execute only the initial layers of the model~\cite{gupta2018distributed}. The resulting intermediate activations, termed smashed data, are then transmitted to the server, which completes the remaining forward and backward passes. SL reduces client side computation while keeping raw data on the device. Building on this idea, Split learning based federated learning (SFL) integrates the computational efficiency of SL with the privacy guarantees and scalability of FL~\cite{thapa2022splitfed}.

Even with these advantages of SFL, it introduces a critical optimization challenge, which is determining the optimal split point, that means the specific layer where the neural network is divided between clients and the server~\cite{liang2025communication}. The split point determines how computation and communication are balanced between clients and the server~\cite{thapa2022splitfed}. When the split is shallow, clients execute only a small portion of the network, which reduces client side compute but increases the size of intermediate activations sent to the server and increases server side processing. On the other hand, when the split is deep, more layers run on the client, which reduces activation transmission but can overload resource constrained devices~\cite{liu2022wireless}. So in practice, the split point can't be fixed because client capability changes over time and varies across devices, as well as capability can fluctuate across federated rounds due to battery drain, competing applications, network congestion, and hardware heterogeneity within client clusters~\cite{chen2023personalized}. In real-world deployments, some devices can support deeper splits, while others cannot~\cite{wu2022fedadapt}. This context motivates adaptive, capability aware split point selection.

Existing split point selection methods mainly use heuristic rules, exhaustive search, or reinforcement learning that relies on high dimensional state representations derived from model weights~\cite{samikwa2022ares}. These approaches often introduce nontrivial overhead, slow adaptation, and limited interpretability. Methods that build states from model weights typically require dimensionality reduction, often via Principal Component Analysis (PCA)~\cite{wang2020optimizing}. This step adds substantial complexity for collecting weights and computing the projection, which is difficult to justify in resource constraint device deployments.

For mitigating these issues, we propose QSplitFL, a capability aware reinforcement learning (RL) based Deep Q Network framework that selects split points dynamically in SFL settings. Here, we have replaced weight based state representations with a lightweight, interpretable state built from client capability metrics. We formulate split selection as a Markov Decision Process~\cite{wang2020optimizing} in which the state summarizes aggregated cluster capability using normalized metrics for CPU availability, memory utilization, battery level, and network latency, together with a heterogeneity indicator derived from capability distribution across clients. The action selects a split layer from a feasible range, for example, from mid network layers up to the last executable layer, which aligns with real deployments where devices and hospitals differ widely in computing capacity. Resource constrained sites in rural areas may only execute a small portion of the model, while larger hospitals can support deeper computation. This heterogeneity makes a single fixed split point ineffective and motivates a mechanism that adapts split depth to the capability of each participating client cluster. The reward follows a decayed loss drop objective, where the reward function assigns higher credit to early improvements to speed up initial convergence.

QSplitFL uses standard DQN stabilization techniques as described by Wang et al.~\cite{wang2020optimizing}. This includes experience replay, where a finite-capacity buffer stores previous transitions to decorrelate gradient updates and allow the model to reuse informative experiences. Stability in temporal difference (TD) learning is further maintained through a target network that is periodically synchronized with the main network. Additionally, the framework employs committee-based action selection to mitigate the problem of reward hacking, which ensures more robust and reliable decision making for action selection~\cite{wu2024towards}. The principal contributions of this paper are summarized as follows:
\begin{itemize}
\item \textbf{Capability-Aware State}: A lightweight and interpretable state representation based on normalized client capability metrics (CPU, memory, battery, network) and cluster heterogeneity.
\item \textbf{Decayed Loss Drop Reward}: An exponentially decayed loss drop reward that prioritizes early round improvements to accelerate split point discovery. 
\item \textbf{Committee-Based DQN}: A committee-based DQN with multiple CNN models as a single committee member to vote on action selection to improve robustness and reduce reward hacking.
\item \textbf{First DQN-Based Adaptive Split Point Selection}: To the best 
of our knowledge, QSplitFL is the first work to apply Deep Q-Learning 
for optimal split point selection in SFL.
\item \textbf{Comprehensive Evaluation}: Extensive experiments on MNIST, Fashion MNIST, CIFAR-10, and CIFAR-100 datasets using CNN, ResNet50, MobileNetV4, and ConvNeXt across 5, 10, 100, and 200 clients, demonstrate scalability and effectiveness of our method. 
\end{itemize}

\section{Related Work}
\label{sec:related_work}
SFL combines the computation reduction of SL with the collaborative scalability of FL~\cite{thapa2022splitfed,hukkeri2025comprehensive}. In SFL, clients perform SL with a server, while client side updates are periodically aggregated in a federated manner. This hybrid design enables local participation with resource-constrained devices, while allowing parallel processing of multiple clients on the server side. Recent SFL research has emphasized optimization and resource management. Fan et al.~\cite{fan2025madrl} proposed a multi-agent deep reinforcement learning framework for cloud edge device collaborative SFL that jointly optimizes partitioning, resource allocation, and client scheduling. Yu et al.~\cite{yu2025model} introduced U-shaped SFL for vehicular environments and used deep reinforcement learning to handle dynamic resource allocation and split selection under mobility constraints. ESFL formulates workload and server resource allocation for heterogeneous wireless devices using optimization techniques tailored to system constraints~\cite{zhu2024esfl}. Privacy and integrity aspects have also been studied. For example, differential privacy mechanisms that perturb smashed data can reduce the success of label inference attacks~\cite{wu2023split}. IV-FED uses trusted execution environments to support training integrity in healthcare IoT scenarios~\cite{li2024integrity}.

Reinforcement learning (RL), and deep reinforcement learning in particular, has been applied to decision making problems in distributed machine learning systems~\cite{lee2019survey}. Deep Q-Networks (DQN) are widely used for discrete action spaces common in scheduling, allocation, and partitioning tasks~\cite{wang2020optimizing}. In distributed learning, RL has been used for split point selection, client selection, and task offloading. Goal oriented DNN splitting methods use RL to control split decisions under resource constraints and accuracy objectives~\cite{binucci2024enabling}. Earlier approaches applied Q-learning to split decisions using PCA compressed weight based representations~\cite{samikwa2022ares}. RL has also been used to select clients in heterogeneous IoT FL settings to balance convergence and resource usage~\cite{yan2023node}, and to optimize task offloading between edge and cloud resources~\cite{yuan2022dqn}. Improvements to experience replay, such as prioritized experience replay and diversity aware replay, have further improved sample efficiency in DRL training~\cite{schaul2015prioritized}.


\textbf{Research Gap and Our Contribution:} Existing RL-based approaches represent the state using high-dimensional model weights with PCA compression~\cite{samikwa2022ares}, which introduces substantial overhead and limits the scalability. They also represent dynamic client capability changes, which makes split point selection decisions slow to adapt. In addition, single agent decision making can be vulnerable to reward hacking, where the policy exploits weaknesses in the reward signal rather than improving true training performance~\cite{wu2024towards}. Finally, reward formulations that treat all rounds equally can miss the importance of early round decisions for shaping later convergence behavior. QSplitFL addresses these gaps through a lightweight capability-aware state, a committee-based DQN with majority voting, and a decayed loss-drop reward for adaptive split point selection in resource-constrained SFL environments.

\section{System Model}
\label{sec:system_model}


\begin{figure*}[ht]
\centering
\includegraphics[width=\textwidth]{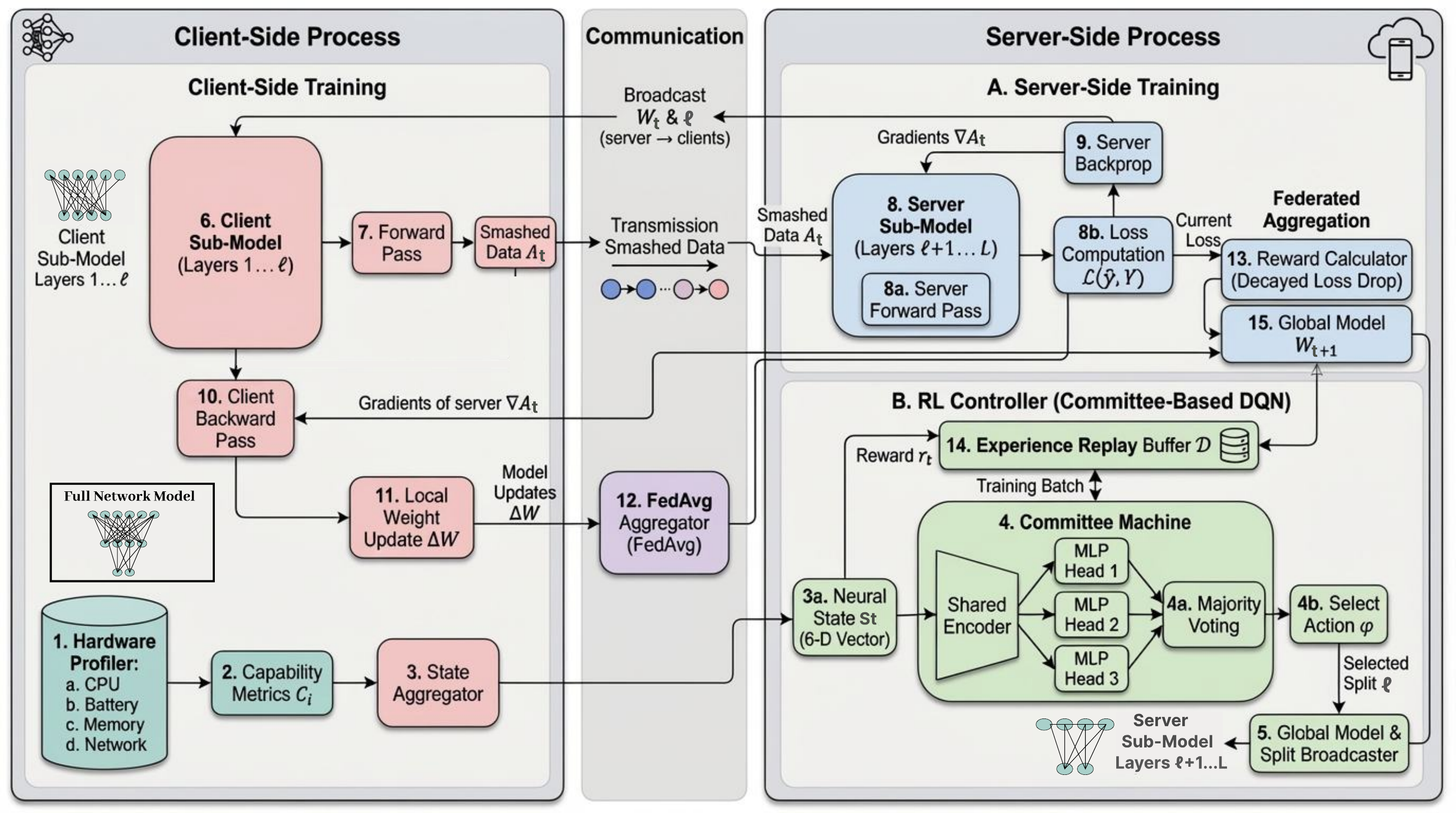}
\caption{\textbf{QSplitFL Workflow Architecture.} (1)~\textit{Client-Side}: Clients has hardware metrics, receive split layer $\ell$, and run forward propagation through layers 1 to $\ell$, which produces smashed data ($A_k$). (2)~\textit{Server-Side}: The server completes training through layers $\ell+1$ to $L$ and returns gradients to clients. (3)~\textit{Aggregation}: Client updates are aggregated via FedAvg; reward ($r_t$) is computed on the based of the loss function. (4)~\textit{RL Controller}: A committee of MLP models votes on the optimal split layer based on client capabilities.}
\label{fig:qsplitfl_architecture}
\end{figure*}

Figure~\ref{fig:qsplitfl_architecture} presents the QSplitFL workflow architecture.
On the \textbf{Client-Side Process}, each client begins with a Hardware Profiler (Step~1) that measures four real-time metrics: CPU utilization, battery level, memory availability, and network latency. These raw readings are converted into normalized capability metrics $C_i$ (Step~2) and then passed to a state aggregator (Step~3), which constructs a six-dimensional neural state vector $s_t$ (Step~3a) summarizing the cluster's overall capability and heterogeneity. After receiving the selected split layer $\ell$ and global model with $L$ layers and weights $W_t$ from the server (Step~5), each client loads its client sub-model with layers $1$ to $\ell$ (Step~6) and runs a forward pass (Step~7) to produce smashed data $A_\ell$, which is transmitted to the server.

On the \textbf{Server-Side Process}, in Block~A, the server receives the smashed data and executes the server forward pass (Step~8a) through the server sub-Model covering layers $\ell+1$ to $L$ (Step~8). The server calculates $\mathcal{L}(\hat{y}, Y)$ by comparing predictions with true labels (Step~8b). Server backpropagation (Step~9) computes gradients $\nabla A_\ell$, which are sent back to clients. Each client then performs its client backward pass (Step~10), computes local weight updates $\Delta W$ (Step~11), and sends them to the Aggregator (Step~12) for federated aggregation. The aggregated loss is fed into the reward calculator (Step~13), which computes the decayed loss-drop reward $r_t$, and the updated global model $W_{t+1}$ (Step~15) is prepared for the next round.

In Block~B (\textbf{RL Controller}), the server executes the committee-based DQN. The state vector $s_t$ enters the committee stage (Step~4), which consists of a shared encoder feeding into multiple independent MLP Heads. Each head proposes a split action, and majority voting (Step~4a) determines the consensus. The winning action is selected (Step~4b) as the split layer $\ell$, which is broadcast to all clients through Step~5. The reward $r_t$ from Step~13 is stored in the experience replay buffer $\mathcal{D}$ (Step~14), from which training batches are sampled to update the committee networks.

Consider a federated network with $K$ edge clients partitioned into $C$ clusters, denoted $\{\mathcal{K}_c\}_{c=1}^{C}$, where each cluster $\mathcal{K}_c$ contains clients with similar hardware capabilities. The central server maintains the server-side portion of the neural network and coordinates training across all clusters. Within each cluster, a dedicated RL agent (the Q-controller, Block~B in Figure~\ref{fig:qsplitfl_architecture}) observes cluster-level capability states and selects appropriate split points for each training round.


\textbf{Computation-Communication Trade-off:}
SFL addresses the computational limitations of resource-constrained devices by allowing each client to execute only the initial layers of the neural network model. The intermediate activations (smashed data) along with the corresponding true labels are transmitted to a central server, which completes the more computation-intensive portions of the training including the deeper layers and loss computation. This approach significantly reduces the workload on devices with limited power, memory, or battery, making participation possible even in resource constrained settings like rural healthcare facilities. Although SFL increases communication costs because activations and gradients must be exchanged between clients and the server; this trade-off is necessary and acceptable~\cite{singh2019detailed}. In critical applications like collaborative AI in healthcare, where enabling weak devices to participate matters more than reducing network latency. If a device cannot run the full model locally, the lower communication cost of traditional FL is irrelevant as the device simply cannot participate. QSplitFL optimizes this trade-off by adaptively choosing split points that balance client computation against communication overhead.

\section{QSplitFL Framework}
\label{sec:qsplitfl_framework}

\subsection{Per-Round Operation}
\label{sec:per_round_detail}
\label{sec:framework_overview}
\begin{figure}[!ht]
\centering
\includegraphics[width=\columnwidth]{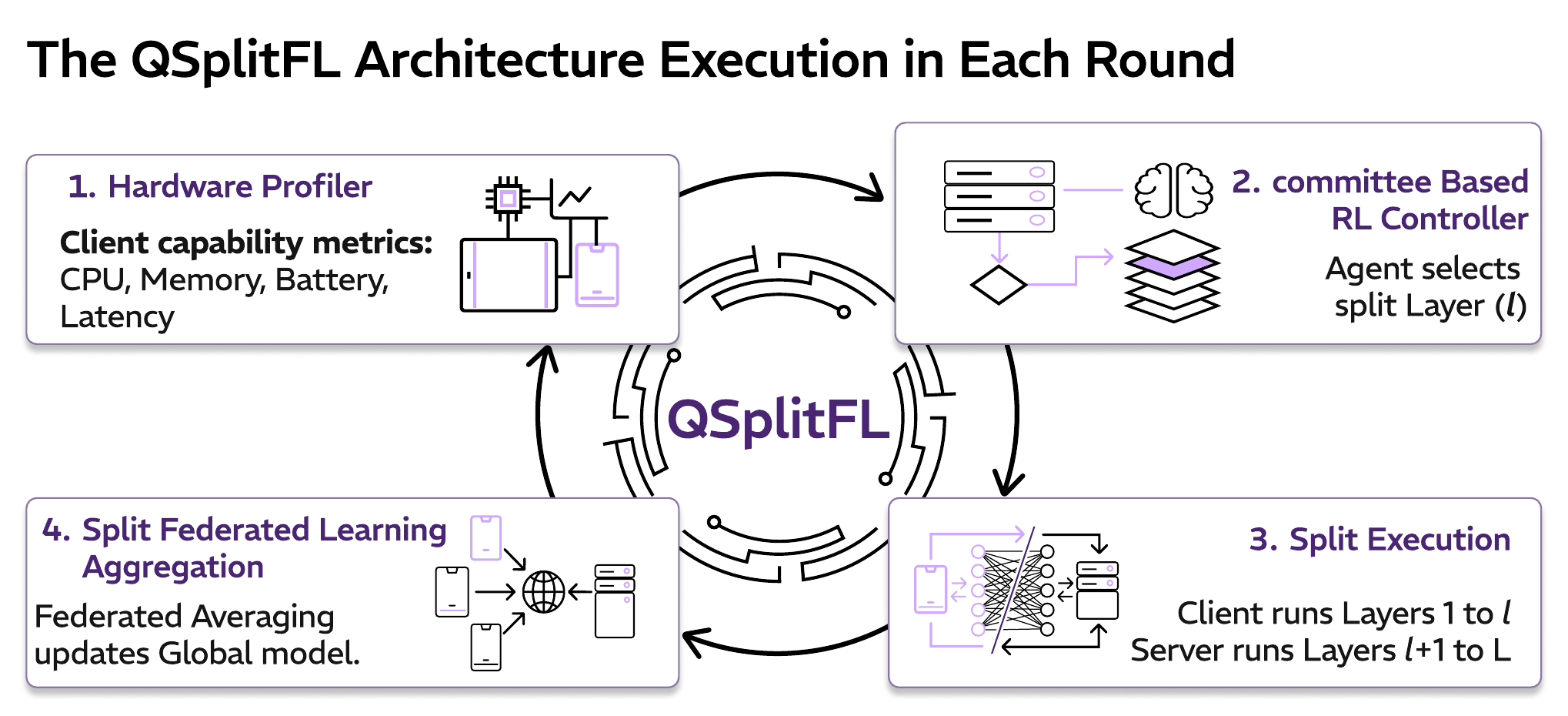}
\caption{\textbf{High-Level Overview of QSplitFL Framework Operation in Each Training Round.} The diagram illustrates the complete workflow: (1) capability metric collection and state construction, (2) committee-based split point selection, (3) SFL execution with smashed data transmission, and (4) federated aggregation with reward computation.}
\label{fig:qsplitfl_round}
\end{figure}

Before formalizing the individual components, we first describe how they interact within a single training round, which gives an end-to-end view of the framework. Figure~\ref{fig:qsplitfl_round} provides a detailed view of how the QSplitFL framework operates during each training round. It shows the integration of capability-aware state construction (corresponding to Steps~1--3 in Figure~\ref{fig:qsplitfl_architecture}), RL-based split point selection (Step~4), SFL execution (Steps~5--11), and federated aggregation with reward computation (Steps~12--15). The per-round workflow proceeds as follows:
\begin{enumerate}
    \item \textbf{State Construction:} At the beginning of each round, the Hardware Profiler (Step~1) collects capability metrics (CPU, memory, battery, network) from all clients, which are normalized into Capability Metrics $C_i$ (Step~2) and aggregated by the State Aggregator (Step~3) into the six-dimensional neural state $s_t$.
    \item \textbf{Action Selection:} The RL Controller's Committee Machine evaluates the current state through a Shared Encoder and multiple MLP Heads. Majority Voting (Step~4a) determines the consensus split layer, and the selected action is output (Step~4b).
    \item \textbf{SFL Execution:} The Global Model and Split Broadcaster (Step~5) sends $W_t$ and $\ell$ to all clients. Each client loads its Sub-Model (Step~6) and runs the Forward Pass (Step~7) to produce smashed data $A_\ell$. The server executes the Server Forward Pass (Step~8a) through layers $\ell+1$ to $L$, computes the loss (Step~8b), and performs Server Backpropagation (Step~9). Gradients $\nabla A_\ell$ are returned to clients for the Client Backward Pass (Step~10) and Local Weight Update (Step~11).
    \item \textbf{Federated Aggregation:} The FedAvg Aggregator (Step~12) combines all client updates. The Reward Calculator (Step~13) computes the decayed loss-drop reward $r_t$, which is stored in the Experience Replay Buffer $\mathcal{D}$ (Step~14). The updated Global Model $W_{t+1}$ (Step~15) is prepared for the next round.
\end{enumerate}

With this per-round view in place, we now formalize each component, beginning with the Markov Decision Process formulation that underpins split point selection.

\subsection{Markov Decision Process Formulation}
We formulate the split point selection problem as a Markov Decision Process (MDP) defined by the tuple $(\mathcal{S}, \mathcal{A}, P, R, \gamma)$, where $\mathcal{S}$ denotes the state space, $\mathcal{A}$ the action space, $P$ the transition, $R$ the reward function, and $\gamma$ the discount factor. The MDP connects the key blocks shown in Figure~\ref{fig:qsplitfl_architecture}: the state is constructed from client capability metrics (Steps~1--3), the action determines the split layer via the RL Controller (Step~4), the transition executes one SFL round (Steps~5--12), and the reward is derived from the aggregated loss (Step~13).

\subsubsection{State Space:} The state at round $t$ for cluster $c$ encodes the aggregated hardware capabilities and heterogeneity of participating clients. As shown in 
Steps~1--3 of Figure~\ref{fig:qsplitfl_architecture}, each client's 
Hardware Profiler (Step~1) reports raw metrics, which are normalized into 
Capability Metrics $C_i$ (Step~2) and aggregated by the State Aggregator 
(Step~3) into a compact neural state vector $s_t$. Unlike prior approaches 
that employ high-dimensional model weight representations requiring PCA 
compression, we define a compact, interpretable state vector based on four 
normalized capability metrics. For each client $k$ at time $t$, we compute: 
CPU availability as $C_{\text{CPU}}^{(k)}(t) = \text{CPU}_{\text{avail}}^{(k)}(t) / \text{CPU}_{\max}$, 
memory availability as $C_{\text{Mem}}^{(k)}(t) = \text{Mem}_{\text{avail}}^{(k)}(t) / \text{Mem}_{\max}$, 
battery level as $C_{\text{Bat}}^{(k)}(t) = \text{Bat}_{\text{level}}^{(k)}(t) / \text{Bat}_{\max}$, 
and network quality as $C_{\text{Net}}^{(k)}(t) = 1 - \text{Latency}^{(k)}(t) / \text{Latency}_{\max}$. All metrics are normalized to the interval $[0, 1]$ using min-max normalization across the federated network; see Appendix~\ref{appendix:metrics} for detailed descriptions. The overall capability score for client $k$ combines these metrics as $C_{\text{Overall}}^{(k)}(t) = \sum_{i=1}^{4} w_i \cdot C_i^{(k)}(t)$, where $w_1, w_2, w_3, w_4 \geq 0$ and $\sum_{i=1}^{4} w_i = 1$ are importance weights which are tunable according to the deployment requirements. The cluster-level state vector aggregates individual client metrics into a six-dimensional representation: $s_t^{(c)} = [\bar{C}_{\text{CPU}}^{(c)}(t), \bar{C}_{\text{Memory}}^{(c)}(t), \bar{C}_{\text{Battery}}^{(c)}(t), \bar{C}_{\text{Network}}^{(c)}(t), \bar{C}_{\text{Overall}}^{(c)}(t), \sigma_c(t)]$. Here, each mean capability is computed as $\bar{C}_i^{(c)}(t) = \frac{1}{|\mathcal{K}_c|} \sum_{k \in \mathcal{K}_c} C_i^{(k)}(t)$, which represents the average of metric $i$ across all clients in cluster $c$. The heterogeneity indicator $\sigma_c(t) = \sqrt{\frac{1}{|\mathcal{K}_c|} \sum_{k \in \mathcal{K}_c} ( C_{\text{Overall}}^{(k)}(t) - \bar{C}_{\text{Overall}}^{(c)}(t) )^2}$ captures the standard deviation of overall capability scores within the cluster. This six-dimensional state representation provides direct interpretability: high $\bar{C}_{\text{Overall}}^{(c)}(t)$ indicates the cluster can accommodate deeper split points, while high $\sigma_c(t)$ shows heterogeneous capabilities which require adaptive split point selection to avoid overloads from weaker client devices.

\subsubsection{Action Space:} The action space consists of feasible split layer indices $\mathcal{A} = \left\{ \ell \in \mathbb{N} : \ell_{\min} \leq \ell \leq \ell_{\max} \right\}$,
where $\ell_{\min} = \lceil L/2 \rceil$ represents the minimum split point set at approximately half the total network depth $L$, and $\ell_{\max} = L - 1$ is the maximum split point. The RL Controller's Committee Machine (Step~4 in Figure~\ref{fig:qsplitfl_architecture}) selects one split layer from this range through majority voting (Step~4a) and outputs the selected action (Step~4b). The lower bound $\ell_{\min} = \lceil L/2 \rceil$ ensures that clients execute at least half of the network layers, which serves two purposes: (1) it prevents clients from sending raw or nearly-raw data to the server, which would increase privacy risks and communication overhead; and (2) it ensures client devices perform meaningful local feature extraction before transmitting smashed data. For example, in ResNet50 with $L=50$ layers, this constraint sets the split point at least in the half layer ($\ell_{\min} = 25$). The upper bound $L-1$ excludes the output layer, which must reside on the server for loss computation and gradient initialization. Each action $a_t = \ell$ specifies that clients compute layers $1$ through $\ell$ while the server computes layers $\ell + 1$ through $L$.

\subsubsection{Transition:} After the RL Controller selects a split layer $\ell$ at round $t$ (Step~4 in Figure~\ref{fig:qsplitfl_architecture}), the environment executes one complete SFL training round following Steps~5 through~12 as shown in Figure~\ref{fig:qsplitfl_architecture}, using Algorithm~\ref{alg:train_round} from Appendix~\ref{sec:algorithms}:
\begin{equation}
W_t = \texttt{TrainRoundSFL}(c, s_t^{(c)}, W_{t-1}; Q_\theta, \epsilon),
\label{eq:transition}
\end{equation}
where the server broadcasts parameters and split configuration (Step~5), clients perform forward passes and transmit smashed data (Steps~6--7), the server completes forward and backward passes (Steps~8--9), clients perform backpropagation and weight updates (Steps~10--11), and updates are aggregated via FedAvg (Step~12). The next state is then computed:
\begin{equation}
s_{t+1}^{(c)} = \Phi\left( \{C_i^{(k)}(t+1)\}_{k \in \mathcal{K}_c} \right),
\label{eq:next_state}
\end{equation}
where $\Phi(\cdot)$ applies the aggregation to updated client capability metrics.

\subsection{Reward Function Design}
The reward function guides the RL agent toward optimized split point selections that minimize cluster loss while considering capability constraints. As indicated in Step~13 of Figure~\ref{fig:qsplitfl_architecture}, the Reward Calculator computes the reward ($r_t$) based on the decayed loss-drop during the aggregation phase. We employ a decayed loss-drop reward that prioritizes early-round improvements.

\subsubsection{Cluster Loss Computation:} At each round $t$, we calculate the overall validation loss for a cluster by averaging the losses from all clients, where we assign more weight to clients with more data. The cluster loss is computed as $L_t = \sum_{k=1}^{K} \omega_k \cdot L_t^{(k)}$, where $\omega_k = \frac{n_k}{\sum_{j=1}^{K} n_j}$ represents the weight of client $k$ based on its number of training samples $n_k$.

\subsubsection{Decay Factor:} Since the model improves most in early rounds, we use an exponential decay factor to give more credit to early round gains and less to the later ones for optimal split point selection. So, the exponential decay factor can be defined as, $\rho_t = e^{-\lambda(t-1)}, \text{where} \quad \lambda > 0$.
Here, at round $t = 1$, we have $\rho_1 = e^0 = 1$, ensuring the first round receives full credit. As training progresses, $\rho_t$ decays exponentially toward zero, causing later rounds to receive diminishing credit for equivalent loss improvements. The decay rate $\lambda$ controls the speed of this transition: larger values of $\lambda$ result in sharper decay, while smaller values produce more gradual decay. In practice, we set $\lambda$ such that $\rho_T \ll 1$ by the final round, which ensures the RL agent prioritizes early stage convergence.

\subsubsection{Loss-Drop Reward:} The reward at round $t$ measures the improvement in cluster loss, weighted by the temporal decay:
$\boxed{r_t = -\left( L_t - L_{t-1} \right) \cdot \rho_t}$.
If the selected split point leads to loss reduction ($L_t < L_{t-1}$), the reward is positive; and if the loss increases it yields negative reward. The decay factor $\rho_t$ ensures that equivalent improvements in later rounds receive progressively smaller rewards, encouraging the agent to prioritize early convergence. Detailed reward behavior examples are provided in Table~\ref{tab:reward_scenarios} in Appendix~\ref{sec:reward_table}.

\subsection{Deep Q-Network Architecture}
The RL Controller shown in Block~B of Figure~\ref{fig:qsplitfl_architecture} is implemented using a Deep Q-Network (DQN) to approximate the action value function $Q(s, a)$. The Q-network is an MLP that maps the six-dimensional capability state (constructed through Steps~1--3) to Q-values for each possible split action. To improve learning stability, we maintain an experience replay buffer $\mathcal{D}$ that stores transition tuples $e_t = (s_t, a_t, r_t, s_{t+1}, d_t)$, where $d_t \in \{0, 1\}$ indicates round termination. The buffer operates under a First-In-First-Out (FIFO) policy with fixed capacity. During training, mini-batches of $B$ transitions are sampled uniformly from $\mathcal{D}$, which breaks the correlation between consecutive samples and helps the model learn more effectively. We also use a target network $Q_{\bar{\theta}}$ with parameters $\bar{\theta}$ that are periodically synchronized with the current network. For each sampled transition $(s_i, a_i, r_i, s'_i, d_i)$, the TD target is computed as $y_i = r_i + \gamma (1 - d_i) \max_{a'} Q_{\bar{\theta}}(s'_i, a')$. The DQN parameters are updated by minimizing the mean-squared TD error in $B$ mini-batches is:
\begin{equation}
\mathcal{L}(\theta) = \frac{1}{B} \sum_{i=1}^{B} \left( Q_\theta(s_i, a_i) - y_i \right)^2.
\label{eq:dqn_loss}
\end{equation}
For action selection, we use the standard $\epsilon$-greedy exploration strategy with exponential decay~\cite{hariharan2022brief}.

\begin{algorithm}[!ht]
\caption{SFL with RL-based DQN Training}
\label{alg:committee_dqn}
\begin{algorithmic}[1]
\renewcommand{\algorithmicrequire}{\textbf{Input:}}
\renewcommand{\algorithmicensure}{\textbf{Output:}}
\REQUIRE $M$ (odd), $L$, $\mathcal{A}$, $\eta$, $\gamma$, $\epsilon_0$, $\epsilon_{\min}$, $\kappa$, $\mathcal{C}$, $N_{\max}$, $B$, $\{\mathcal{K}_c\}$, $T$, $E$
\ENSURE Trained committee $\{Q_\theta^{(m)}\}_{m=1}^{M}$; policy $\pi$
\STATE \textbf{Model form (shared encoder, per-member head):}
\[
f_s(\cdot;\phi_s)\text{ (shared)}, \quad g^{(m)}(\cdot;\psi^{(m)})\text{ (head)}
\]
\[
Q^{(m)}_\theta(s,a)=\big[g^{(m)}(f_s(s;\phi_s);\psi^{(m)})\big]_a;
\theta_m = f_s + g^{(m)}
\]
\STATE Initialize shared encoder $f_s(\cdot; \phi_s)$, heads $\{g^{(m)}\}_{m=1}^{M}$; set $Q_{\bar{\theta}}^{(m)} \leftarrow Q_\theta^{(m)}$ for all $m$
\STATE Initialize replay buffer $\mathcal{D} \leftarrow \emptyset$
\FOR{episode $e = 1$ to $E$}
    \FOR{cluster $c = 1$ to $C$}
        \STATE Set $\epsilon \leftarrow \epsilon_0$; $L_0 \leftarrow \infty$
        \FOR{round $t = 1$ to $T$}
            \STATE Compute capability metrics $C_{\text{CPU}}^{(k)}(t)$, $C_{\text{Memory}}^{(k)}(t)$, $C_{\text{Battery}}^{(k)}(t)$, $C_{\text{Network}}^{(k)}(t)$ for all $k \in \mathcal{K}_c$
            \STATE Construct state $s_t^{(c)} = [\bar{C}_{\text{CPU}}^{(c)}, \bar{C}_{\text{Memory}}^{(c)}, \bar{C}_{\text{Battery}}^{(c)}, \bar{C}_{\text{Network}}^{(c)}, \bar{C}_{\text{Overall}}^{(c)}, \sigma_c(t)]$
            \STATE Select action $a_t$: with prob. $\epsilon$ random from $\mathcal{A}$, else $a_t \leftarrow \texttt{CommitteeVote}(s_t^{(c)}, \{Q_\theta^{(m)}\})$ (Alg.~\ref{alg:committee_vote})
            \STATE Execute SFL round: $(W_t, L_t) \leftarrow \texttt{TrainRoundSFL}(c, s_t^{(c)}, W_{t-1}, \ell = a_t)$ (Alg.~\ref{alg:train_round})
            \STATE Compute reward $r_t = -(L_t - L_{t-1}) \cdot \rho_t$ where $\rho_t = e^{-\lambda(t-1)}$
            \STATE Obtain next state $s_{t+1}^{(c)}$ by recomputing capability metrics
            \STATE Set terminal flag: $d_t = \mathbf{1}$ [if t = T]
            \STATE Store $(s_t^{(c)}, a_t, r_t, s_{t+1}^{(c)}, d_t)$ in $\mathcal{D}$; if $|\mathcal{D}| > N_{\max}$, pop oldest
            \STATE Sample mini-batch $\{(s_i, a_i, r_i, s'_i, d_i)\}_{i=1}^{B}$ from $\mathcal{D}$
            \FOR{$m = 1$ to $M$}
                \STATE Compute TD target: $y_i^{(m)} = r_i + \gamma(1-d_i)\max_{a'} Q_{\bar{\theta}}^{(m)}(s'_i, a')$
                \STATE Update: $\theta^{(m)} \leftarrow \theta^{(m)} - \eta \nabla_{\theta^{(m)}} \frac{1}{B}\sum_{i}(Q_\theta^{(m)}(s_i, a_i) - y_i^{(m)})^2$
                \IF{$t \mod \mathcal{C} = 0$}
                    \STATE $Q_{\bar{\theta}}^{(m)} \leftarrow Q_\theta^{(m)}$
                \ENDIF
            \ENDFOR
            \STATE Decay: $\epsilon \leftarrow \epsilon_{\min} + (\epsilon_0 - \epsilon_{\min})e^{-\kappa t}$; $L_{t-1} \leftarrow L_t$
        \ENDFOR
    \ENDFOR
\ENDFOR
\RETURN $\{Q_\theta^{(m)}\}_{m=1}^{M}$, $\pi(s) = \texttt{CommitteeVote}(s, \{Q_\theta^{(m)}\})$
\end{algorithmic}
\end{algorithm}

\textbf{Committee-Based Solution for Mitigating Reward Hacking:} A single agent can sometimes learn to cheat by finding shortcuts that increase its reward without actually improving the model's real performance; this phenomenon is known as reward hacking~\cite{wu2024towards}. As shown in Step~4 of Figure~\ref{fig:qsplitfl_architecture}, the RL Controller's Committee Machine employs $M$ MLP Heads with a Shared Encoder (with $M$ being an odd number) that vote together via Majority Voting (Step~4a) to decide the best split action (Step~4b). Each committee member shares a common shallow encoder $f_s(\cdot; \phi_s)$ for stable low-level state representation, but maintains an independent deep head $g^{(m)}(\cdot; \psi^{(m)})$ for diverse Q-value estimation:
\begin{equation}
Q_\theta^{(m)}(s, a) = \left[ g^{(m)}\left( f_s(s; \phi_s); \psi^{(m)} \right) \right]_a.
\label{eq:committee_q}
\end{equation}
By sharing the early layers but keeping separate deeper layers, the committee members learn different perspectives on which actions are best, without adding too much computational cost. During action selection, each committee member proposes its preferred split point by selecting the action with the highest Q-value. We then count the votes for each action and select the action with the most votes as the final decision. In case of ties, we break by selecting the action with the highest mean Q-value across committee members. This majority voting mechanism ensures that no single model can exploit loopholes in the reward function.
It is worth clarifying the relationship between the committee structure and the broader RL formulation. QSplitFL is designed as a single-agent RL system. A single centralized RL Controller on the server observes one global state $s_t^{(c)}$ aggregating all client capability metrics and outputs one global action, the split layer $\ell$, applied uniformly in that round. The $M$ MLP heads are not independent agents. Rather, they form an ensemble of Q-value estimators within the single agent, each sharing the same encoder but maintaining separate deeper layers~\cite{wu2024towards}. A true multi-agent formulation would require each client to act as an independent decision maker, introducing non-stationary dynamics and coordination overhead. Since QSplitFL needs one globally consistent split point per cluster at each round, the single-agent MDP formulation is both sufficient and appropriate. Beyond architectural clarity, this ensemble design serves a critical functional purpose which is mitigating reward hacking. To verify this, we conducted a controlled comparison between a single-head DQN ($M$=1) and the committee DQN ($M$=3). The single-head agent converged to a fixed split point early in training, accumulating high reward through early loss-drops while accuracy plateaued. The committee DQN, in contrast, continued adapting its split selection across rounds and achieved significantly higher final accuracy, which validates majority voting as an effective defense against reward hacking.

\subsection{Training Procedure}
Algorithm~\ref{alg:committee_dqn} presents the main controller that orchestrates this complete training loop. It manages the experience replay buffer $\mathcal{D}$ and coordinates all committee members. At each round $t$, it computes capability metrics, constructs the six-dimensional state $s_t^{(c)}$, and determines whether to explore (random action with probability $\epsilon$) or exploit (committee voting with probability $1-\epsilon$). Two supporting algorithms are provided in Appendix~\ref{sec:algorithms}: Algorithm~\ref{alg:train_round} (SFL Training Round) executes the FL round corresponding to Steps~5--12 in Figure~\ref{fig:qsplitfl_architecture}, and Algorithm~\ref{alg:committee_vote} (Committee Majority Voting) implements Step~4.

\subsection{Computational Complexity Analysis} A key advantage of the capability-based state representation is its computational efficiency compared to weight-based approaches. Traditional methods using PCA-compressed model weights require high complexity but our capability-based states require only $\mathcal{O}(|\mathcal{K}|)$, which corresponds to simple aggregation operations (mean, standard deviation) over client capability metrics. This becomes especially significant in large scale settings when the network has many layers ($d$) and there are many participating clients ($|\mathcal{K}|$).

\section{Experiments and Results}
\label{sec:experiments}

\subsection{Experimental Setup}
\textbf{Datasets and Architectures:} We evaluate QSplitFL on MNIST, Fashion-MNIST, CIFAR-10, and CIFAR-100, covering a range of complexity from simple grayscale to challenging RGB classification tasks. To assess performance across varying computational settings, we have paired these datasets with four DNN architectures: a 10-layer CNN (split points 5--9), ResNet50 (50 layers, split points 25--49), MobileNetV4 (53 layers, split points 27--52), and ConvNeXt (59 layers, split points 30--58).

\textbf{Heterogeneity and Reproducibility:} To emulate heterogeneous, resource-constrained clients, each client samples its four capability metrics from device-tier distributions (strong, medium, and weak) that fluctuate across rounds, and clients are grouped into capability clusters served by a per-cluster controller. Data are partitioned with a Dirichlet distribution ($\alpha = 0.5$) for non-IID conditions, and for fairness all baselines share the same backbone, data partition, optimizer, and number of rounds, with split-based baselines using the same client and server partition where applicable. The full device-tier ranges, clustering procedure, hyperparameter settings, and baseline tuning protocol are detailed in Appendix~\ref{appendix:setup}, and Appendix~\ref{appendix:extended_discussion} discusses the rationale for the split depth bound, its privacy implications, system-level cost, feature selection, and limitations.



\begin{figure*}[ht]
\centering
\begin{minipage}{\textwidth}
\centering
\begin{subfigure}{0.23\textwidth}\includegraphics[width=\textwidth]{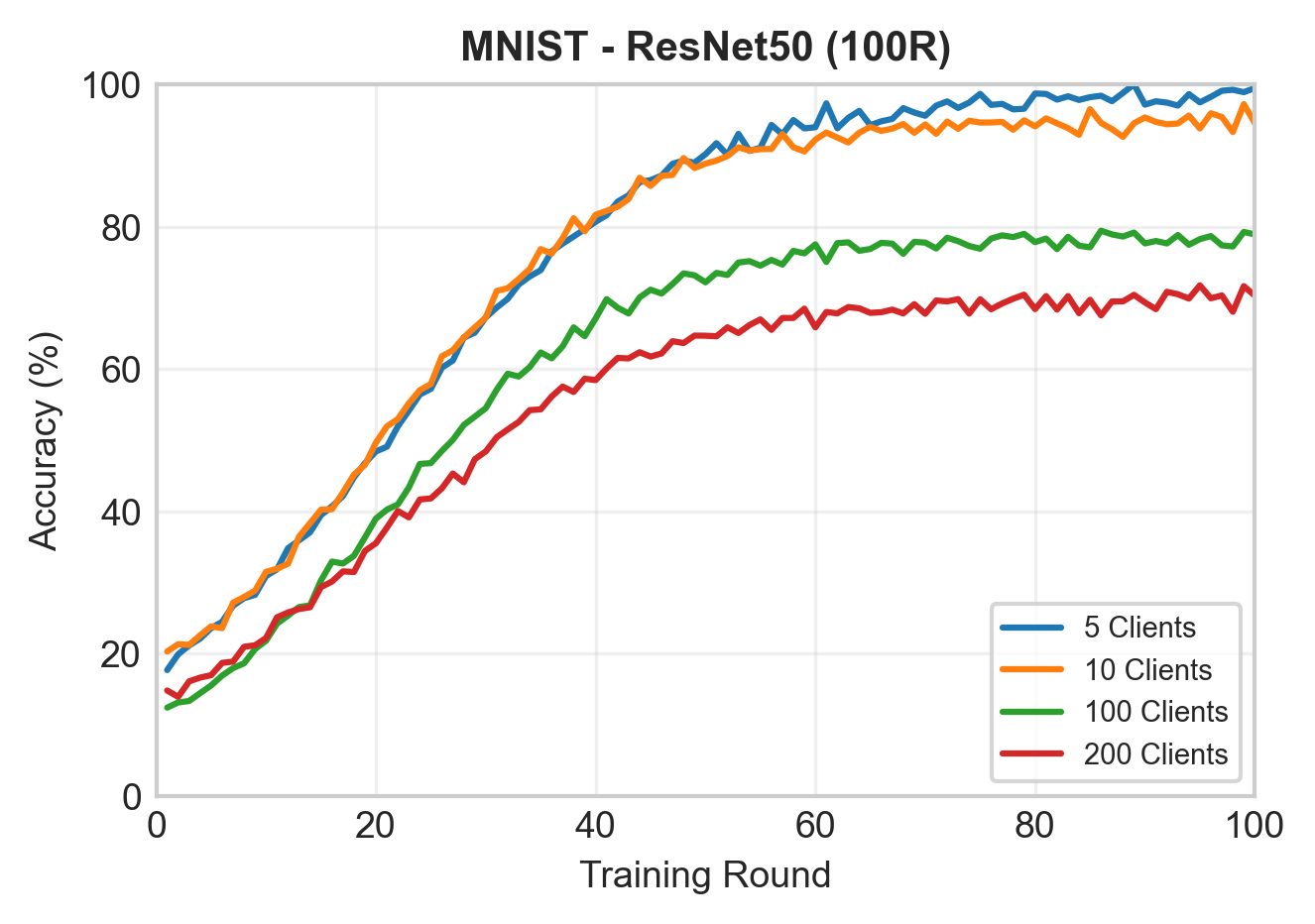}\caption{MNIST: ResNet50}\end{subfigure}
\begin{subfigure}{0.23\textwidth}\includegraphics[width=\textwidth]{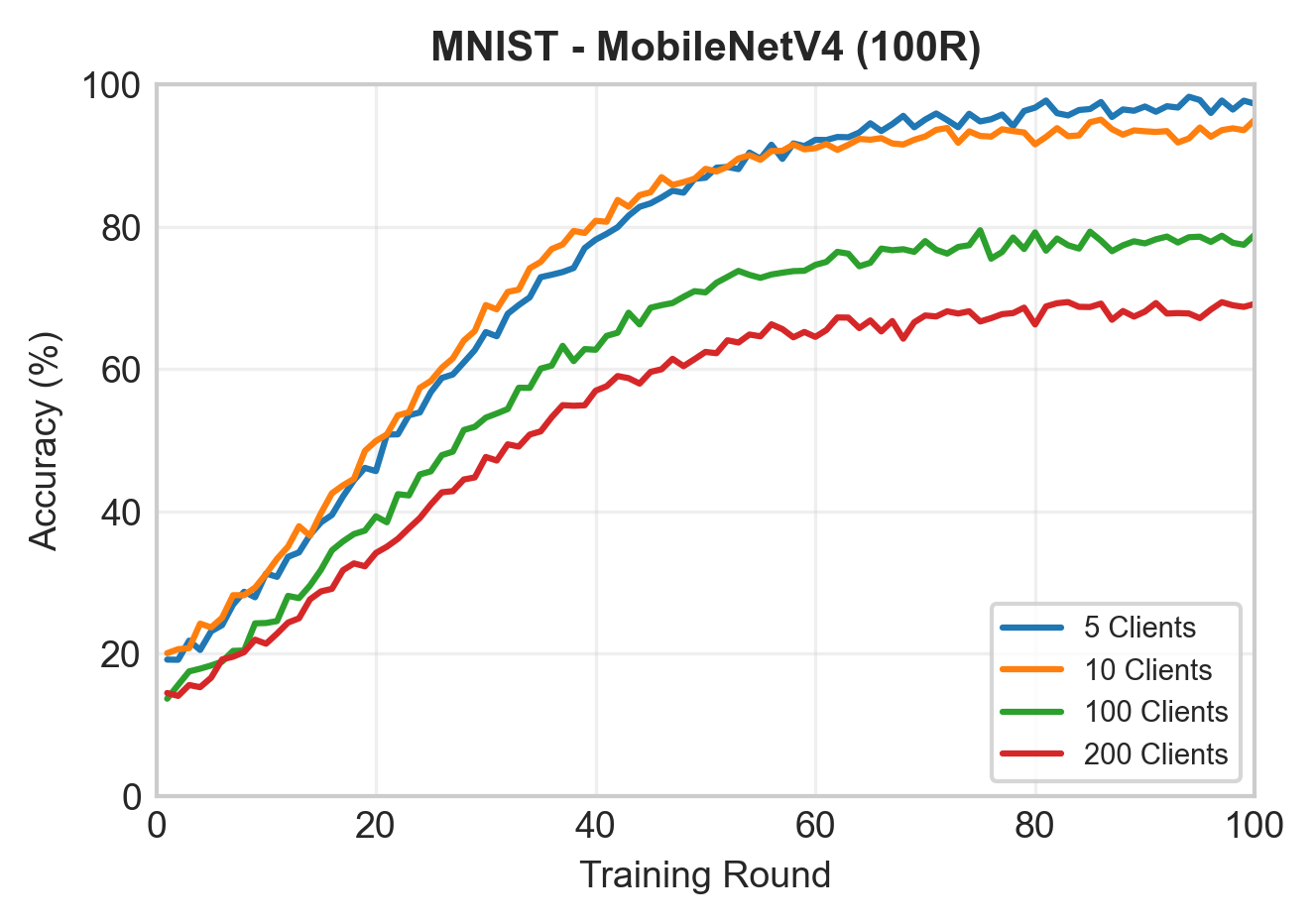}\caption{MNIST: MobileNetV4}\end{subfigure}
\begin{subfigure}{0.23\textwidth}\includegraphics[width=\textwidth]{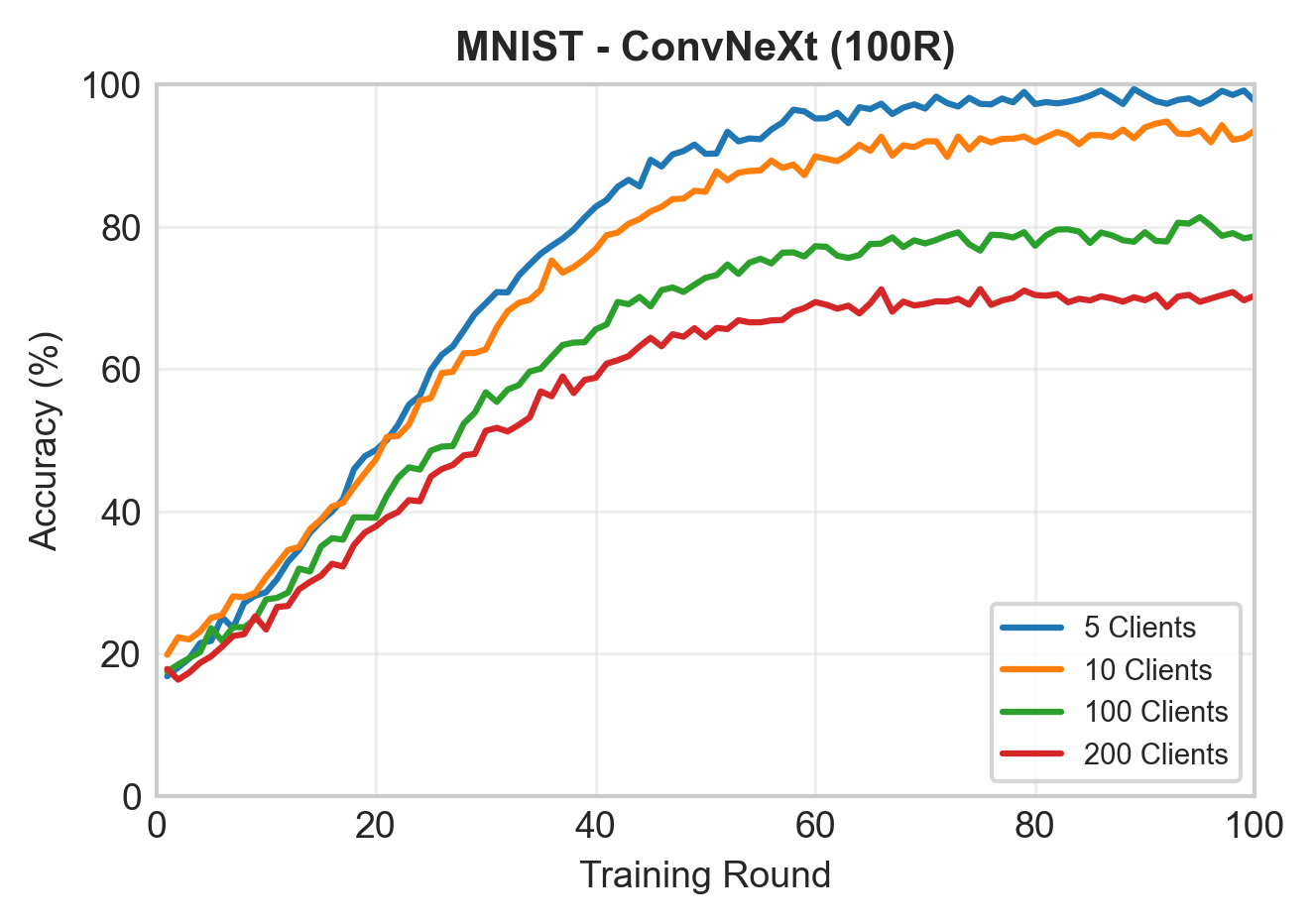}\caption{MNIST: ConvNeXt}\end{subfigure}
\begin{subfigure}{0.23\textwidth}\includegraphics[width=\textwidth]{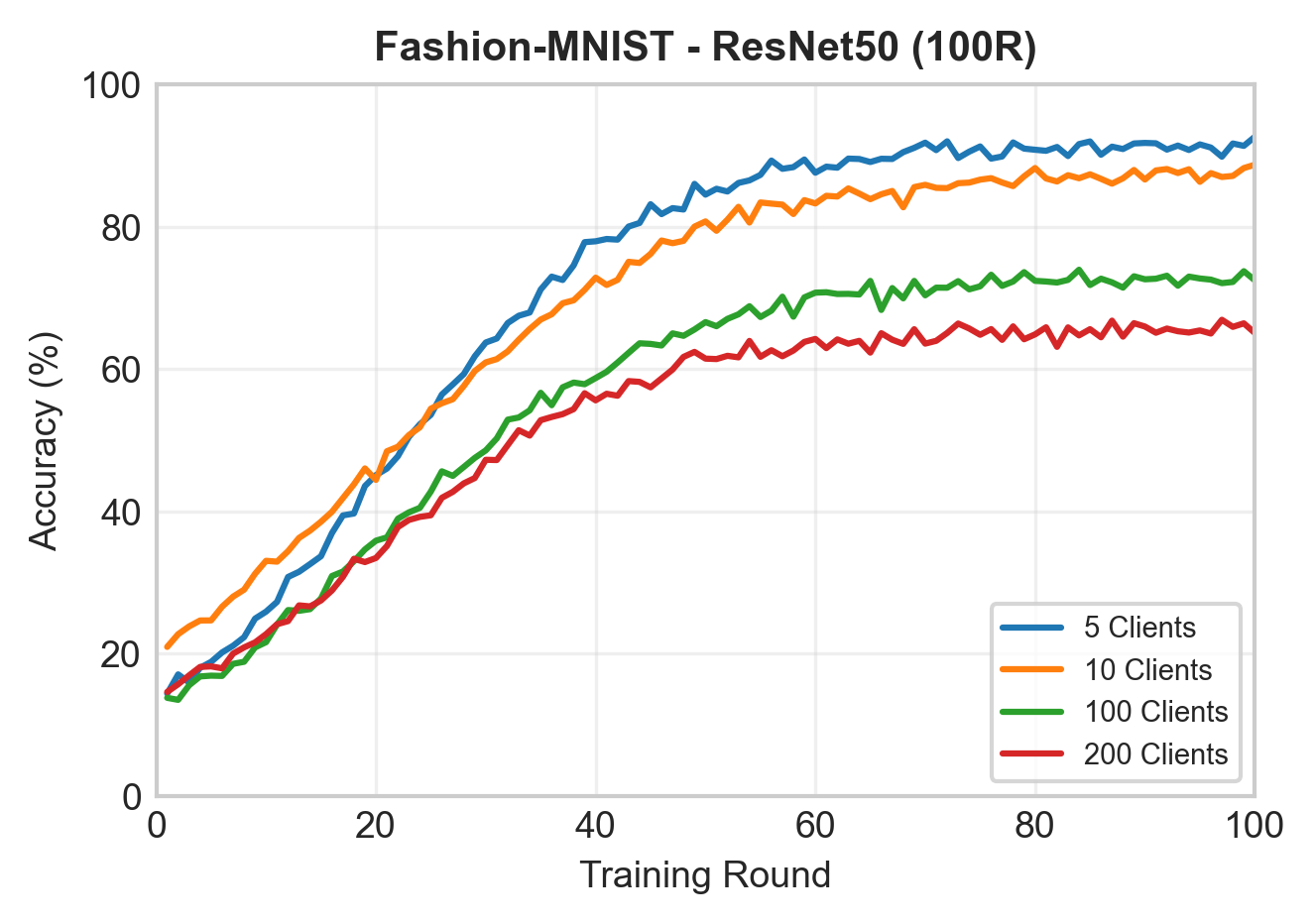}\caption{FMNIST: ResNet50}\end{subfigure}
\begin{subfigure}{0.23\textwidth}\includegraphics[width=\textwidth]{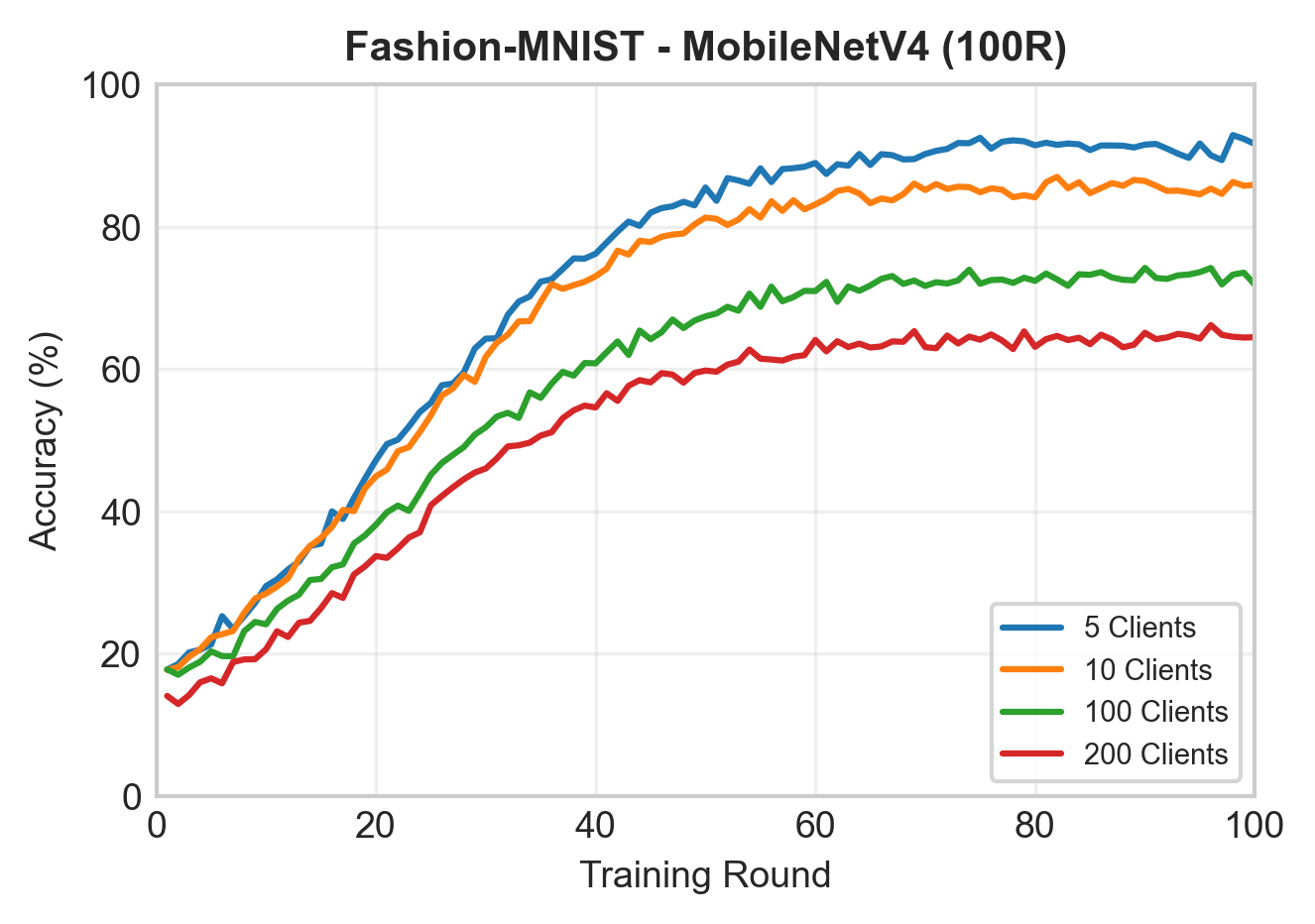}\caption{FMNIST: MobileNetV4}\end{subfigure}
\begin{subfigure}{0.23\textwidth}\includegraphics[width=\textwidth]{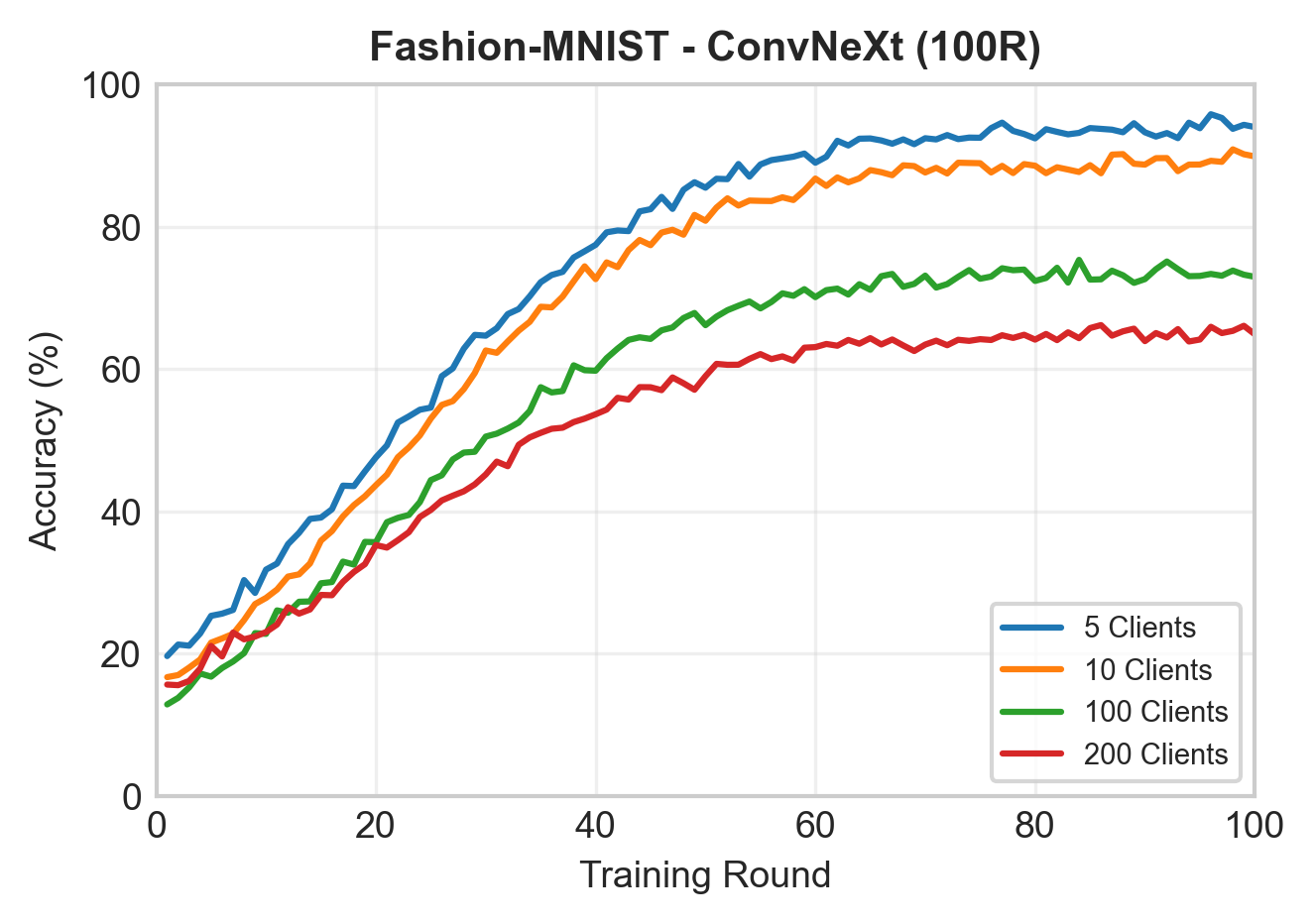}\caption{FMNIST: ConvNeXt}\end{subfigure}
\begin{subfigure}{0.23\textwidth}\includegraphics[width=\textwidth]{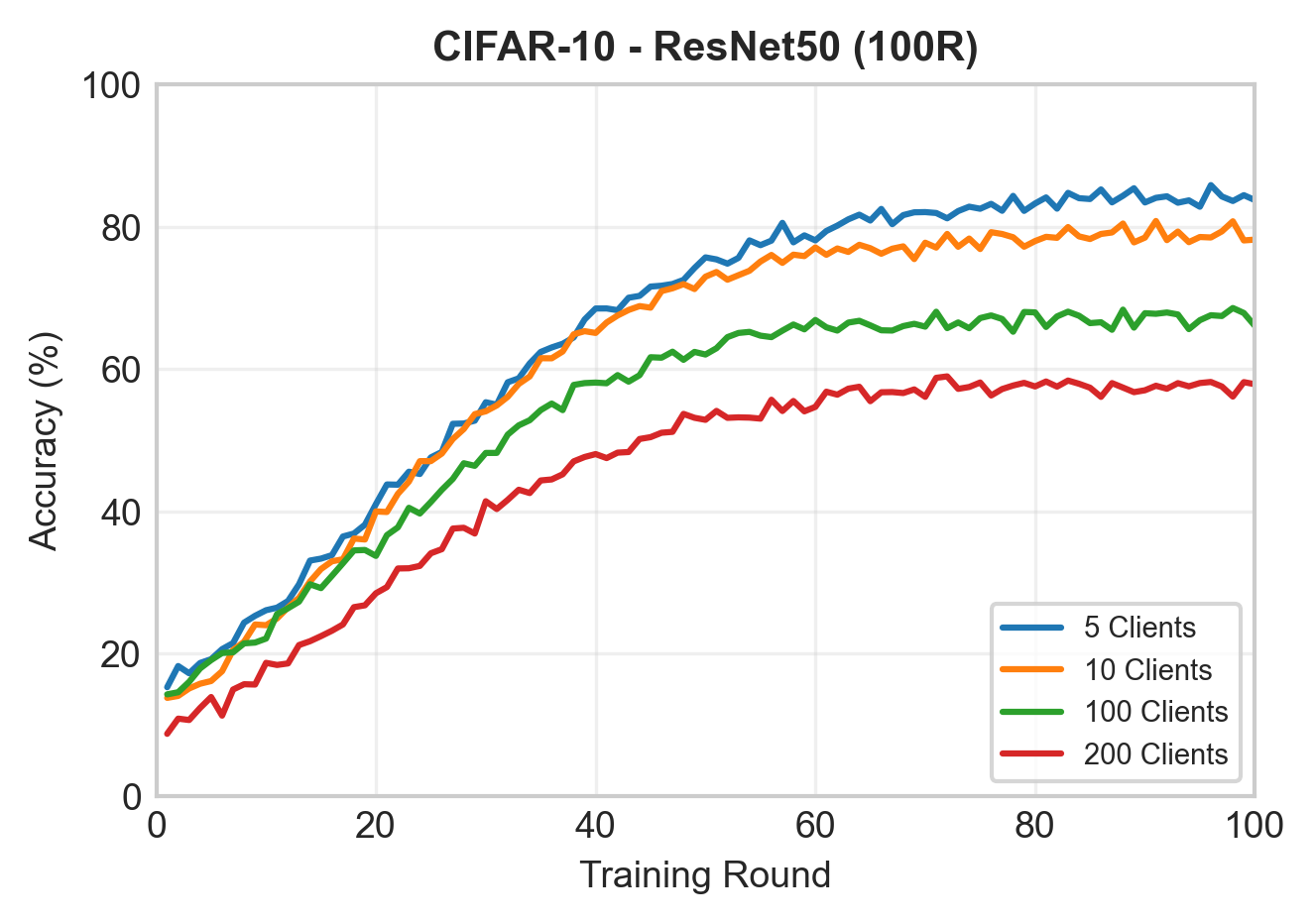}\caption{CIFAR10: ResNet50}\end{subfigure}
\begin{subfigure}{0.23\textwidth}\includegraphics[width=\textwidth]{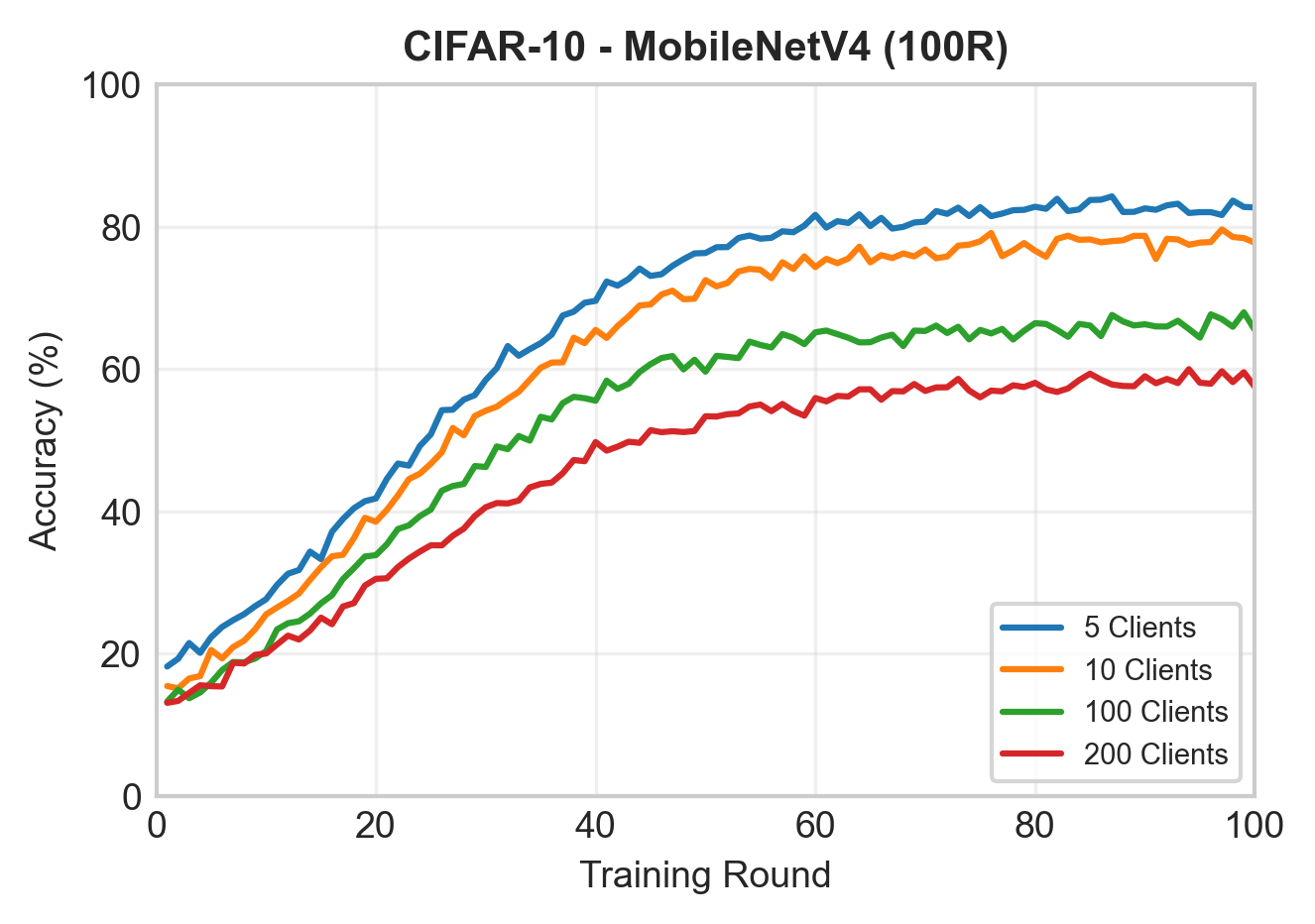}\caption{CIFAR10: MobileNetV4}\end{subfigure}
\begin{subfigure}{0.23\textwidth}\includegraphics[width=\textwidth]{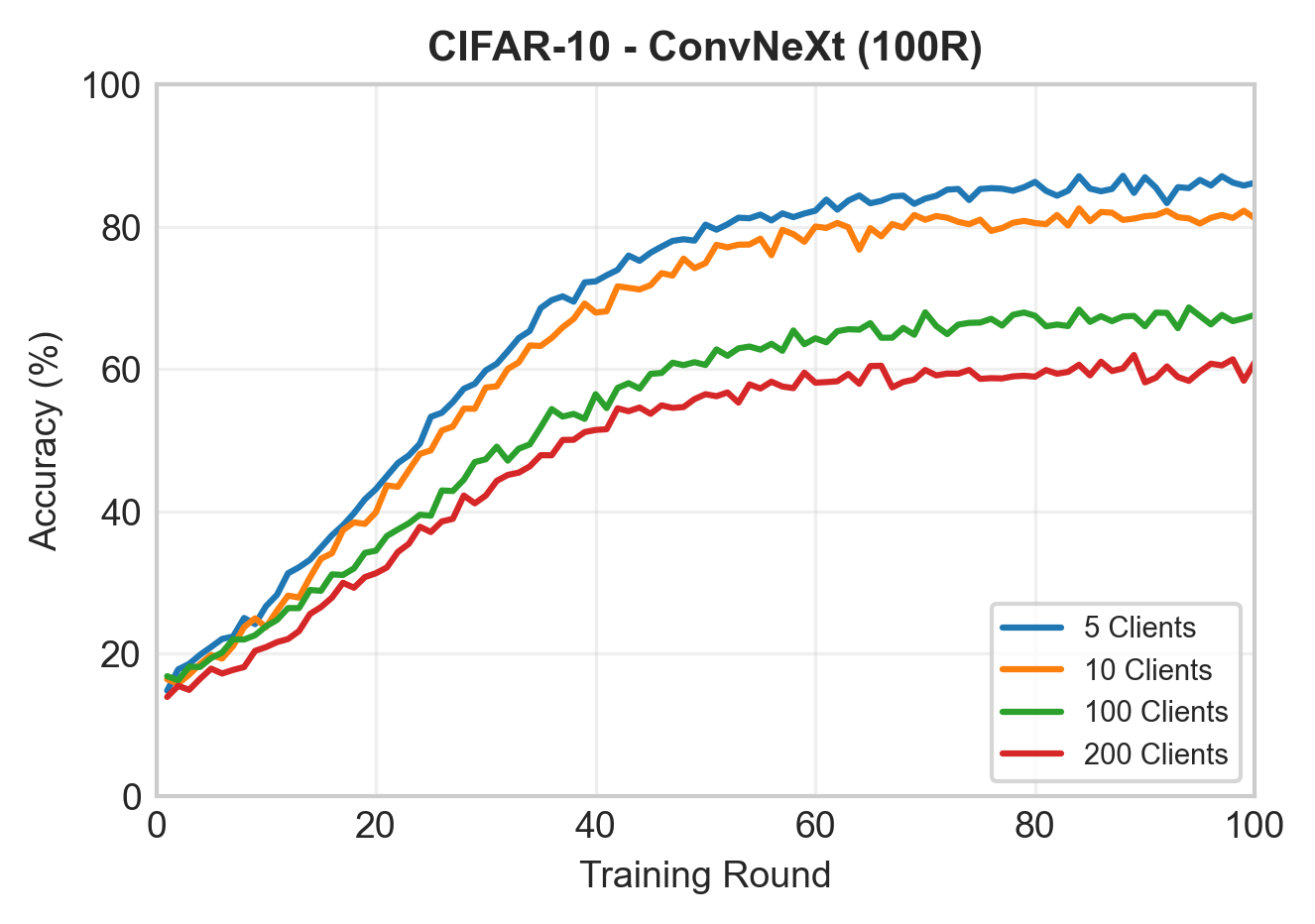}\caption{CIFAR10: ConvNeXt}\end{subfigure}
\begin{subfigure}{0.23\textwidth}\includegraphics[width=\textwidth]{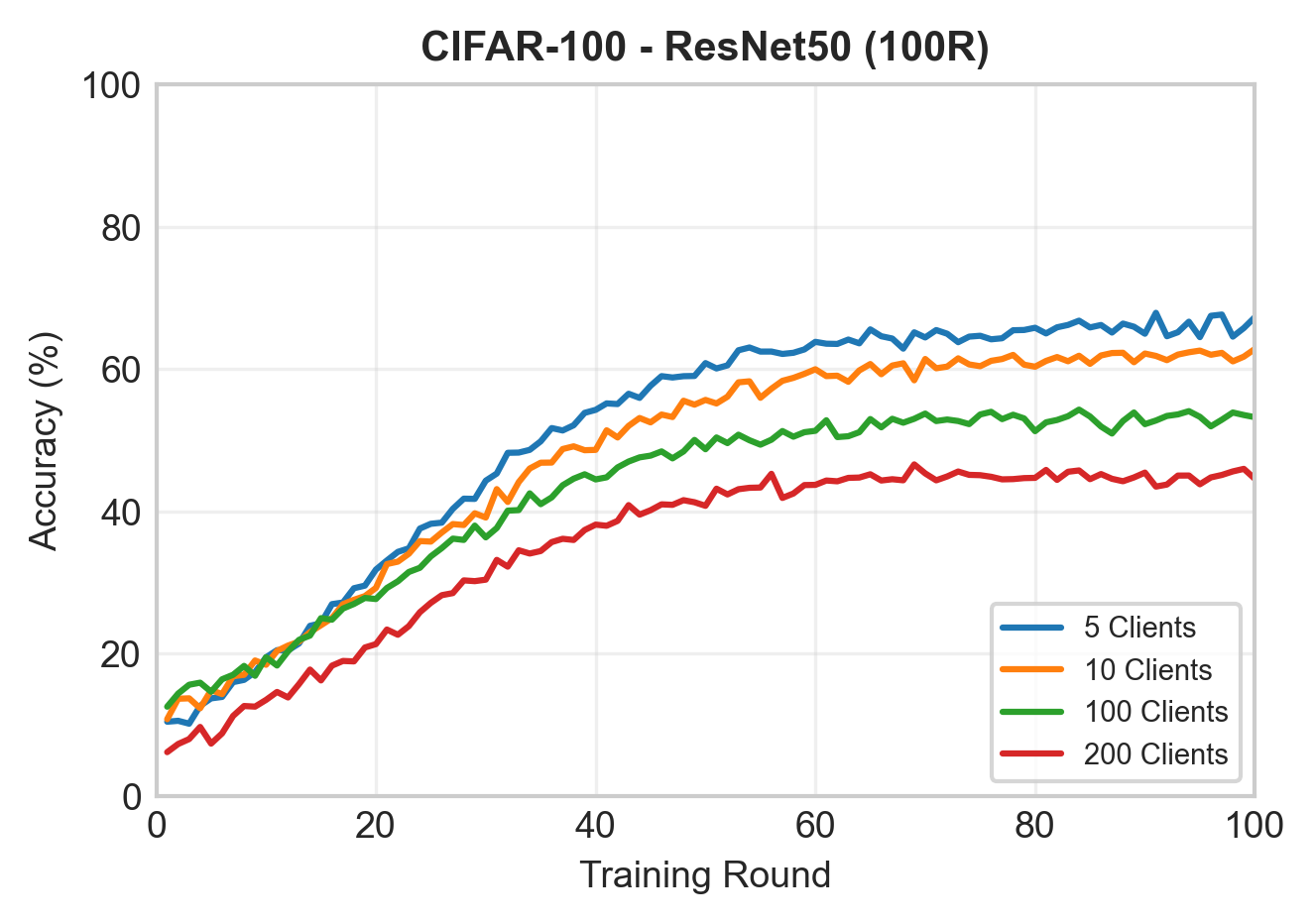}\caption{CIFAR100: ResNet50}\end{subfigure}
\begin{subfigure}{0.23\textwidth}\includegraphics[width=\textwidth]{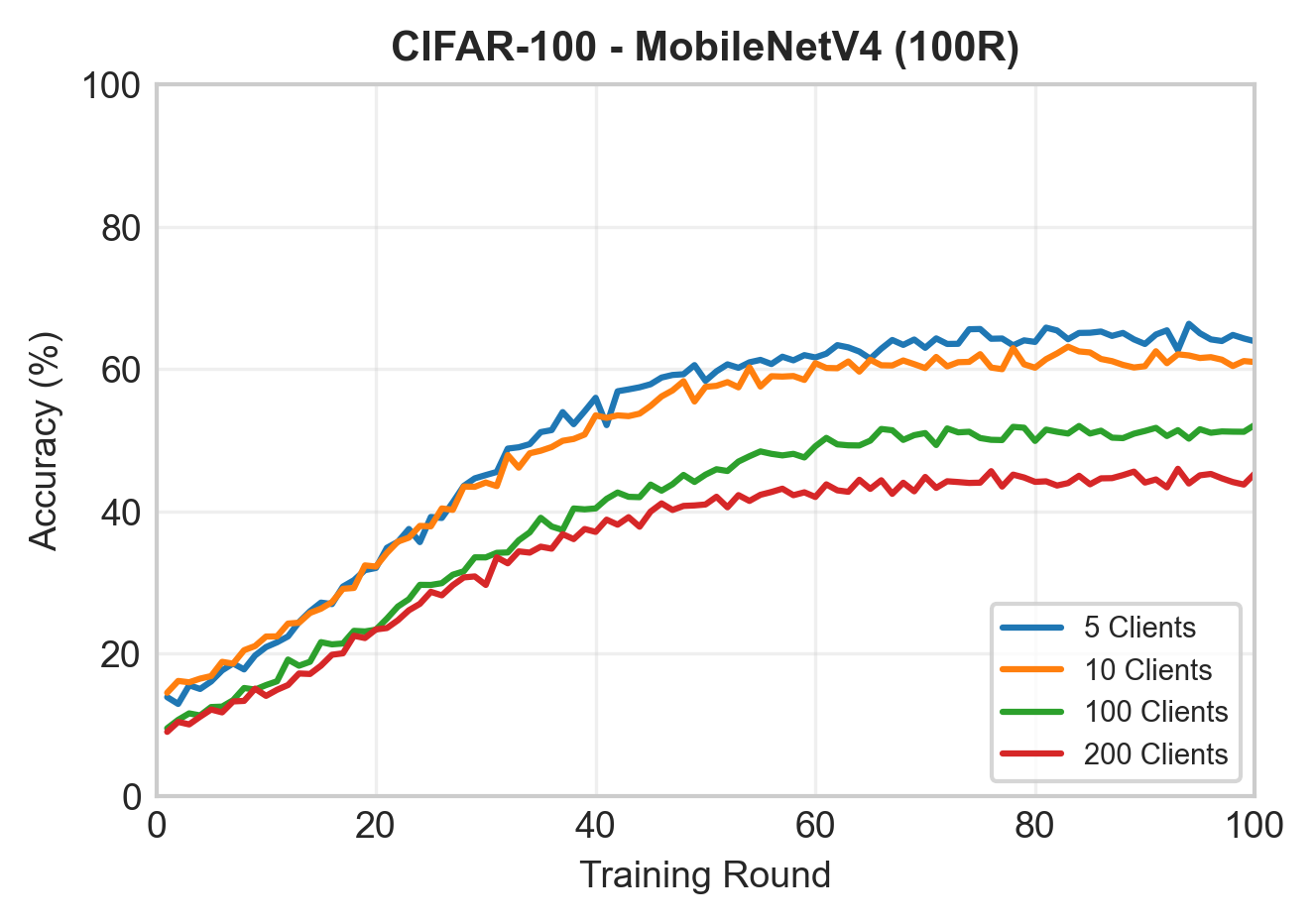}\caption{CIFAR100: MobileNetV4}\end{subfigure}
\begin{subfigure}{0.23\textwidth}\includegraphics[width=\textwidth]{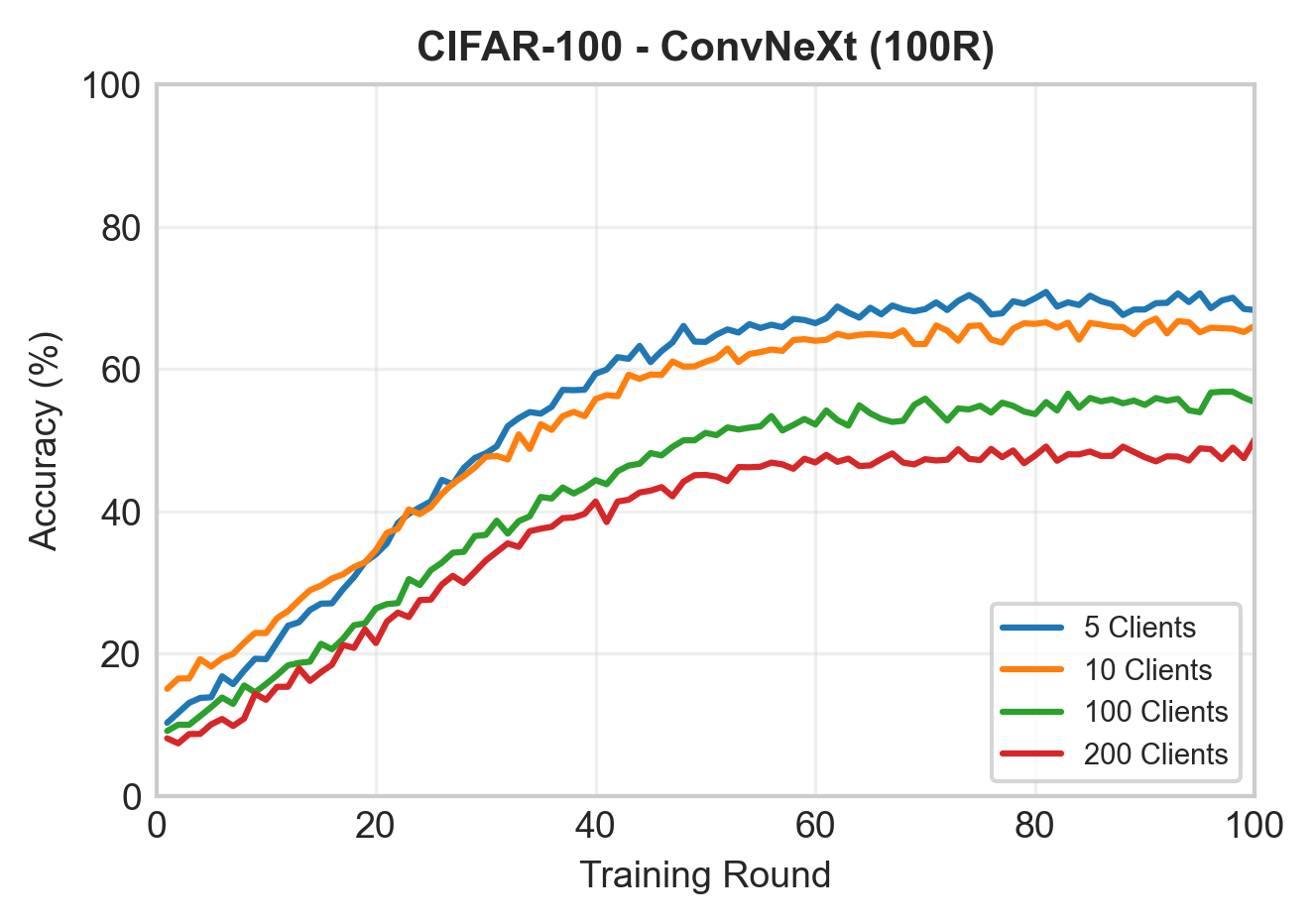}\caption{CIFAR100: ConvNeXt}\end{subfigure}
\caption{\textbf{Accuracy Convergence Analysis (100 Rounds).} Comparison of ResNet50, MobileNetV4, and ConvNeXt across MNIST, Fashion-MNIST, CIFAR-10, and CIFAR-100 with varying client counts. CNN results are detailed in Appendix~\ref{sec:additional_results}.}
\label{fig:acc_100r_all}
\end{minipage}
\end{figure*}

\subsection{Accuracy Convergence Analysis (100 Rounds)}
Figure~\ref{fig:acc_100r_all} shows how accuracy improves over 100 training rounds for all datasets and architectures (except CNN). Deeper models consistently outperform shallow CNNs. ConvNeXt achieves the best results: 99.6\% on MNIST, 94.1\% on Fashion-MNIST, 86.5\% on CIFAR-10, and 68.3\% on CIFAR-100. Across all datasets, we observe that accuracy increases sharply in the first 20--30 rounds and then gradually stabilizes. This pattern holds regardless of the number of clients (5, 10, 50, 100, or 200), showing that QSplitFL scales well to larger federated settings. The performance gap between architectures becomes more noticeable on harder datasets like CIFAR-100, where ConvNeXt performs better than ResNet50 and MobileNetV4. CNN results and additional comparisons are provided in Appendix~\ref{sec:additional_results}.

\subsection{Split Point Selection Analysis}
\begin{figure*}[ht]
\centering
\begin{minipage}{\textwidth}
\centering
\begin{subfigure}{0.23\textwidth}\includegraphics[width=\textwidth]{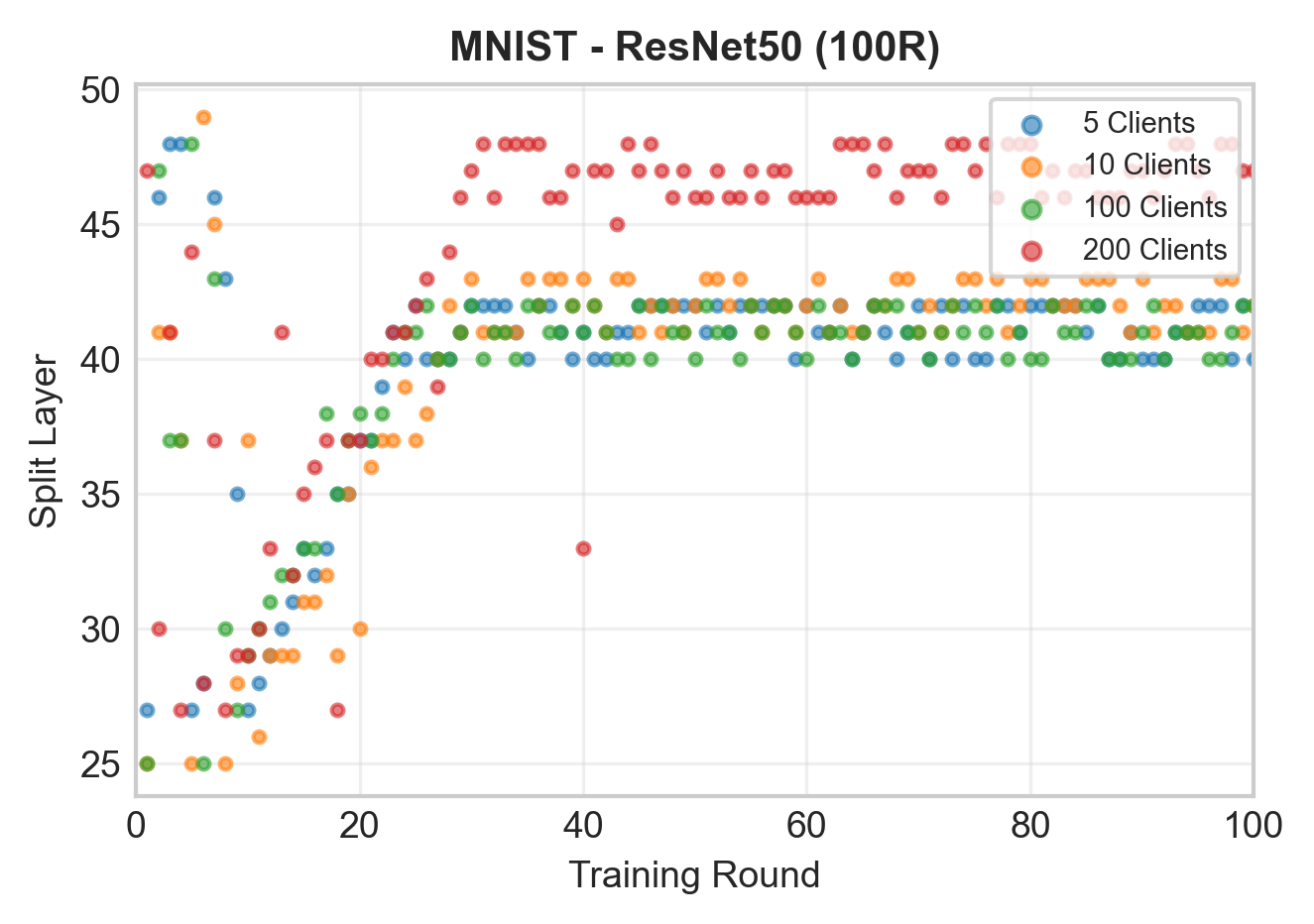}\caption{MNIST: ResNet50}
\end{subfigure}
\begin{subfigure}{0.23\textwidth}\includegraphics[width=\textwidth]{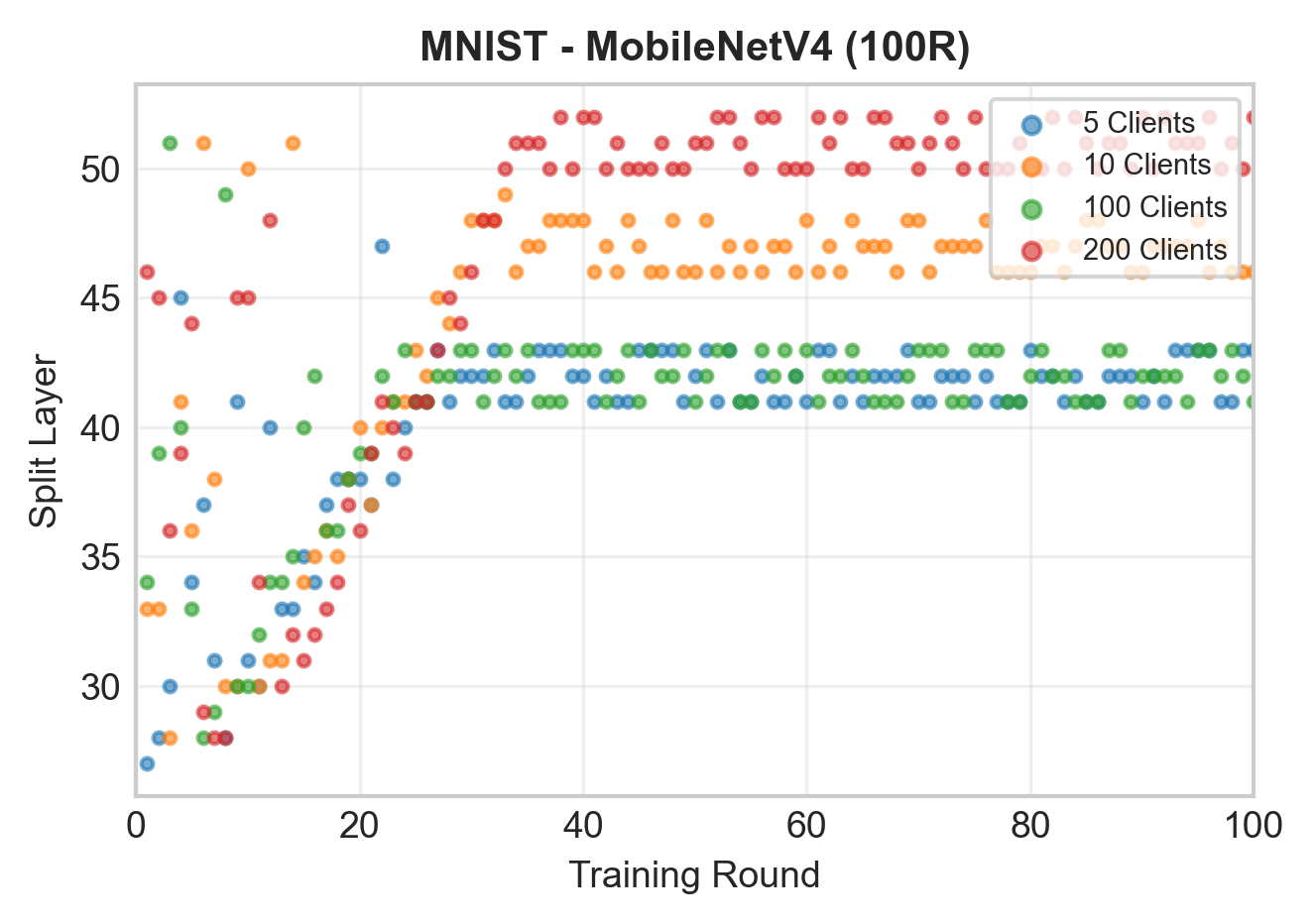}\caption{MNIST: MobileNetV4}
\end{subfigure}
\begin{subfigure}{0.23\textwidth}\includegraphics[width=\textwidth]{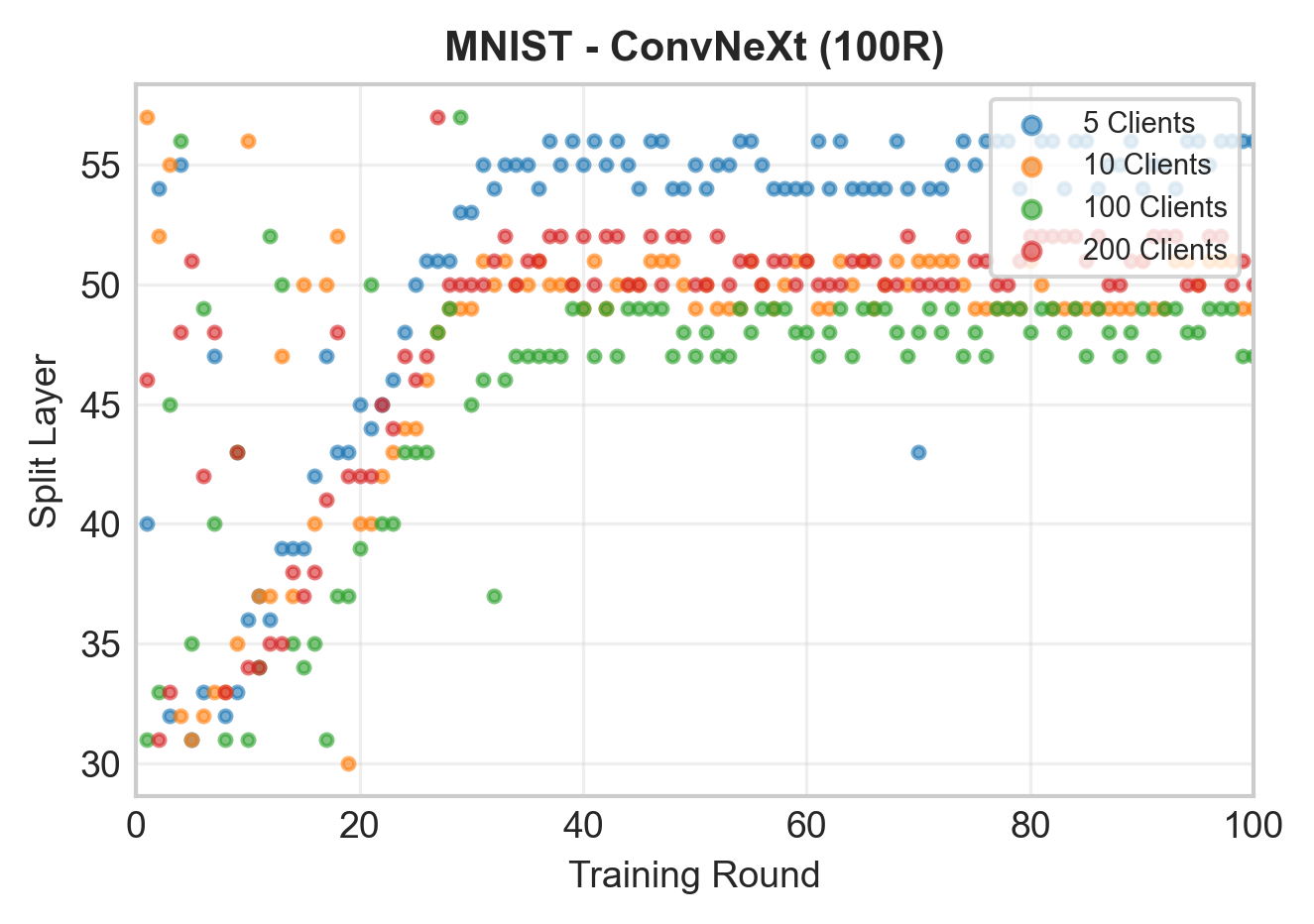}\caption{MNIST: ConvNeXt}
\end{subfigure}
\begin{subfigure}{0.23\textwidth}\includegraphics[width=\textwidth]{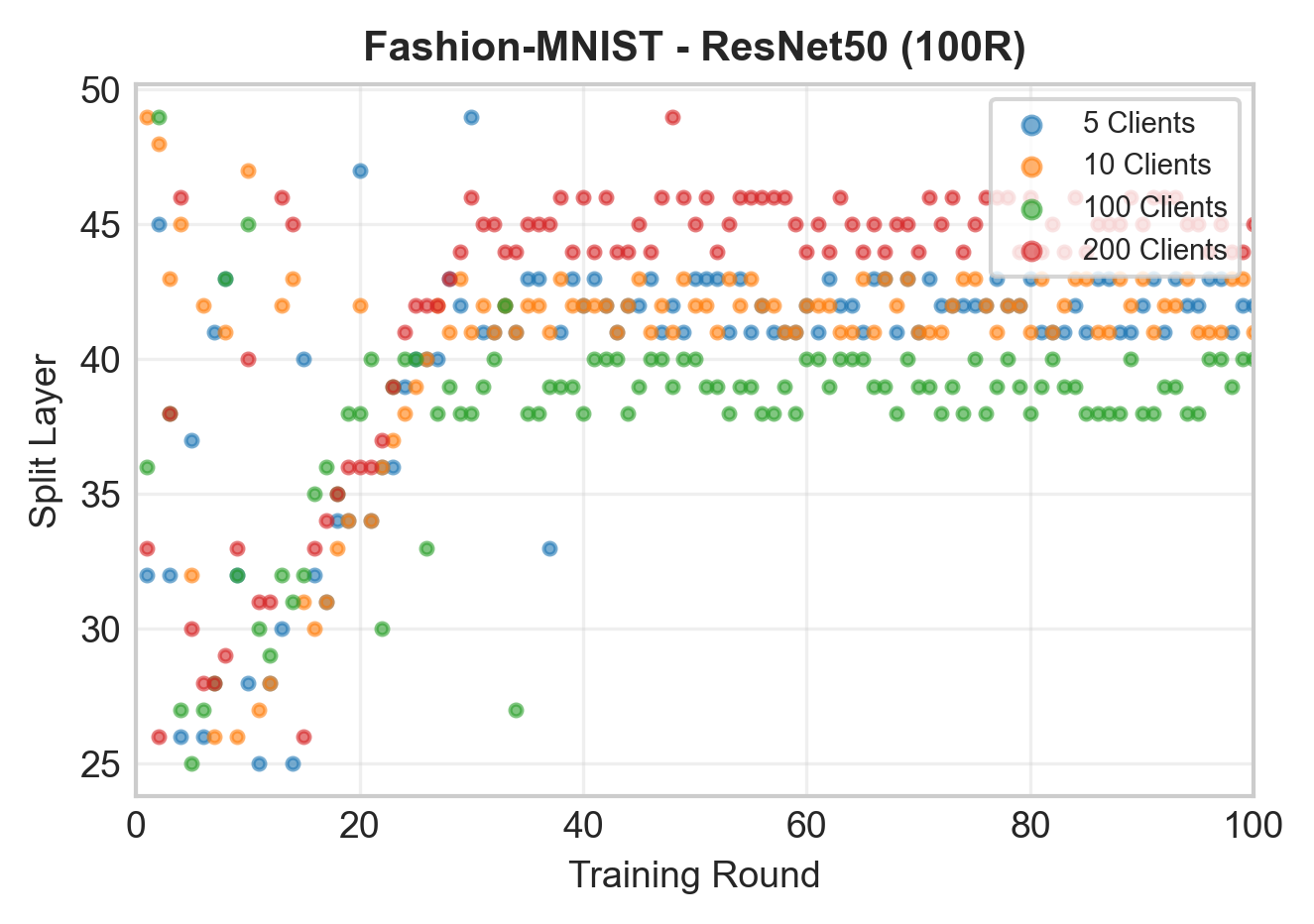}\caption{FMNIST: ResNet50}
\end{subfigure}
\begin{subfigure}{0.23\textwidth}\includegraphics[width=\textwidth]{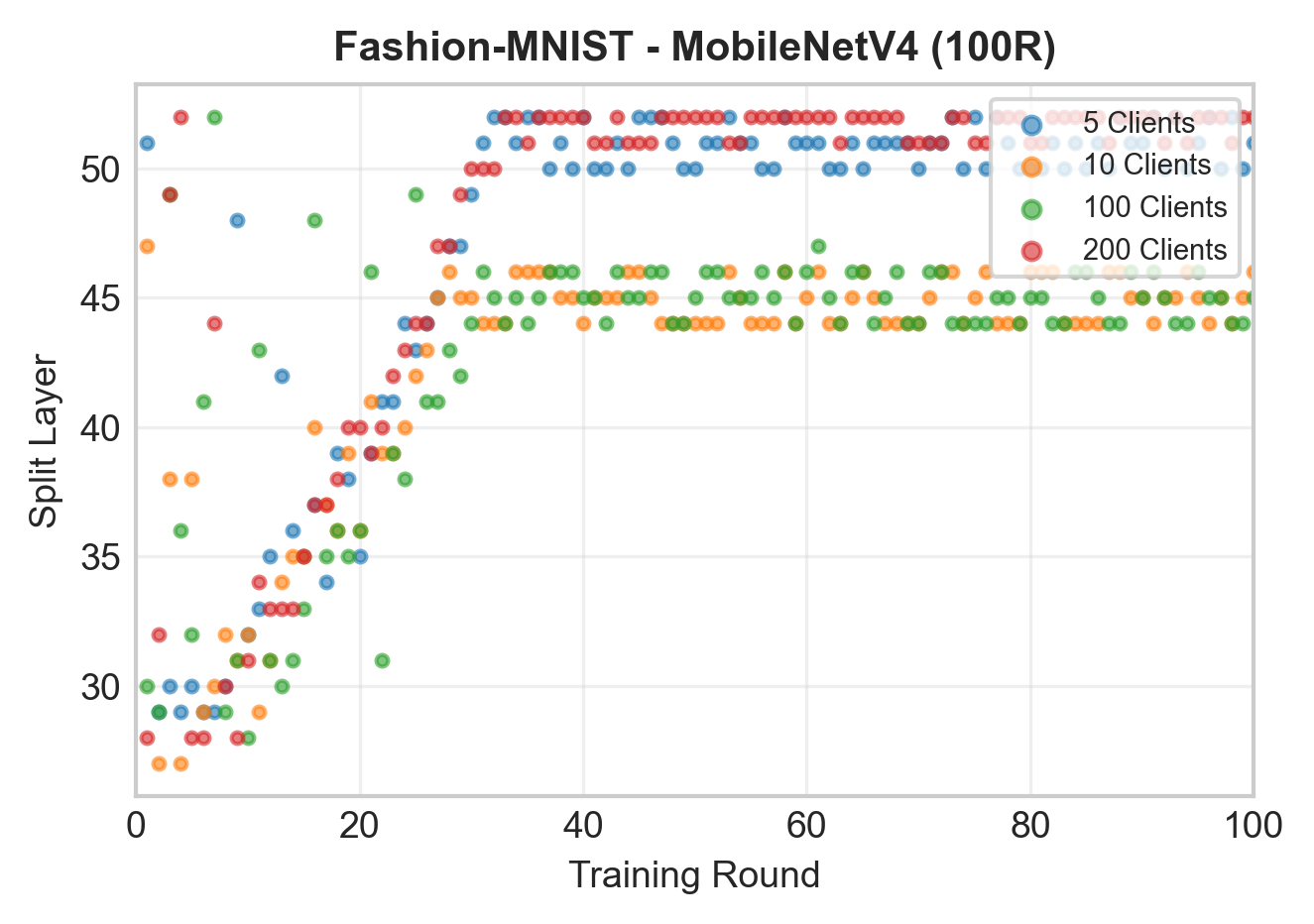}\caption{FMNIST: MobileNetV4}
\end{subfigure}
\begin{subfigure}{0.23\textwidth}\includegraphics[width=\textwidth]{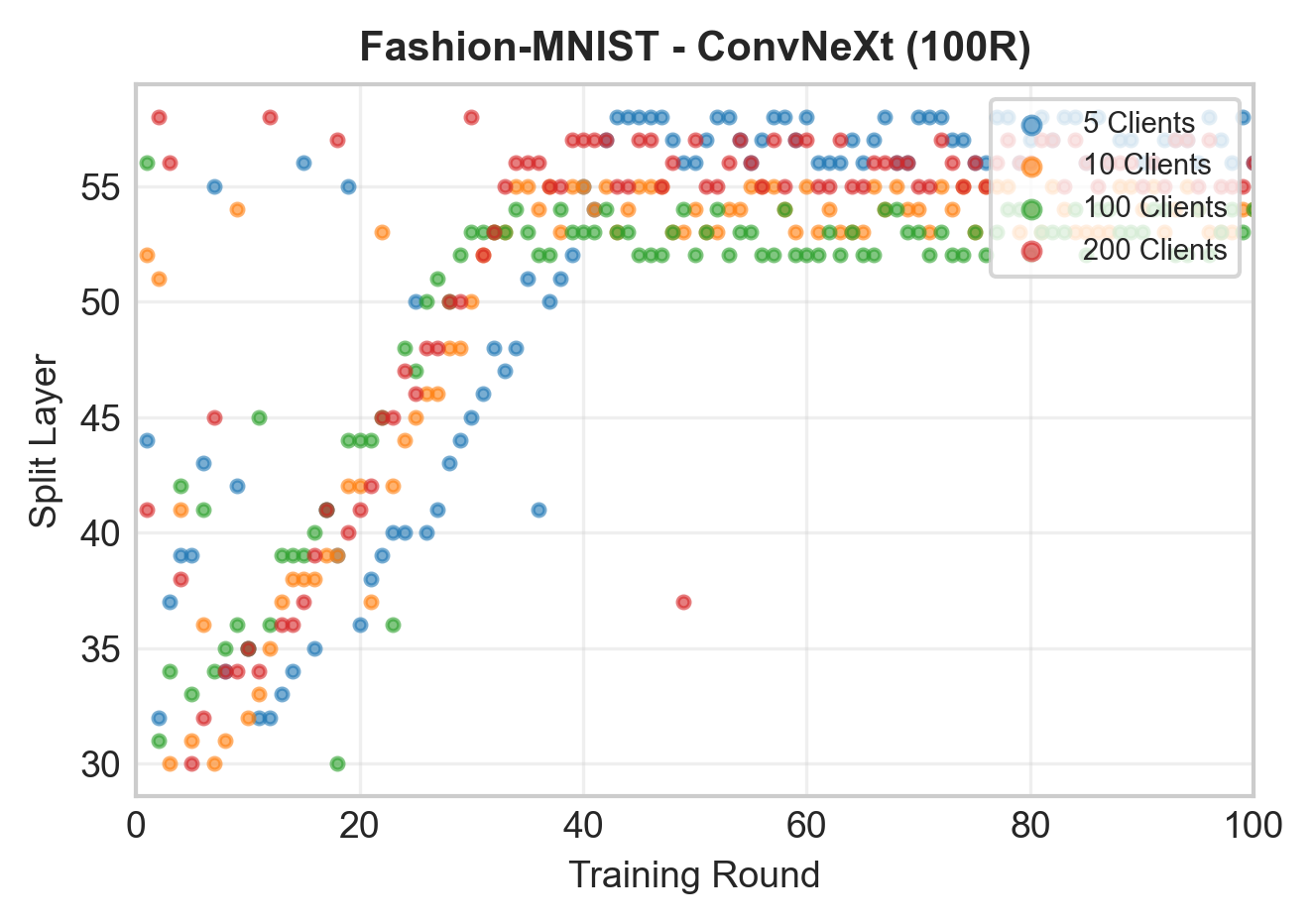}\caption{FMNIST: ConvNeXt}
\end{subfigure}
\begin{subfigure}{0.23\textwidth}\includegraphics[width=\textwidth]{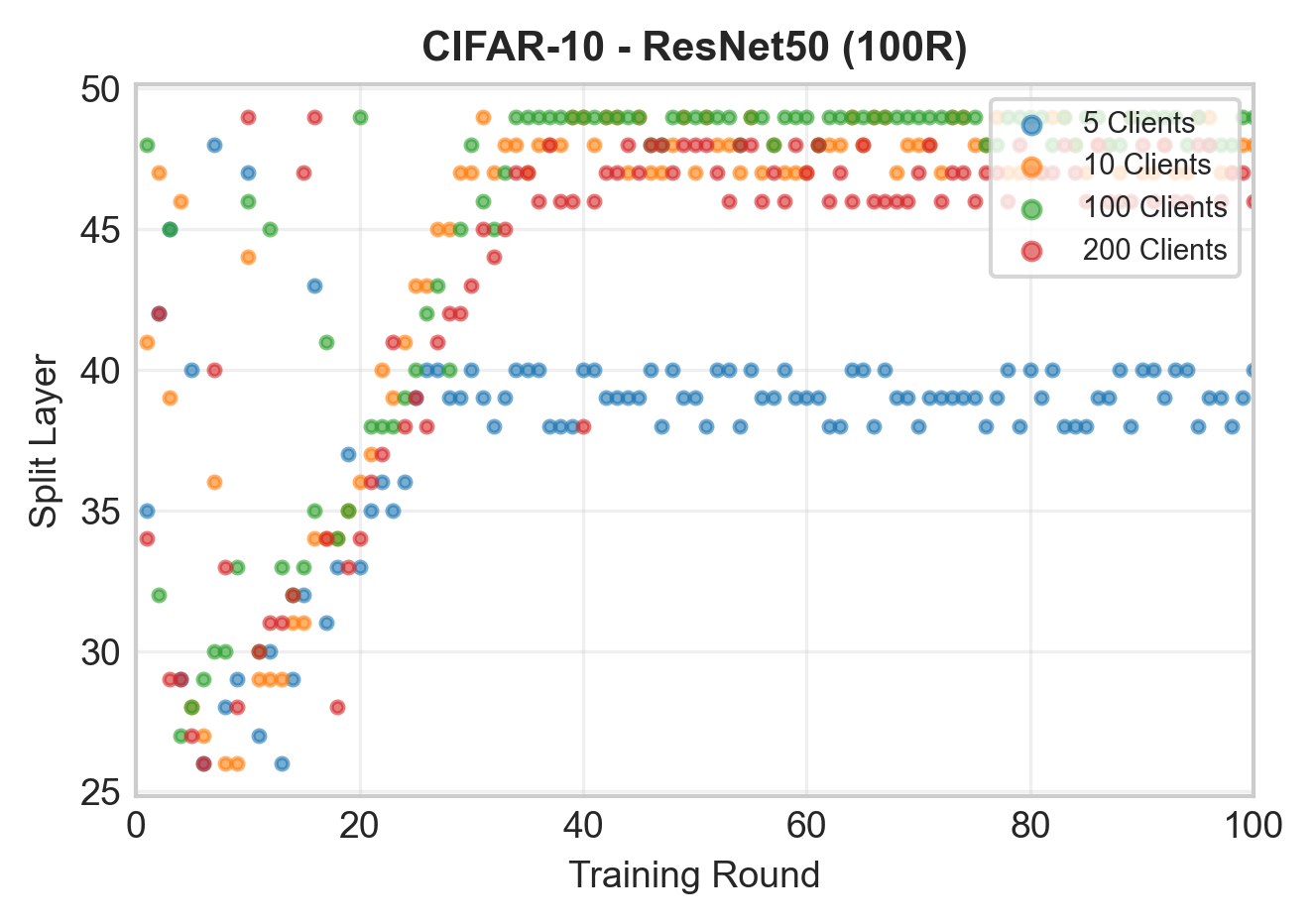}\caption{CIFAR10: ResNet50}
\end{subfigure}
\begin{subfigure}{0.23\textwidth}\includegraphics[width=\textwidth]{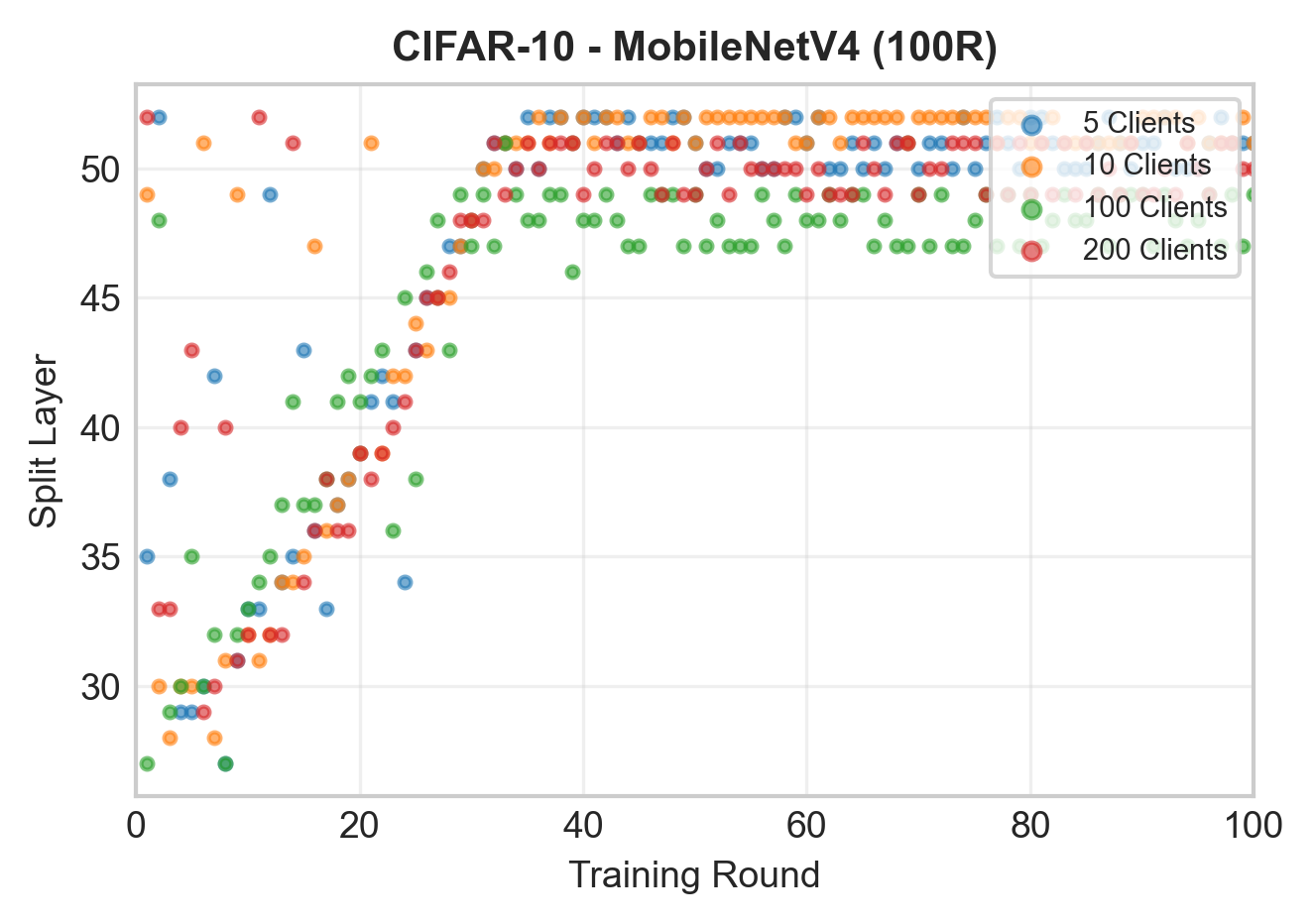}\caption{CIFAR10:MobileNetV4}
\end{subfigure}
\begin{subfigure}{0.23\textwidth}\includegraphics[width=\textwidth]{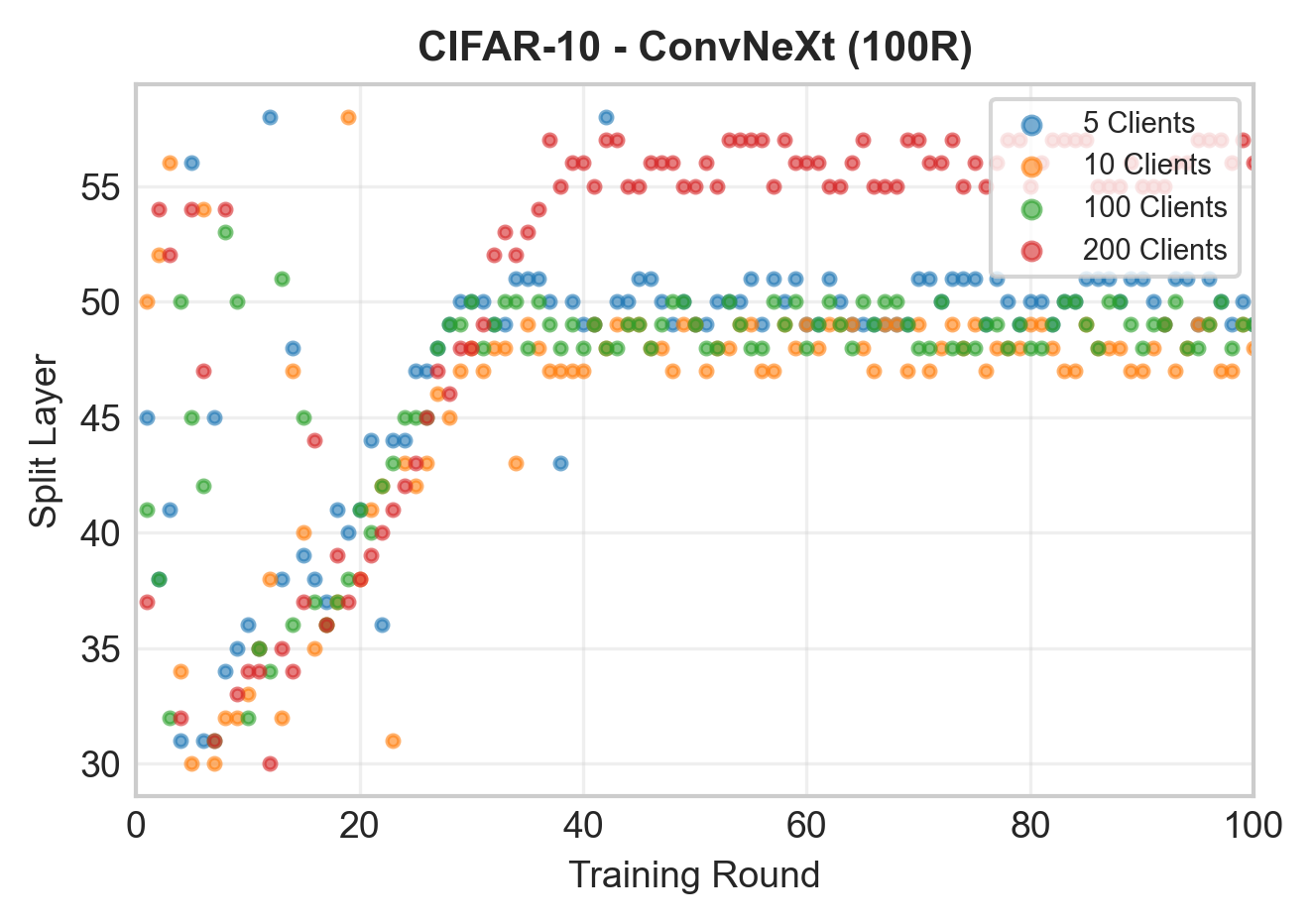}\caption{CIFAR10: ConvNeXt}
\end{subfigure}
\begin{subfigure}{0.23\textwidth}\includegraphics[width=\textwidth]{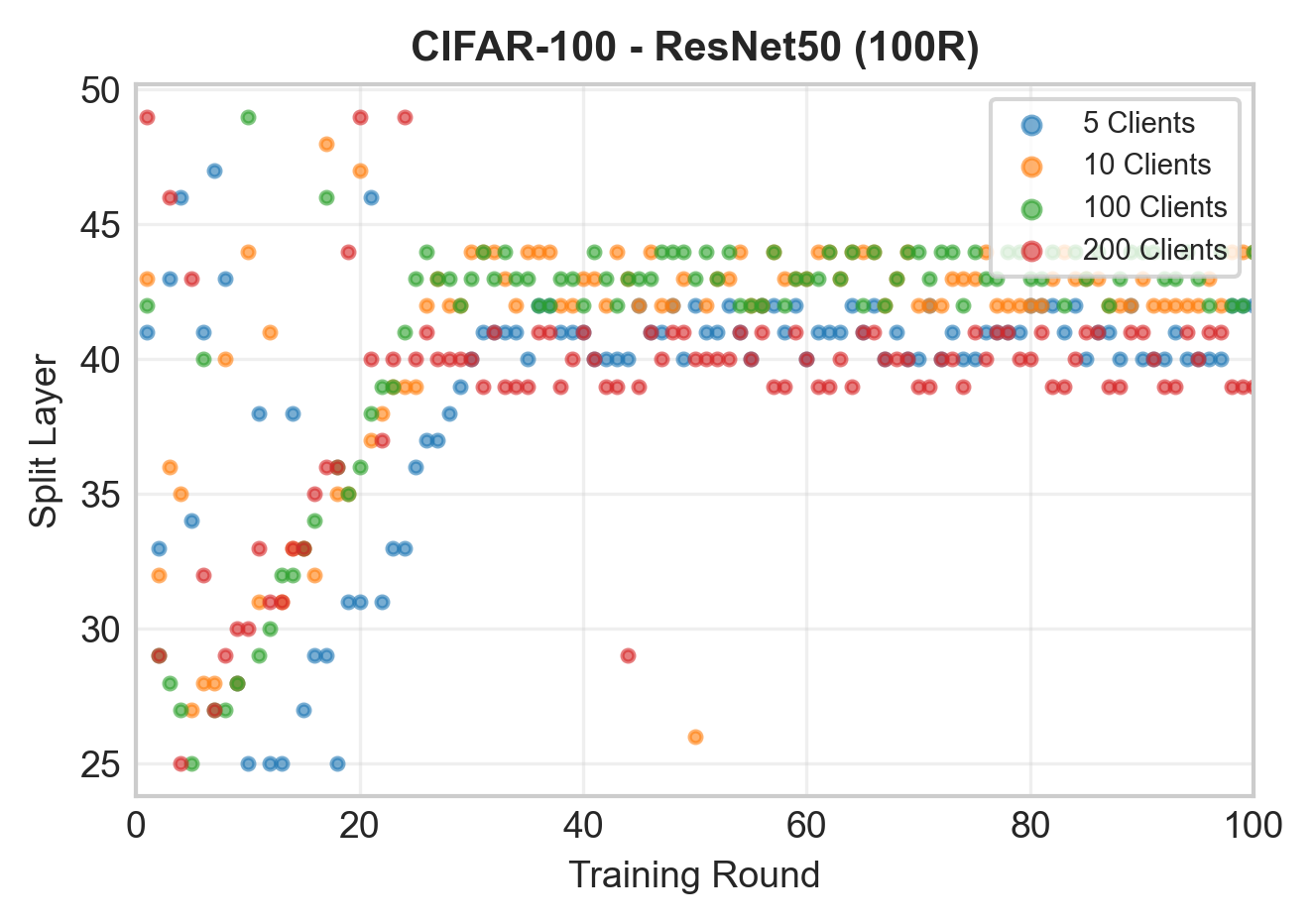}\caption{CIFAR100: ResNet50}
\end{subfigure}
\begin{subfigure}{0.23\textwidth}\includegraphics[width=\textwidth]{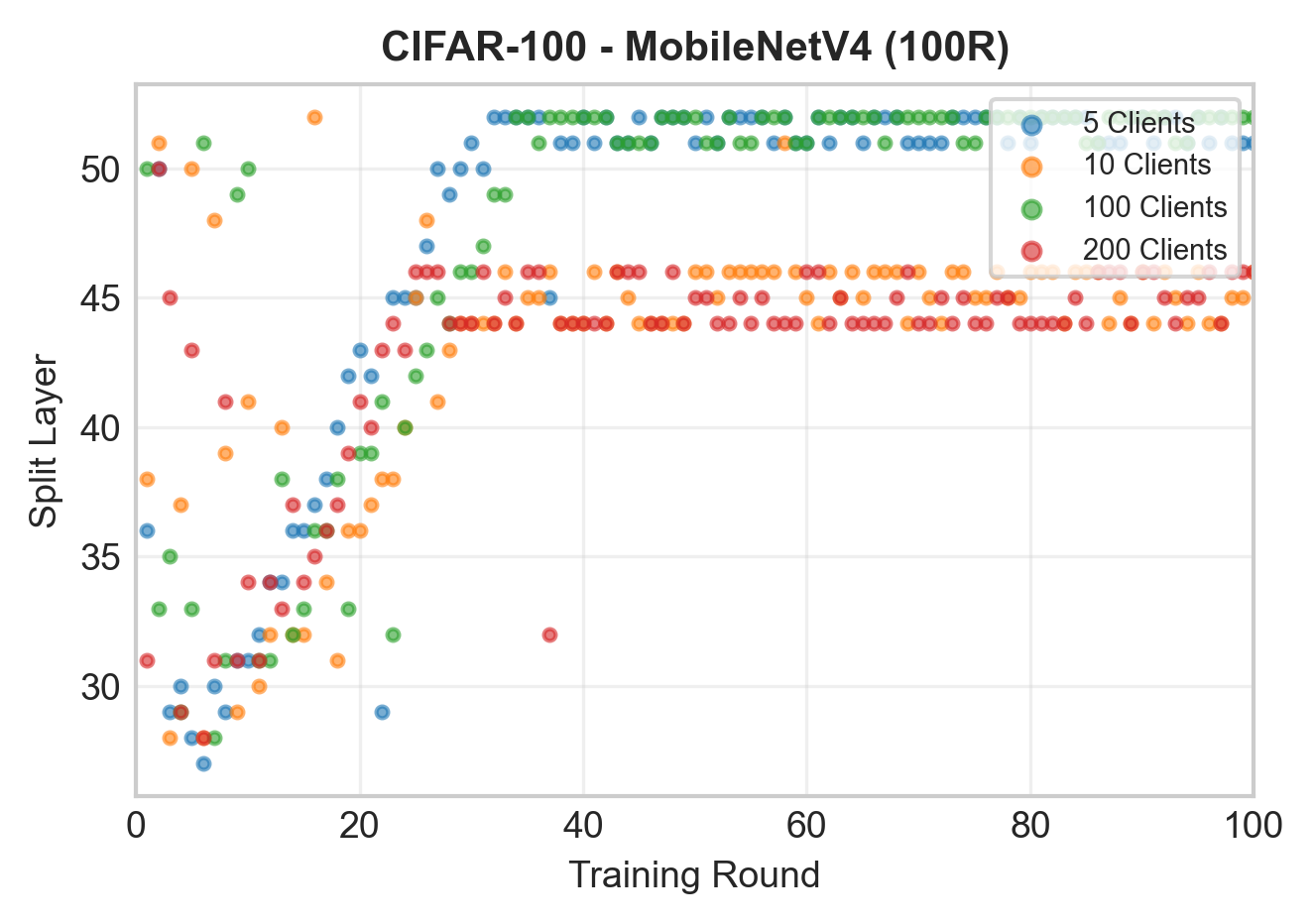}\caption{CIFAR100: MobileNetV4}
\end{subfigure}
\begin{subfigure}{0.23\textwidth}\includegraphics[width=\textwidth]{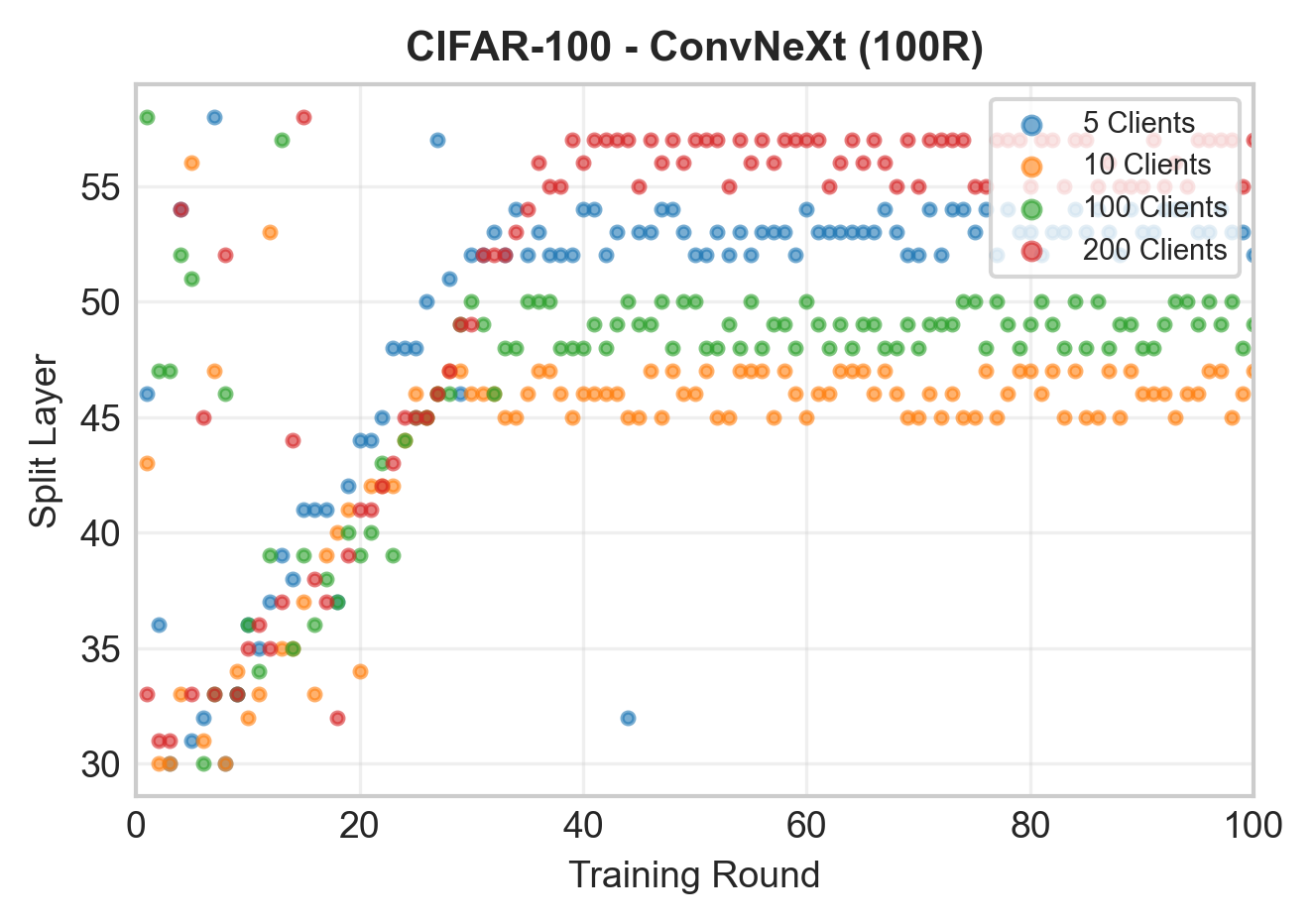}\caption{CIFAR100: ConvNeXt}
\end{subfigure}
\caption{\textbf{Dynamic Split Point Selection (100 Rounds).} Adaptive split point selection for ResNet50, MobileNetV4, and ConvNeXt. The agent consistently selects optimal split layers (30--59 depending on architecture) to balance computation and communication.}
\label{fig:split_main}
\end{minipage}
\end{figure*}

Figure~\ref{fig:split_main} shows how the RL agent adaptively selects split points for deep architectures across all datasets over 100 rounds. The agent learns to choose higher split points (layers 30--59) for deeper networks like ResNet50, MobileNetV4, and ConvNeXt. This means more layers run on the client side, which is possible because these architectures have more parameters to distribute. The split point selection remains relatively stable across different client counts. This suggests that the agent focuses primarily on the model architecture rather than the number of participating clients when making split decisions. The consistency of these choices across datasets also indicates that the learned policy generalizes well to different data distributions.

\subsection{Comparison with Baseline Methods}
We have compared the QSplitFL framework against a
few state-of-the art techniques, such as: Centralized Learning (theoretical
upper bound), FedAvg~\cite{li2019convergence}, FedProx~\cite{li2020federated},
q-FedAvg~\cite{li2019fair}, SplitFed~\cite{thapa2022splitfed},
ClusterSFL~\cite{arafeh2025efficient}, HeteroSFL~\cite{11008529},
SHeRL-FL~\cite{tran2025sherl}, and FLUID/SFL-V2~\cite{dachille2024impact}. Table~\ref{tab:baseline_comparison} presents the comparison of QSplitFL against
baseline methods using 5 clients and 100 training rounds. A detailed ablation 
study isolating the contribution of each design component is provided 
in Appendix~\ref{appendix:ablation}.

\subsection{Summary of Findings}
Across all datasets and training settings, we observe a consistent performance ranking: ConvNeXt $>$ ResNet50 $>$ MobileNetV4 $>$ CNN. This ordering holds regardless of the model architectures, number of clients, and training rounds, which ensures that deeper and more modern architectures benefit the most from QSplitFL's adaptive split point selection. Extended training with 100 rounds improves accuracy by 15--28\% over 10 rounds, with the largest gains on harder datasets like CIFAR-100. Furthermore, all configurations scale stably from 5 to 200 clients, which demonstrates QSplitFL's effectiveness in both small and large federated settings. In terms of split behavior, shallow CNNs use early 
layers (5--9), while deeper networks use mid to late layers (30--59), which enables efficient resource utilization across heterogeneous devices. Beyond accuracy, this adaptive split selection also improves communication efficiency, where deeper splits reduce the size of transmitted activations, while the capability-aware state naturally penalizes poor connectivity clients through $C_{\text{Net}}^{(k)}(t) = 1 - \text{Latency}^{(k)}(t) / 
\text{Latency}_{\max}$, pushing the agent to favor deeper splits that transmit less data. As a result, QSplitFL implicitly optimizes communication overhead without an explicit bandwidth term in the reward, making QSplitFL well suited for bandwidth-constrained deployments such as rural healthcare facilities and IoT edge environments.

\begin{table}[!t]
\centering
\caption{Baseline Accuracy Comparison (QSplitFL with 5 Clients, 100 Rounds)}
\label{tab:baseline_comparison}
\renewcommand{\arraystretch}{1.1}
\footnotesize
\resizebox{\columnwidth}{!}{%
\begin{tabular}{|l|c|c|c|c|c|c|}
\hline
\textbf{Method} & \textbf{MNIST(\%)} & \textbf{FMNIST(\%)} & \textbf{CIFAR-10(\%)} & \textbf{CIFAR-100(\%)} & \textbf{Devices} \\
\hline
Centralized & 99.8 & 94.5 & 93.8 & 76.5 & Server \\
\hline
FedAvg~\cite{li2019convergence} & 97.8 & 89.2 & 79.5 & 58.4 & High-end \\
\hline
FedProx~\cite{li2020federated} & 98.1 & 90.1 & 81.2 & 60.1 & High-end \\
\hline
q-FedAvg~\cite{li2019fair} & 97.5 & 88.5 & 78.9 & 57.8 & High-end \\
\hline
SplitFed~\cite{thapa2022splitfed} & 99.2 & 81.0 & 63.8 & -- & All \\
\hline
ClusterSFL~\cite{arafeh2025efficient}
  & -- & 89.2 & 78.5 & -- & All (IoT) \\
\hline
HeteroSFL~\cite{11008529}
  & -- & -- & 69.45 & 54.54 & Heterogeneous \\
\hline
SHeRL-FL~\cite{tran2025sherl}
  & -- & -- & 65.44 & 55.23 & All \\
\hline
FLUID/SFL-V2~\cite{dachille2024impact}
  & -- & -- & 69.45 & 52.38 & All \\
\hline
\textbf{QSplitFL} & \textbf{99.47} & \textbf{92.62} & \textbf{83.73} & \textbf{67.26} & \textbf{All} \\
\hline
\end{tabular}
}
\end{table}

\section{Conclusion}
\label{sec:conclusion}
This paper presented QSplitFL, a capability-aware Deep Q-Network framework for dynamic split point selection in SFL. The proposed approach addresses the fundamental challenge of enabling FL on resource-constrained edge devices by intelligently partitioning neural network models based on client hardware capabilities. Our framework introduces three key innovations: (1) a lightweight state representation that reduces computational complexity to $\mathcal{O}(|\mathcal{K}|)$ by directly encoding client metrics instead of model weights, (2) a committee-based RL architecture with majority voting that mitigates reward hacking and ensures adaptive split point decisions, and (3) a decayed loss-drop reward function that prioritizes early-round convergence. Experimental evaluation across four benchmark datasets and four neural network architectures demonstrates that QSplitFL achieves accuracy of 99.47\% on MNIST, 93.99\% on Fashion-MNIST, 86.16\% on CIFAR-10, and 68.27\% on CIFAR-100, while scaling gracefully from 5 to 200 clients. Finally, QSplitFL outperforms traditional FL methods while enabling participation of resource-constrained devices.

\begin{credits}
\subsubsection{\ackname} The work of Nazmus Shakib Shadin and Xinyue Zhang is partly supported by the U.S. National Science Foundation (NSF-2348417 and NSF-2431597). The work of Jingyi Wang is partly supported by the U.S. National Science Foundation (CNS-2431594).
\end{credits}

\bibliographystyle{splncs04}
\bibliography{references}


\clearpage
\appendix

\section{Appendix}
\label{sec:appendix}

This appendix provides detailed supplementary materials that correspond to the main paper. We begin with a discussion of the key challenges in existing split learning based federated learning (SFL) and our proposed solutions, followed by the committee-based reinforcement learning (RL) mechanism, neural network architectures, notation summary, reward function, supporting algorithm pseudocode, detailed experimental setup, and additional experimental results.

\subsection{Challenges in Split Learning based Federated Learning and Our Solution}
\label{sec:challenges_solution}
\begin{figure}[!ht]
\centering
\includegraphics[width=\columnwidth]{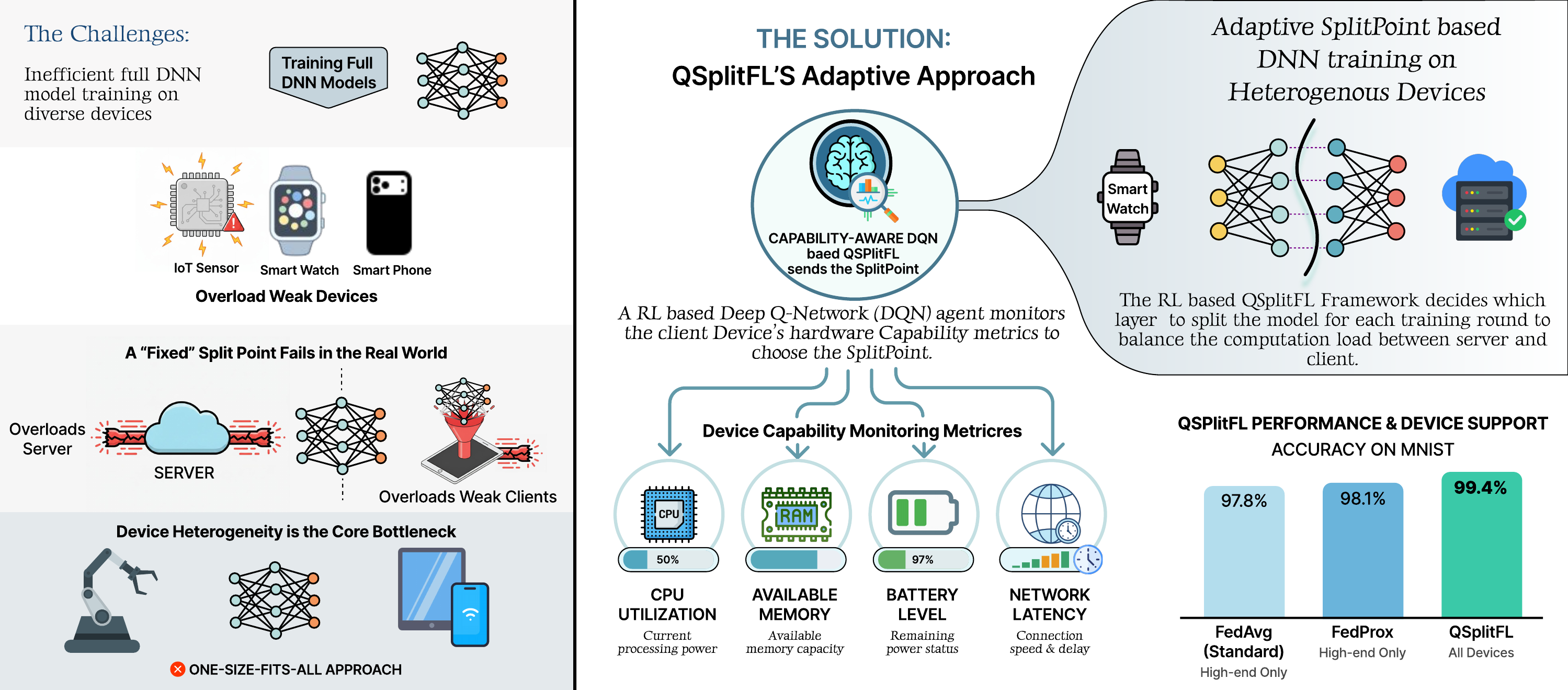}
\caption{\textbf{Challenges in Existing Split Federated Learning vs Our Proposed Solution.} The left side illustrates the key limitations of current state-of-the-art approaches, including static split points, inability to handle device heterogeneity. The right side presents QSplitFL's capability-aware SFL based reinforcement learning solution that dynamically adapts split points based on current client resource metrics.}
\label{fig:challenge_solution}
\end{figure}

A fundamental challenge in SFL is to determine the optimal split point between client and server portions of the deep neural network (DNN). As illustrated in Figure~\ref{fig:challenge_solution}, existing state-of-the-art approaches face several critical limitations when addressing device heterogeneity. The key challenges in existing approaches are:
\begin{itemize}
    \item \textbf{Static Split Point Selection:} Traditional methods use fixed split points that fail to adapt to varying client capabilities, leading to suboptimal resource utilization and training inefficiency.
    \item \textbf{Device Heterogeneity:} Clients possess diverse computational resources (CPU, memory, battery, network bandwidth), which makes a one-size-fits-all approach inadequate.
    \item \textbf{Resource Fluctuations:} Client capabilities can change over time due to concurrent processes, battery drain, and network conditions, which require adaptive split points in SFL environments.
\end{itemize}

\textbf{Our Solution:} QSplitFL addresses these challenges through a capability-aware reinforcement learning framework that adaptively selects optimal split points based on current client resource measurements. By formulating split point selection as a Markov Decision Process (MDP) and employing a committee based Deep Q-Networks (DQN), our approach achieves robust, adaptive decisions that maximize training efficiency by considering device constraints.

\subsection{Capability Metric Descriptions}
\label{appendix:metrics}
The four normalized capability metrics are defined as follows:
(1)~$C_{\text{CPU}}^{(k)}$ measures available CPU resources normalized
by maximum CPU capacity; (2)~$C_{\text{Mem}}^{(k)}$ represents available
memory normalized by maximum memory; (3)~$C_{\text{Bat}}^{(k)}$ indicates
the current battery level as a fraction of full charge; and
(4)~$C_{\text{Net}}^{(k)}$ inversely measures network quality such that
lower latency yields higher capability.

\subsection{Committee-Based Reward Hacking Mitigation}
\label{sec:reward_hacking}

A critical consideration in RL based optimization is the potential for reward hacking, which means that the agent learns to find loopholes in the reward function rather than achieving the intended objective. In the context of SFL, a single DQN agent might learn suboptimal behaviors that artificially inflate rewards without genuinely improving model performance.

\begin{figure}[!ht]
\centering
\includegraphics[width=\columnwidth]{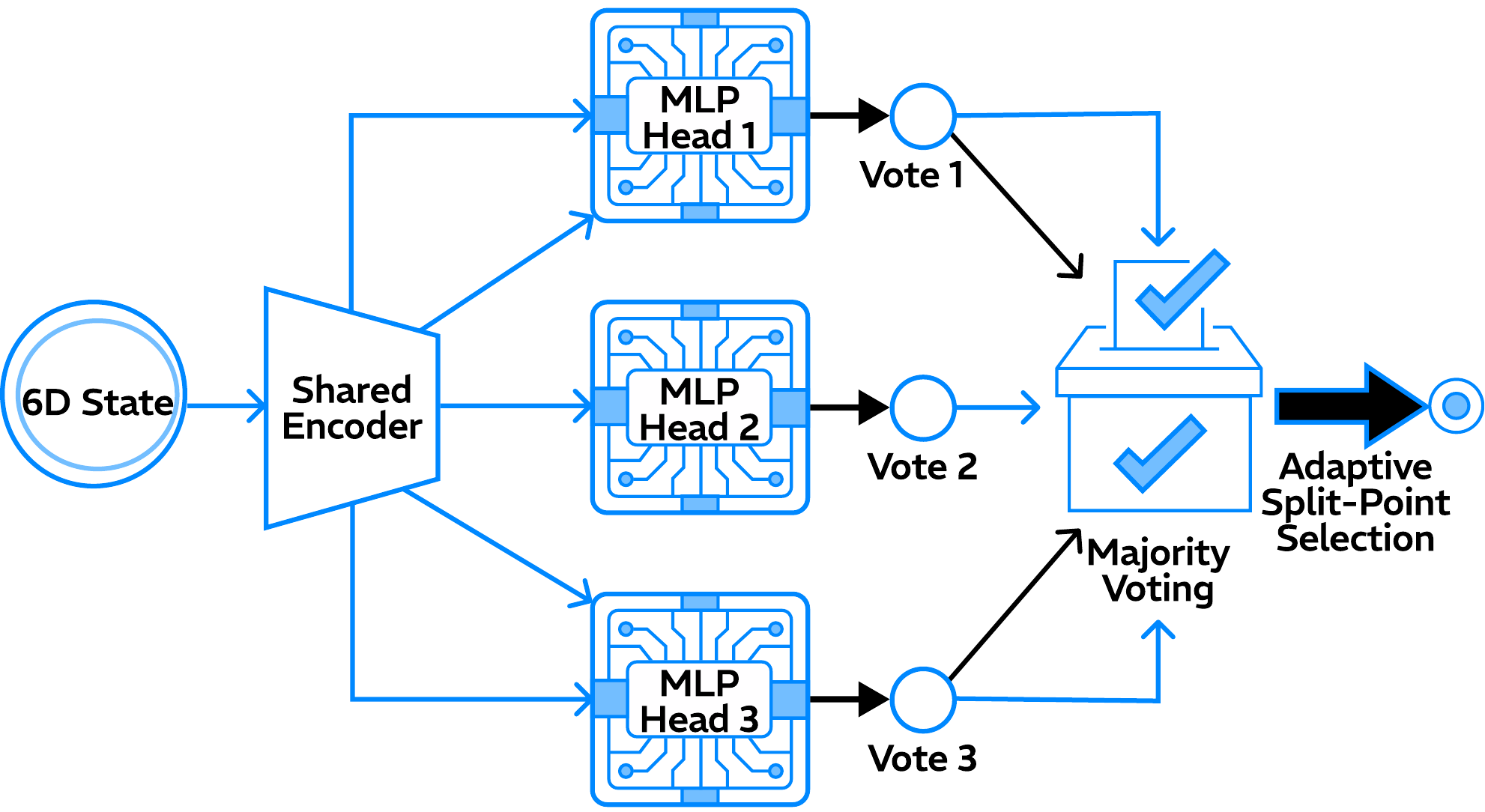}
\caption{\textbf{Committee-Based Reward Hacking Prevention Mechanism.} The architecture employs $M$ DQN members (typically $M=3$ or $M=5$, always odd) which shares a common encoder but maintains independent decision heads. Each member proposes its preferred split action, and the final decision is made through majority voting. This ensemble approach mitigates reward hacking by ensuring decisions reflect consensus across diverse learned policies rather than exploitation by a single agent.}
\label{fig:reward_hacking_committee}
\end{figure}

Figure~\ref{fig:reward_hacking_committee} illustrates how QSplitFL's committee based architecture mitigates reward hacking through ensemble decision making. By employing multiple independent DQN heads with a shared encoder backbone, the framework introduces diverse perspectives in action selection: The mechanisms are given below:
\begin{itemize}
    \item \textbf{Shared Feature Extraction:} All committee members share a common encoder $f_s(\cdot; \phi_s)$ that extracts capability which are relevant features from the state representation.
    \item \textbf{Independent Decision Heads:} Each member $m$ maintains its own decision head $g^{(m)}(\cdot; \psi^{(m)})$, which is potentially different Q-value estimates and action preferences. This approach prevents any single network from dominating decisions based on different patterns in the training data.
    \item \textbf{Majority Voting:} The final split point is determined by majority voting across all committee members.
    \item \textbf{Tie-Breaking via Mean Q-Values:} When votes are tied, the action with the highest average Q-value across all members is selected.
\end{itemize}

Having established the conceptual foundations through the above figures, we now present the detailed technical specifications including DNNs, mathematical notation, reward function behavior, complete algorithm pseudocode, and comprehensive experimental results.

\subsection{Neural Network Architectures}
\label{sec:architectures}
Table~\ref{tab:model_architectures} summarizes the neural network architectures which are used in QSplitFL and their split point configurations.

\begin{table}[!ht]
\centering
\caption{Neural Network Architectures and Split Point Selection Action}
\label{tab:model_architectures}
\renewcommand{\arraystretch}{1.2}
\resizebox{\columnwidth}{!}{%
\begin{tabular}{|l|c|c|c|c|}
\hline
\textbf{Architecture} & \textbf{Total Layers} & \textbf{Split Min} & \textbf{Split Max} & \textbf{Example Action Space} \\
\hline
CNN & 10 & 5 & 9 & 7 \\
\hline
ResNet50 & 50 & 25 & 49 & 33 \\
\hline
MobileNetV4 & 53 & 27 & 52 & 39 \\
\hline
ConvNeXt & 59 & 30 & 58 & 41 \\
\hline
\end{tabular}%
}
\end{table}
\subsection{Notation Summary}
\label{sec:notation}
Table~\ref{tab:notation} provides a comprehensive reference for the mathematical notation used throughout this paper.
\begin{table}[!htb]
\centering
\caption{Summary of Notation}
\label{tab:notation}
\renewcommand{\arraystretch}{0.95}
\scriptsize
\begin{tabular}{|c|p{5.2cm}|}
\hline
\textbf{Symbol} & \textbf{Description} \\
\hline
$K$ & Total number of clients in the federated network \\
\hline
$C$ & Number of client clusters \\
\hline
$\mathcal{K}_c$ & Set of clients in cluster $c$ \\
\hline
$E$ & Number of Episodes \\
\hline
$T$ & Total number of training rounds per episode \\
\hline
$t$ & Current training round index, $t \in \{1, \ldots, T\}$ \\
\hline
$\mathcal{C}$ & Number of rounds between target network updates\\
\hline
$s_t^{(c)}$/$s$/$S$ & State or Cluster state at round $t$ for cluster $c$ \\
\hline
$\mathcal{A}$/$a$ & Action space: $\{\lceil L/2 \rceil, \ldots, L-1\}$ \\
\hline
$r_t$/$R$ & Reward signal at round $t$ \\
\hline
$L$ & Total number of layers in the neural network \\
\hline
$\ell$ & Split layer index (action), $\ell \in \mathcal{A}$ \\
\hline
$C_i^{(k)}(t)$ & Capability metric $i$ for client $k$ at round $t$ \\
\hline
$\bar{C}_i^{(c)}(t)$ & Mean capability metric $i$ for cluster $c$ \\
\hline
$\sigma_c(t)$ & Capability heterogeneity (std. dev.) for cluster $c$ \\
\hline
$L_t$ & Cluster-aggregated validation loss at round $t$ \\
\hline
$L_{t-1}$ & Cluster-aggregated validation loss at round $t-1$ \\
\hline
$L_t^{(k)}$ & Per-client validation loss for client $k$ \\
\hline
$\rho_t$ & Temporal decay factor at round $t$ \\
\hline
$\lambda$ & Decay rate constant \\
\hline
$Q_\theta(s, a)$ & Q-function parameterized by $\theta$ \\
\hline
$Q_{\bar{\theta}}$ & Target Q-network with parameters $\bar{\theta}$ \\
\hline
$\mathcal{D}$ & Experience replay buffer \\
\hline
$N_{\max}$ & Maximum replay buffer capacity \\
\hline
$B$ & Mini-batch size for training \\
\hline
$\eta$ & Learning rate \\
\hline
$\alpha$ & Uniform distribution for non-iid data \\
\hline
$\gamma$ & Discount factor, typically $\gamma\in[0.1,0.3]$ \\
\hline
$\pi$ & Policy of Q-learning with $\epsilon$-greedy exploration \\
\hline
$\epsilon$ & Exploration rate for $\epsilon$-greedy policy \\
\hline
w.p. & With probability \\
\hline
$M$ & Number of DQN committee members \\
\hline
\end{tabular}
\end{table}

\subsection{Reward Function Behavior}
\label{sec:reward_table}
Table~\ref{tab:reward_scenarios} illustrates the reward behavior across representative training scenarios.

\begin{table*}[!htbp]
\centering
\caption{Reward Function Behavior Across Training Scenarios}
\label{tab:reward_scenarios}
\renewcommand{\arraystretch}{1.2}
\setlength{\tabcolsep}{4pt}
\resizebox{\textwidth}{!}{%
\begin{tabular}{|p{2.8cm}|c|c|c|c|p{8.0cm}|}
\hline
\textbf{Scenario} & \textbf{$t$} & \textbf{$L_{t-1} \to L_t$} & \textbf{$\rho_t$} & \textbf{$r_t$} & \textbf{Interpretation} \\
\hline
Early large improvement & 2 & $0.80 \to 0.68$ & 0.707 & $+0.085$ & Large early drop ($\Delta L_t=-0.12$) posts a strong positive reward. \\
\hline
Early modest improvement & 2 & $0.72 \to 0.70$ & 0.707 & $+0.014$ & Small early gain posts a positive reward with smaller magnitude. \\
\hline
Early no change & 2 & $0.70 \to 0.70$ & 0.707 & $0.000$ & No change in loss posts near zero reward. \\
\hline
Early degradation & 2 & $0.70 \to 0.74$ & 0.707 & $-0.028$ & Loss increase posts a negative reward, penalizing the selected split. \\
\hline
Mid episode improvement & 8 & $0.62 \to 0.58$ & 0.545 & $+0.022$ & Improvement receives less credit due to exponential decay. \\
\hline
Late small improvement & 15 & $0.55 \to 0.54$ & 0.523 & $+0.005$ & Late and small gain is strongly down weighted by $\rho_t$. \\
\hline
Late degradation & 15 & $0.54 \to 0.57$ & 0.523 & $-0.016$ & Late regression posts a negative reward, with moderated magnitude due to decay. \\
\hline
\end{tabular}%
}
\end{table*}

\subsection{Supporting Algorithms}
\label{sec:algorithms}
This section presents the supporting algorithmic procedures for the QSplitFL framework. The main training algorithm (Algorithm~\ref{alg:committee_dqn}) is presented in the main paper.

\subsubsection{Algorithm: SFL Training Round}
Algorithm~\ref{alg:train_round} formalizes the execution of a SFL round, corresponding to Steps~5--12 in Figure~\ref{fig:qsplitfl_architecture}. This algorithm is invoked by Algorithm~\ref{alg:committee_dqn} whenever a training round must be executed with a selected split point. The procedure begins with broadcasting model parameters and split configuration to all clients (Step~5). Clients execute parallel forward passes through their assigned layers $1$ to $\ell$ (Steps~6--7), transmitting smashed data $A_{k,t}$ and true labels $Y_k$ to the server. The server completes forward propagation through layers $\ell+1$ to $L$ (Step~8), computes per-client losses (Step~8b), performs backpropagation (Step~9), sends gradients back for client-side backpropagation (Steps~10--11), and aggregates client updates via FedAvg (Step~12).

\begin{algorithm}[!ht]
\caption{SFL Training Round: $\texttt{TrainRoundSFL}(c, s_t^{(c)}, W_{t-1}, \ell)$}
\label{alg:train_round}
\begin{algorithmic}[1]
\renewcommand{\algorithmicrequire}{\textbf{Input:}}
\renewcommand{\algorithmicensure}{\textbf{Output:}}
\REQUIRE Cluster $c$; current state $s_t^{(c)}$; previous model weights $W_{t-1}$; selected split layer $\ell$
\ENSURE Updated weights $W_t$; cluster loss $L_t$; per-client losses $\{L_t^{(k)}\}$
\STATE Broadcast $(W_{t-1}, \ell)$ to all clients $k \in \mathcal{K}_c$
\FOR{each client $k \in \mathcal{K}_c$ \textbf{in parallel}}
    \STATE \textbf{Client-side forward:} Compute $A_{k,t} = f_{\text{client}}(X_k; W_{t-1}^{[1:\ell]})$ \COMMENT{Smashed data}
    \STATE Send $(A_{k,t}, Y_k)$ to server \COMMENT{Activations and true labels}
\ENDFOR
\STATE \textbf{Server-side forward:} For each client $k$, compute $\hat{Y}_k = f_{\text{server}}(A_{k,t}; W_{t-1}^{[\ell+1:L]})$
\STATE \textbf{Compute losses:} $L_t^{(k)} = \mathcal{L}(\hat{Y}_k, Y_k)$ for each $k$; $L_t = \sum_{k} \omega_k L_t^{(k)}$ where $\omega_k = n_k / \sum_j n_j$
\STATE \textbf{Server backpropagation:} Compute $\nabla_{W^{[\ell+1:L]}} L_t$ and gradients $\nabla A_{k,t}$
\STATE Send $\nabla A_{k,t}$ to respective clients
\FOR{each client $k \in \mathcal{K}_c$ \textbf{in parallel}}
    \STATE \textbf{Client-side backpropagation:} Compute $\nabla_{W^{[1:\ell]}} L_t^{(k)}$ using received $\nabla A_{k,t}$
    \STATE Send client-side gradients to server
\ENDFOR
\STATE \textbf{FedAvg aggregation:} $W_t = W_{t-1} - \eta_{\text{model}} \cdot \frac{1}{|\mathcal{K}_c|} \sum_{k \in \mathcal{K}_c} \nabla W_k$
\RETURN $W_t$, $L_t$, $\{L_t^{(k)}\}$
\end{algorithmic}
\end{algorithm}

\subsubsection{Algorithm: Committee Majority Voting}
Algorithm~\ref{alg:committee_vote} implements the majority voting mechanism (Step~4 in Figure~\ref{fig:qsplitfl_architecture}) for robust action selection when using a committee of $M$ models (where $M$ is odd to avoid ties). Each committee member $Q_\theta^{(m)}$ independently proposes its preferred split point. Vote counts are tallied for each candidate action, and the action receiving the most votes is selected. In case of a tie, the algorithm selects the action with the highest mean Q-value across all committee members.

\begin{algorithm}[!ht]
\caption{Committee Majority Voting: $\texttt{CommitteeVote}(s, \{Q_\theta^{(m)}\}_{m=1}^{M})$}
\label{alg:committee_vote}
\begin{algorithmic}[1]
\renewcommand{\algorithmicrequire}{\textbf{Input:}}
\renewcommand{\algorithmicensure}{\textbf{Output:}}
\REQUIRE Current state $s$; action set $\mathcal{A} = \{\lceil L/2 \rceil, \ldots, L-1\}$; committee of Q-networks $\{Q_\theta^{(m)}\}_{m=1}^{M}$
\ENSURE Selected split action $a^* \in \mathcal{A}$
\STATE \textbf{Member proposals:} For each committee member $m = 1, \ldots, M$:
\[
\tilde{a}^{(m)} \leftarrow \arg\max_{a \in \mathcal{A}} Q_\theta^{(m)}(s, a) \quad \text{(each member's action)}
\]
\STATE \textbf{Vote counting:} For each action $a \in \mathcal{A}$, compute vote count:
\[
v(a) \leftarrow \left|\{m \in \{1, \ldots, M\} : \tilde{a}^{(m)} = a\}\right|
\]
\STATE \textbf{Identify majority winner(s):} $\mathcal{A}_{\max} \leftarrow \arg\max_{a \in \mathcal{A}} v(a)$
\IF{$|\mathcal{A}_{\max}| = 1$}
    \RETURN $a^* \leftarrow$ the unique action in $\mathcal{A}_{\max}$
\ELSE
    \STATE \textbf{Tie-breaking via mean Q-value:} For each $a \in \mathcal{A}_{\max}$:
    \[
    \bar{Q}(s, a) = \frac{1}{M}\sum_{m=1}^{M} Q_\theta^{(m)}(s, a)
    \]
    \RETURN $a^* \leftarrow \arg\max_{a \in \mathcal{A}_{\max}} \bar{Q}(s, a)$
\ENDIF
\end{algorithmic}
\end{algorithm}

\subsection{Detailed Experimental Setup}
\label{appendix:setup}
This subsection provides the implementation details needed to reproduce our experiments, including how heterogeneous and resource-constrained devices are modeled, how client capabilities are generated, how clients are clustered, and the full hyperparameter and baseline tuning settings.

\subsubsection{Device Heterogeneity Simulation and Capability Generation.}
\label{appendix:heterogeneity}
We model device heterogeneity by assigning each client a hardware profile drawn from one of three device tiers, namely strong, medium, and weak, which reflect the spread of hardware found in deployments such as rural healthcare networks. For each client $k$, the four raw metrics, which are available CPU fraction, available memory fraction, battery level, and network latency, are sampled from tier-specific ranges. Representative ranges are listed in Table~\ref{tab:device_tiers}. At every round, each metric is perturbed by bounded random fluctuation around its tier value to emulate battery drain, competing background processes, and network congestion, so that a client's capability changes over time rather than staying fixed. The raw values are then normalized to $[0,1]$ using min-max normalization across the network, as defined in Section~\ref{sec:qsplitfl_framework}, which produces the per-client capability metrics $C_i^{(k)}(t)$.

\begin{table}[!ht]
\centering
\caption{Representative Capability Ranges for Simulated Device Tiers}
\label{tab:device_tiers}
\renewcommand{\arraystretch}{1.2}
\resizebox{\columnwidth}{!}{%
\begin{tabular}{|l|c|c|c|c|}
\hline
\textbf{Device Tier} & \textbf{CPU Avail.} & \textbf{Memory Avail.} & \textbf{Battery} & \textbf{Latency (ms)} \\
\hline
Strong & 0.70--1.00 & 0.70--1.00 & 0.60--1.00 & 10--40 \\
\hline
Medium & 0.40--0.70 & 0.40--0.70 & 0.40--0.80 & 40--100 \\
\hline
Weak   & 0.10--0.40 & 0.10--0.40 & 0.10--0.60 & 100--250 \\
\hline
\end{tabular}%
}
\end{table}

\subsubsection{Clustering Strategy.}
Clients are grouped into $C$ capability clusters according to their overall capability score $C_{\text{Overall}}^{(k)}(t)$. Clients with similar scores are placed in the same cluster, which keeps the within-cluster heterogeneity $\sigma_c(t)$ low and lets a single split point serve all members of a cluster without overloading the weakest device. Each cluster is controlled by its own RL agent, which observes the six-dimensional cluster state and selects one split layer per round. This design separates strong and weak devices, so that capable clusters can use deeper splits while constrained clusters use shallower ones.

\subsubsection{Hyperparameter Settings.}
\label{appendix:hyperparameters}
Table~\ref{tab:hyperparameters} lists the hyperparameters for both the SFL training loop and the DQN controller. These values are used in all experiments unless stated otherwise.

\begin{table}[!ht]
\centering
\caption{Hyperparameter Settings for SFL Training and the DQN Controller}
\label{tab:hyperparameters}
\renewcommand{\arraystretch}{1.15}
\resizebox{\columnwidth}{!}{%
\begin{tabular}{|l|l|}
\hline
\textbf{Component / Hyperparameter} & \textbf{Value} \\
\hline
\multicolumn{2}{|l|}{\textbf{SFL Training}} \\
\hline
Client optimizer & SGD (momentum $0.9$) \\
\hline
Model learning rate $\eta_{\text{model}}$ & $0.01$ \\
\hline
Local epochs per round & $1$ \\
\hline
Mini-batch size (model) & $64$ \\
\hline
Aggregation & FedAvg \\
\hline
Training rounds $T$ & up to $100$ \\
\hline
Data partition & Dirichlet ($\alpha = 0.5$) \\
\hline
\multicolumn{2}{|l|}{\textbf{DQN Controller}} \\
\hline
State dimension & $6$ \\
\hline
Hidden layers (shared encoder + head) & $2 \times 64$, ReLU \\
\hline
Committee size $M$ & $3$ (odd) \\
\hline
Discount factor $\gamma$ & $0.2$ \\
\hline
Controller learning rate $\eta$ & $1\times10^{-3}$ \\
\hline
Replay buffer capacity $N_{\max}$ & $10{,}000$ \\
\hline
Mini-batch size $B$ & $32$ \\
\hline
Target sync interval $\mathcal{C}$ & $10$ rounds \\
\hline
Exploration $\epsilon_0 \to \epsilon_{\min}$ & $1.0 \to 0.05$ \\
\hline
Exploration decay $\kappa$ & $0.05$ \\
\hline
Reward decay rate $\lambda$ & $0.05$ \\
\hline
Capability weights $(w_{\text{CPU}}, w_{\text{Mem}}, w_{\text{Bat}}, w_{\text{Net}})$ & $(0.30, 0.30, 0.20, 0.20)$ \\
\hline
\end{tabular}%
}
\end{table}

\subsubsection{Baseline Tuning and Fairness.}
\label{appendix:baseline_tuning}
For a fair comparison, every baseline shares the same backbone architecture, non-IID data partition (Dirichlet $\alpha = 0.5$), client count, optimizer, and number of training rounds as QSplitFL. FedAvg, FedProx, and q-FedAvg are full-model federated baselines: the FedProx proximal weight $\mu$ and the q-FedAvg fairness parameter $q$ were each swept over standard ranges and the best configuration is reported. The split-based baselines, namely SplitFed, ClusterSFL, HeteroSFL, SHeRL-FL, and FLUID/SFL-V2, use the same client and server partition boundaries as QSplitFL where applicable, so that accuracy differences reflect the split-point policy rather than differing model capacity. QSplitFL uses no extra training data or additional rounds relative to the baselines.

\subsection{Additional Experimental Results}
\label{sec:additional_results}

We present detailed experimental results demonstrating the effectiveness of the QSplitFL framework.

\textbf{Federated Configurations:} We have conducted the experiments across four client configurations (5, 10, 100, 200 clients) with data distributed using Dirichlet distribution~\cite{beutel2020flower} ($\alpha = 0.5$) to simulate non-IID settings for data heterogeneity. We have evaluated QSplitFL at training round checkpoints of 10, 20, 50, and 100 rounds.

\begin{figure}[!ht]
\centering
\begin{subfigure}{0.98\linewidth}\includegraphics[width=\textwidth]{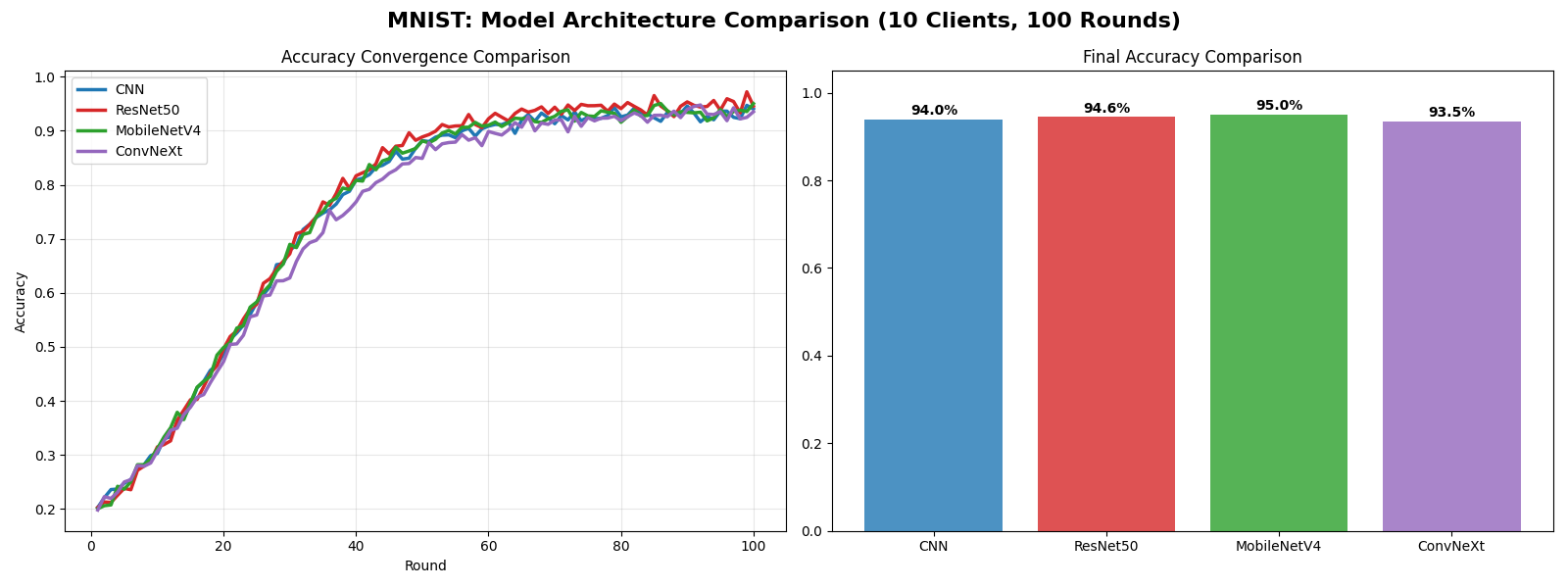}\caption{MNIST}\end{subfigure}
\begin{subfigure}{0.98\linewidth}\includegraphics[width=\textwidth]{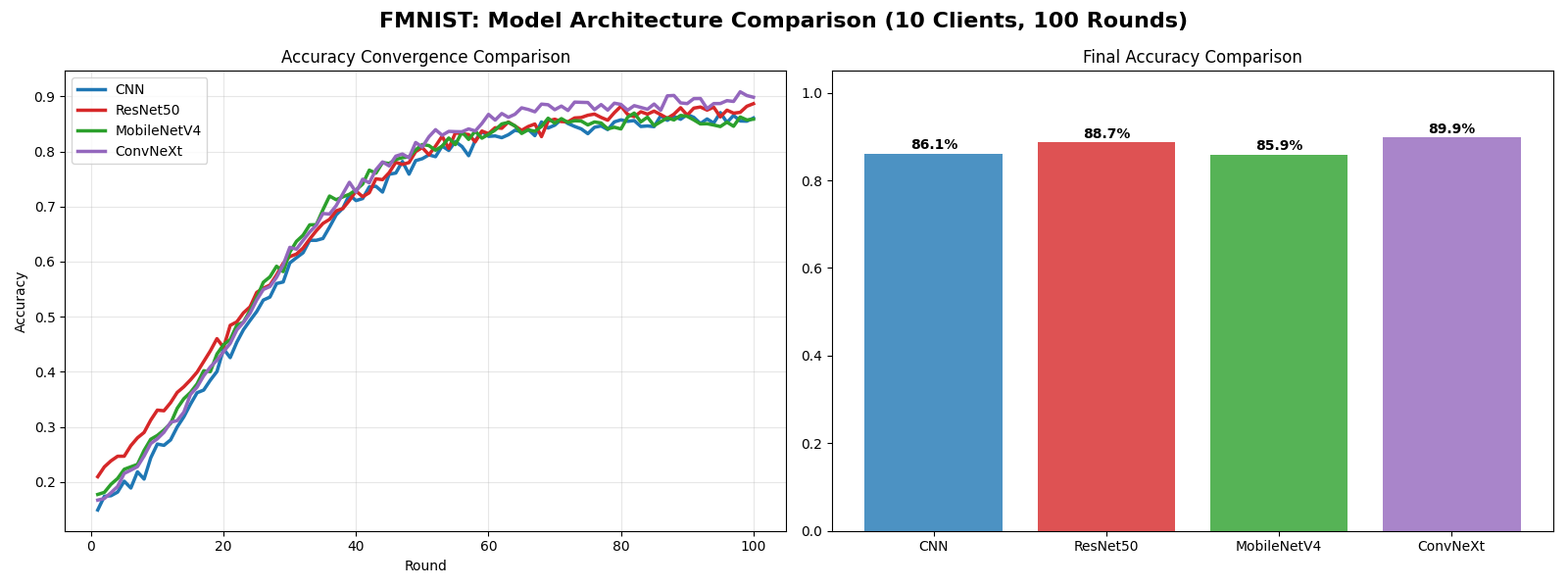}\caption{Fashion-MNIST}\end{subfigure}
\begin{subfigure}{0.98\linewidth}\includegraphics[width=\textwidth]{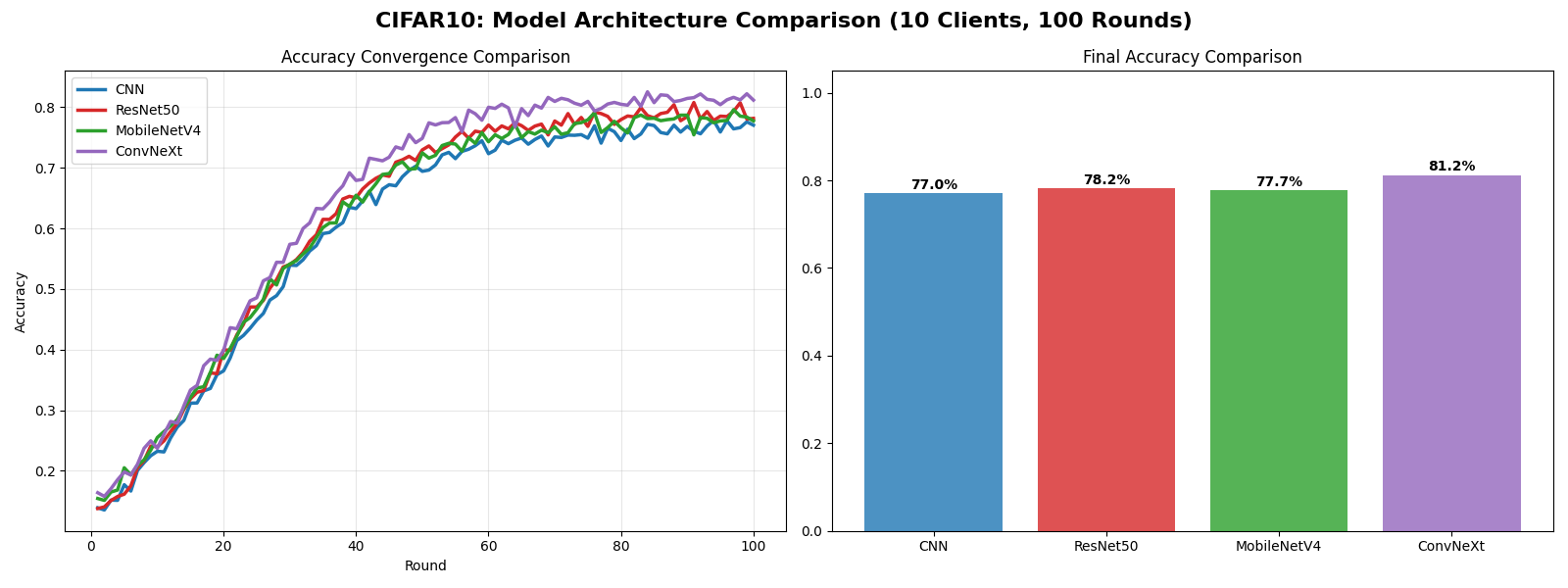}\caption{CIFAR-10}\end{subfigure}
\begin{subfigure}{0.98\linewidth}\includegraphics[width=\textwidth]{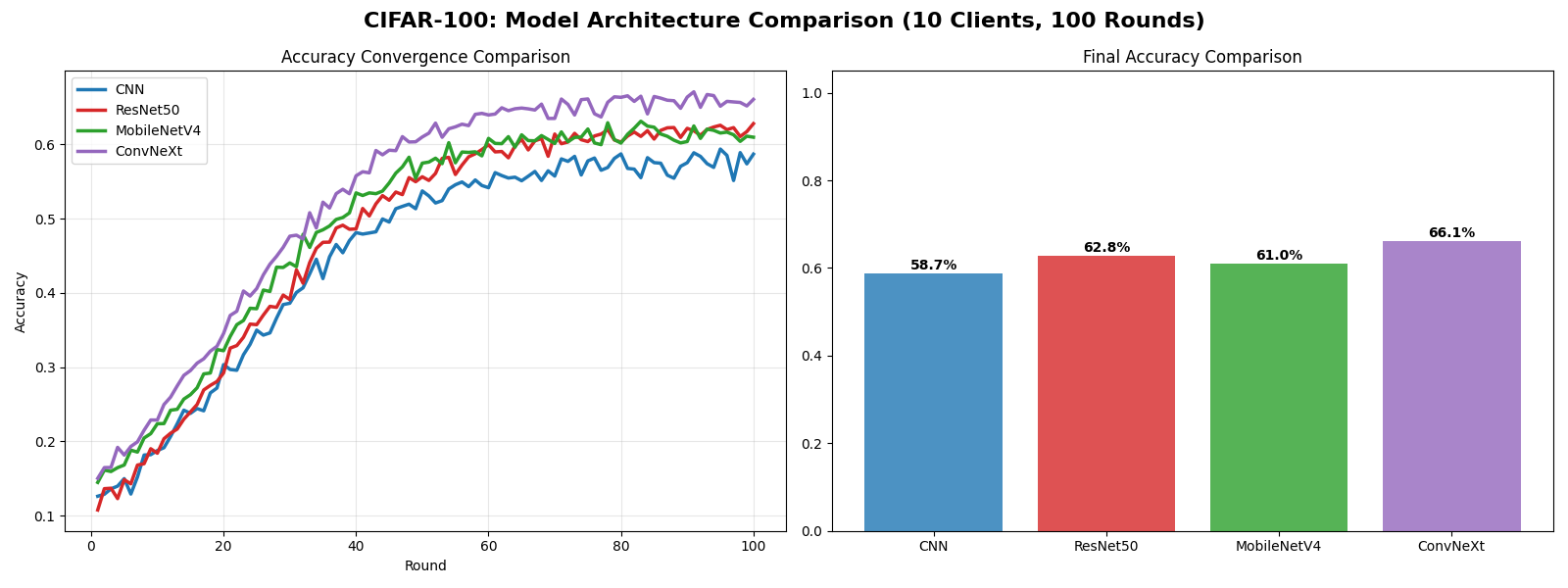}\caption{CIFAR-100}\end{subfigure}
\caption{Comprehensive Model Architecture Performance Comparison Across All Benchmark Datasets.}
\label{fig:model_comparison}
\end{figure}

\textbf{Model Architecture Comparison:}
Figure~\ref{fig:model_comparison} presents a detailed comparison of accuracy achieved by all four neural network architectures (CNN with 10 layers, ResNet50 with 50 layers, MobileNetV4 with 53 layers, and ConvNeXt with 59 layers) evaluated in the QSplitFL framework across all four benchmark datasets. Each subplot displays accuracy for different client configurations (5, 10, 100, 200 clients) after 100 training rounds. The results consistently demonstrate that deeper architectures significantly outperform shallower models: ConvNeXt achieves the highest accuracy across all datasets (99.6\% on MNIST, 94.1\% on Fashion-MNIST, 86.5\% on CIFAR-10, 68.3\% on CIFAR-100), followed by ResNet50, then MobileNetV4, with CNN performing lowest. The performance gap between architectures becomes increasingly high as dataset complexity increases. On MNIST dataset, all architectures achieve almost similar accuracy ($>99\%$), while on CIFAR-100, ConvNeXt outperforms CNN by 5--8\% . This demonstrates that the QSplitFL capability aware split point selection enables effective training of deep neural networks on heterogeneous edge device devices.

\begin{figure}[!ht]
\centering
\begin{subfigure}{0.48\columnwidth}\includegraphics[width=\textwidth]{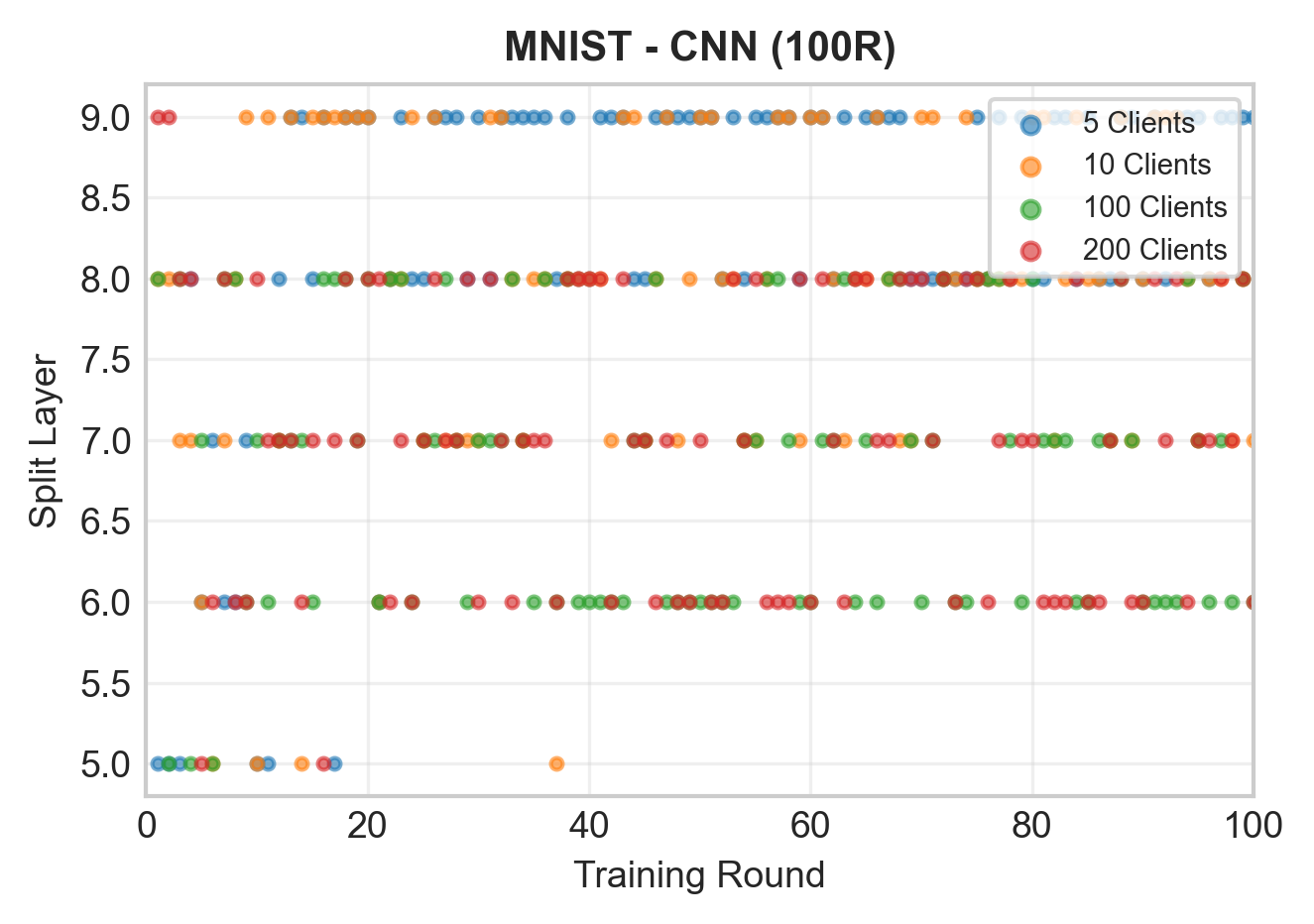}\caption{MNIST}\end{subfigure}
\begin{subfigure}{0.48\columnwidth}\includegraphics[width=\textwidth]{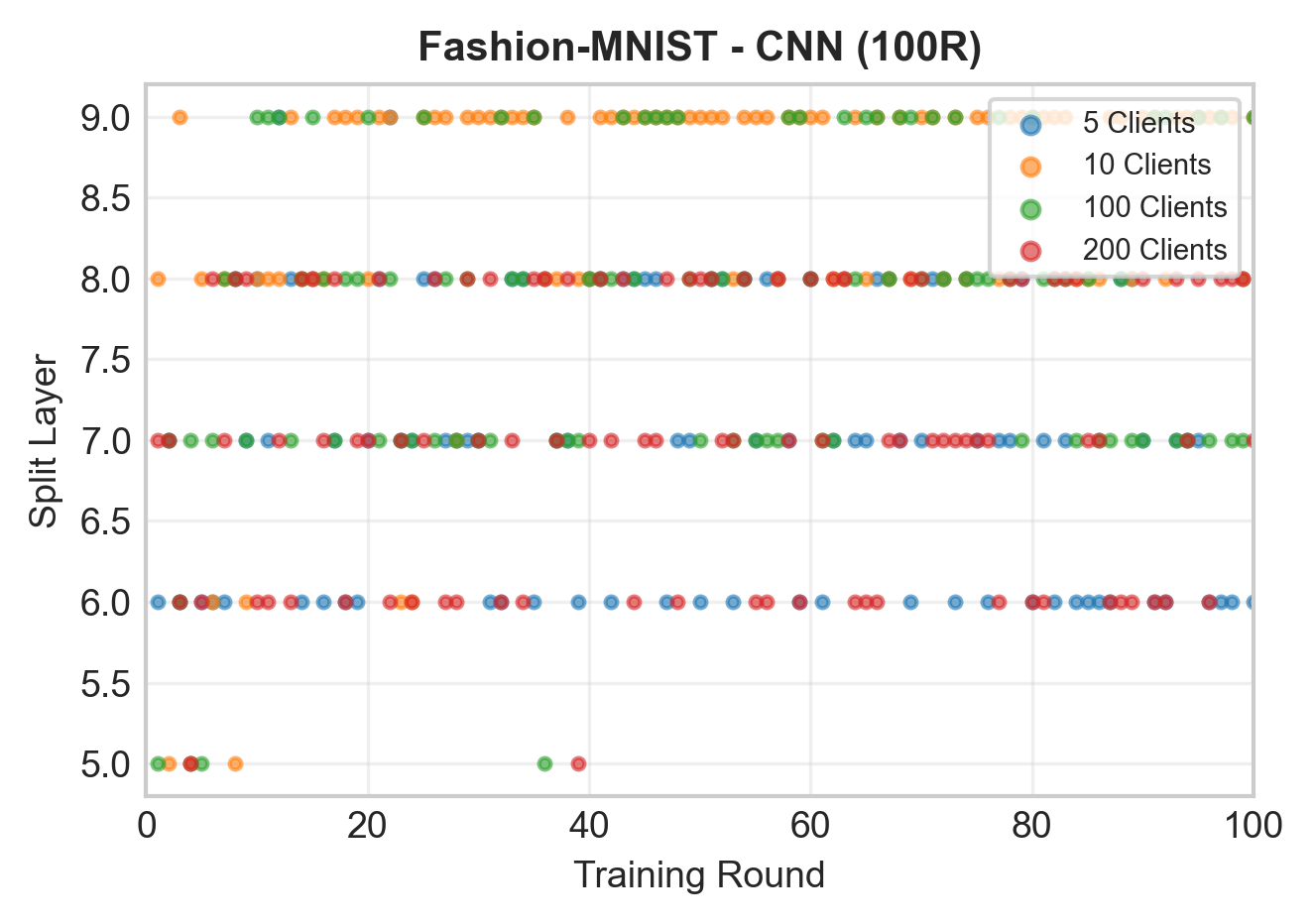}\caption{FMNIST}\end{subfigure}
\begin{subfigure}{0.48\columnwidth}\includegraphics[width=\textwidth]{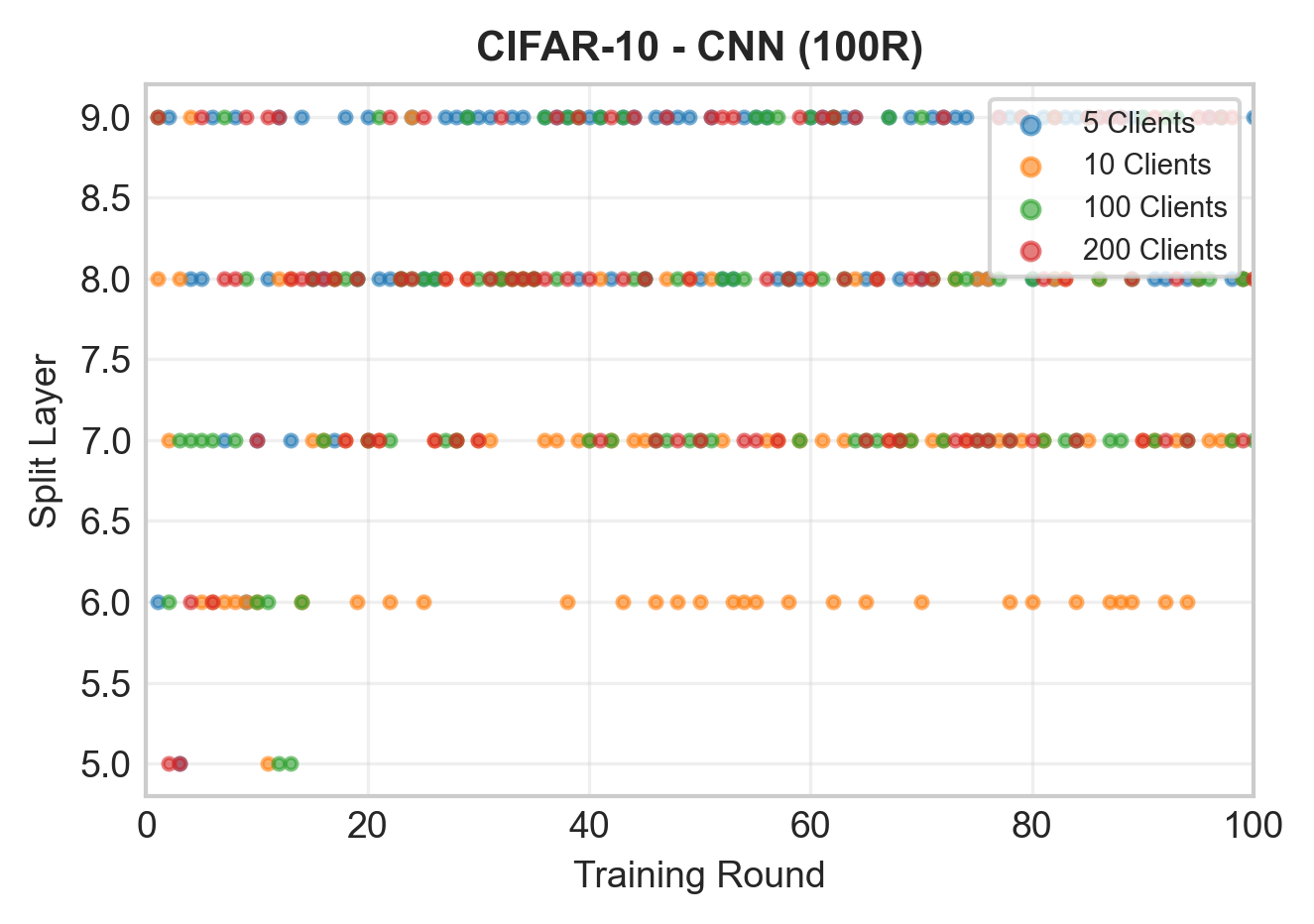}\caption{CIFAR10}\end{subfigure}
\begin{subfigure}{0.48\columnwidth}\includegraphics[width=\textwidth]{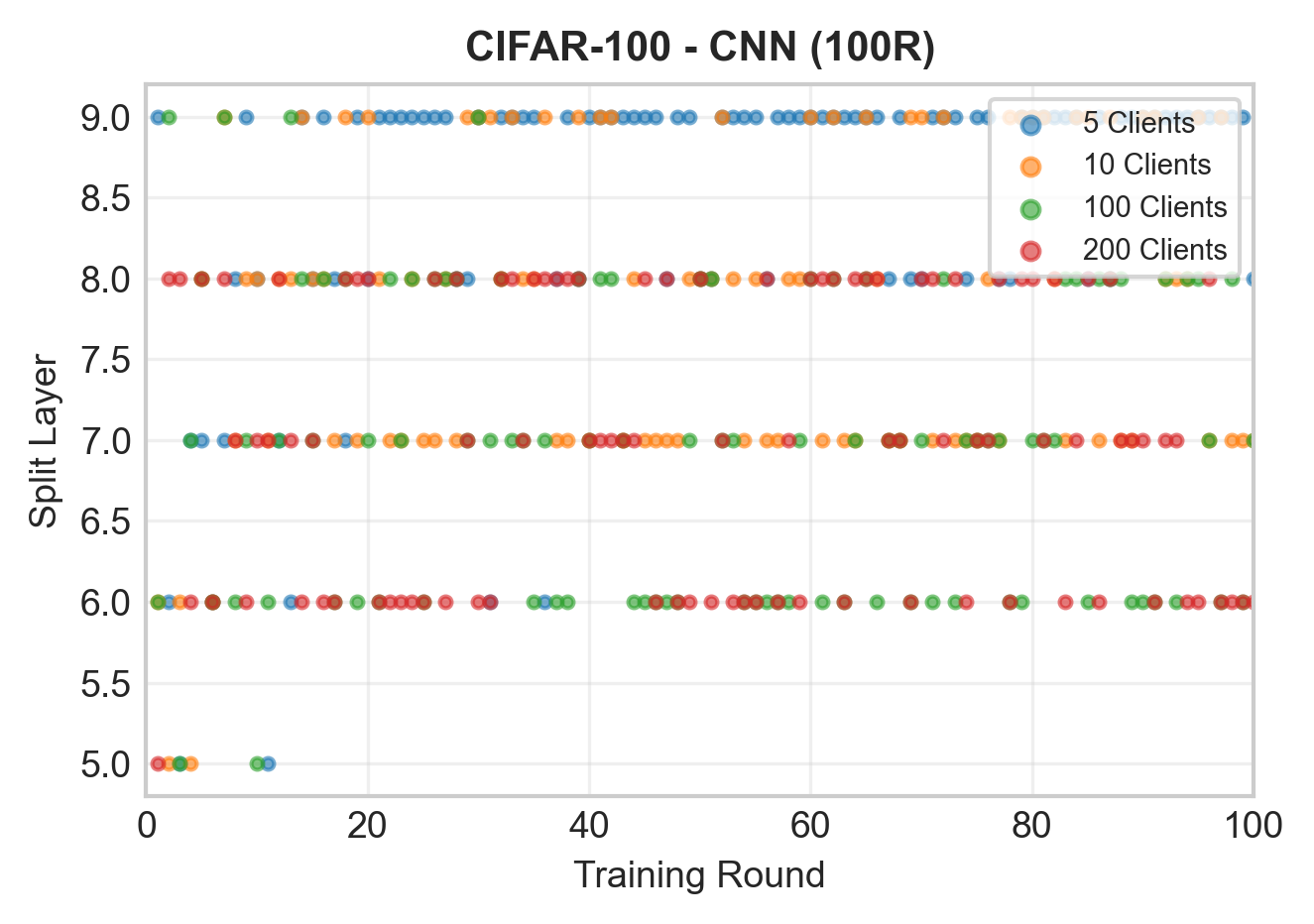}\caption{CIFAR100}\end{subfigure}
\caption{CNN Architecture Split Point Selection Analysis.}
\label{fig:cnn_split}
\end{figure}

Here in Figure~\ref{fig:cnn_split} illustrates the split point selection behavior of the QSplitFL reinforcement learning agent for the lightweight CNN architecture (10 total layers, action space: layers 5--9) across all four benchmark datasets. Unlike deeper architectures that utilize a wide range of split layers (30--50), the CNN consistently converges to earlier split points (typically layers 5--7).
\subsubsection{CNN Accuracy Convergence Analysis}
\begin{figure}[!ht]
\centering
\begin{subfigure}{0.48\columnwidth}\includegraphics[width=\textwidth]{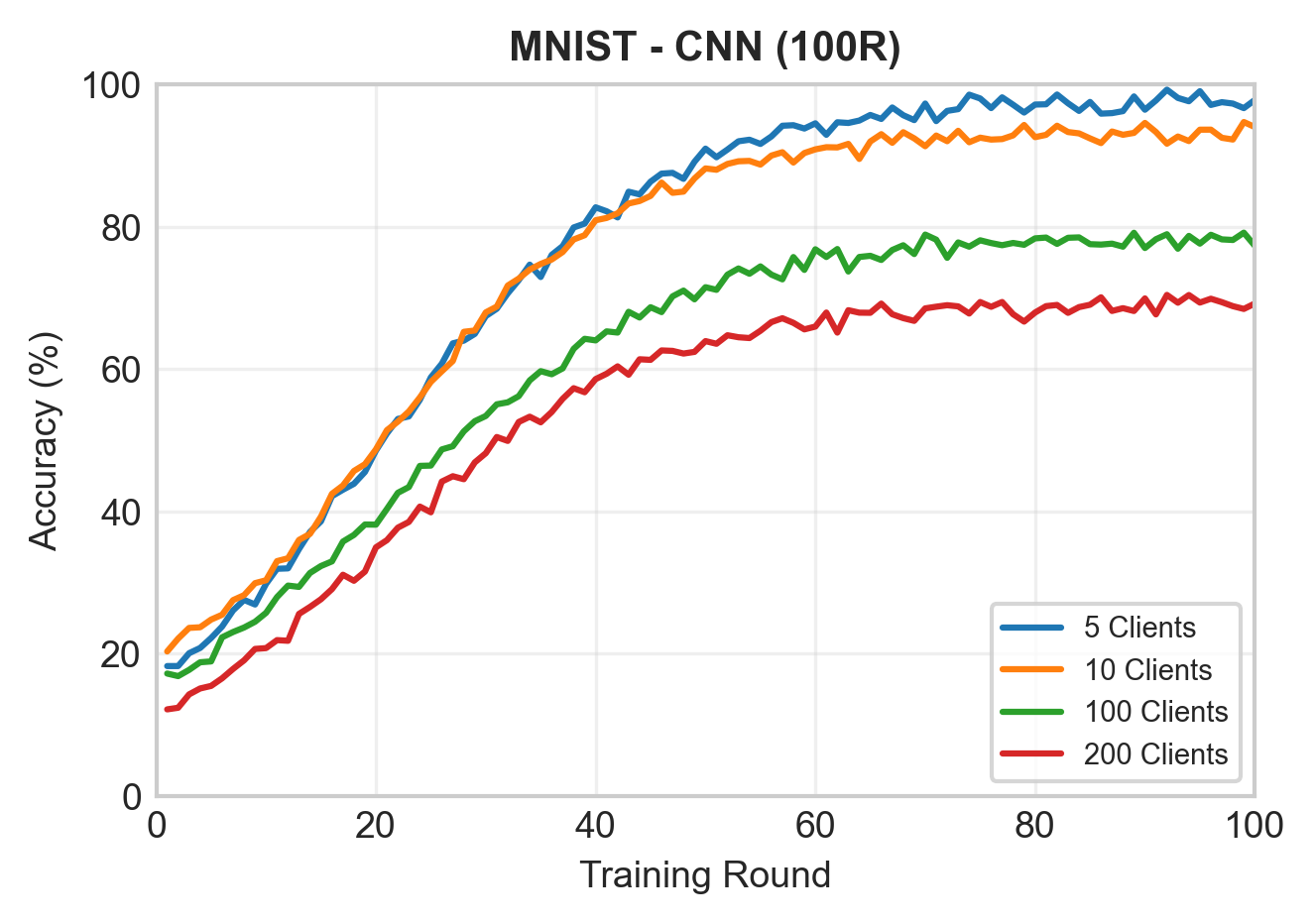}\caption{MNIST}\end{subfigure}
\begin{subfigure}{0.48\columnwidth}\includegraphics[width=\textwidth]{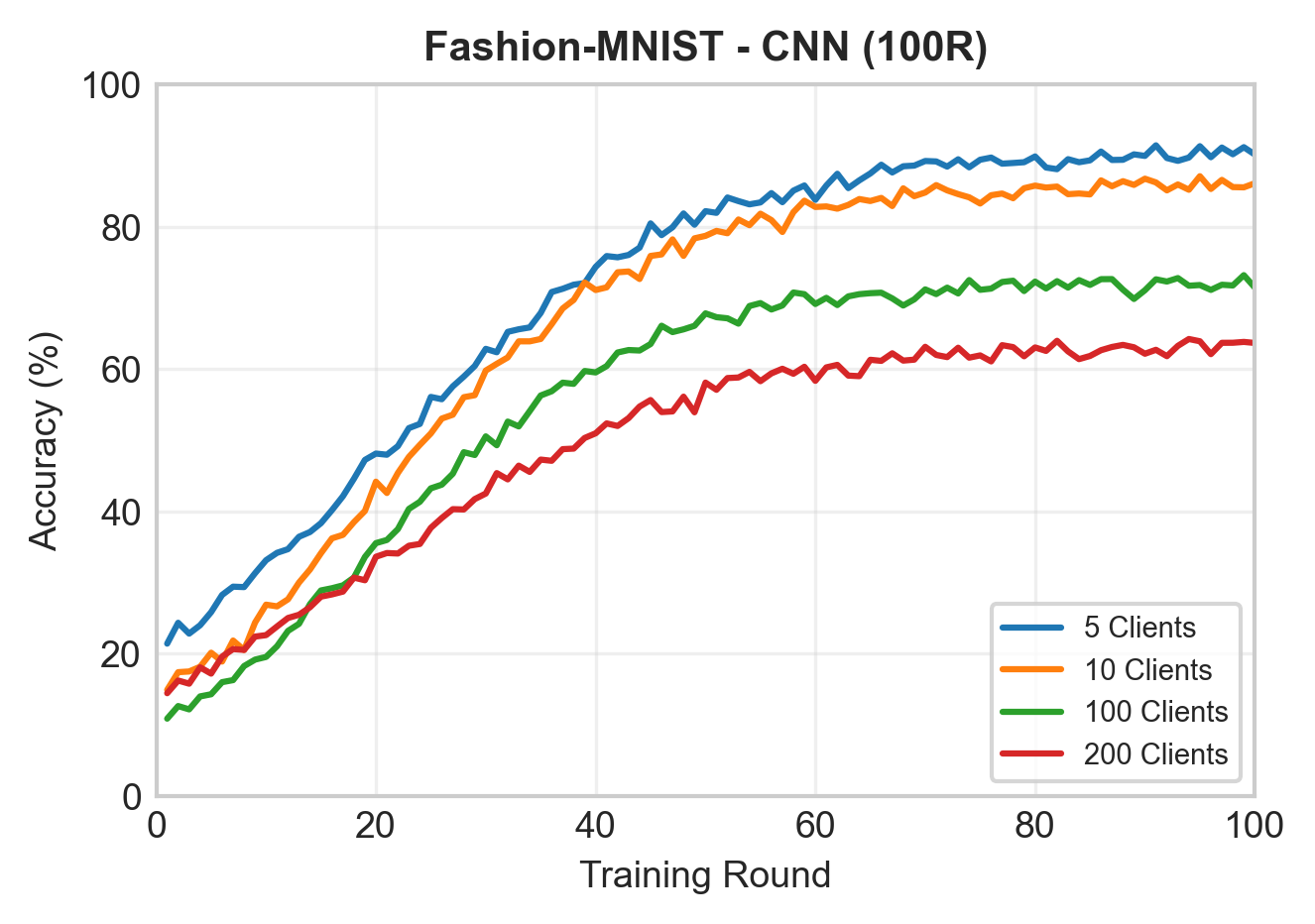}\caption{FMNIST}\end{subfigure}
\begin{subfigure}{0.48\columnwidth}\includegraphics[width=\textwidth]{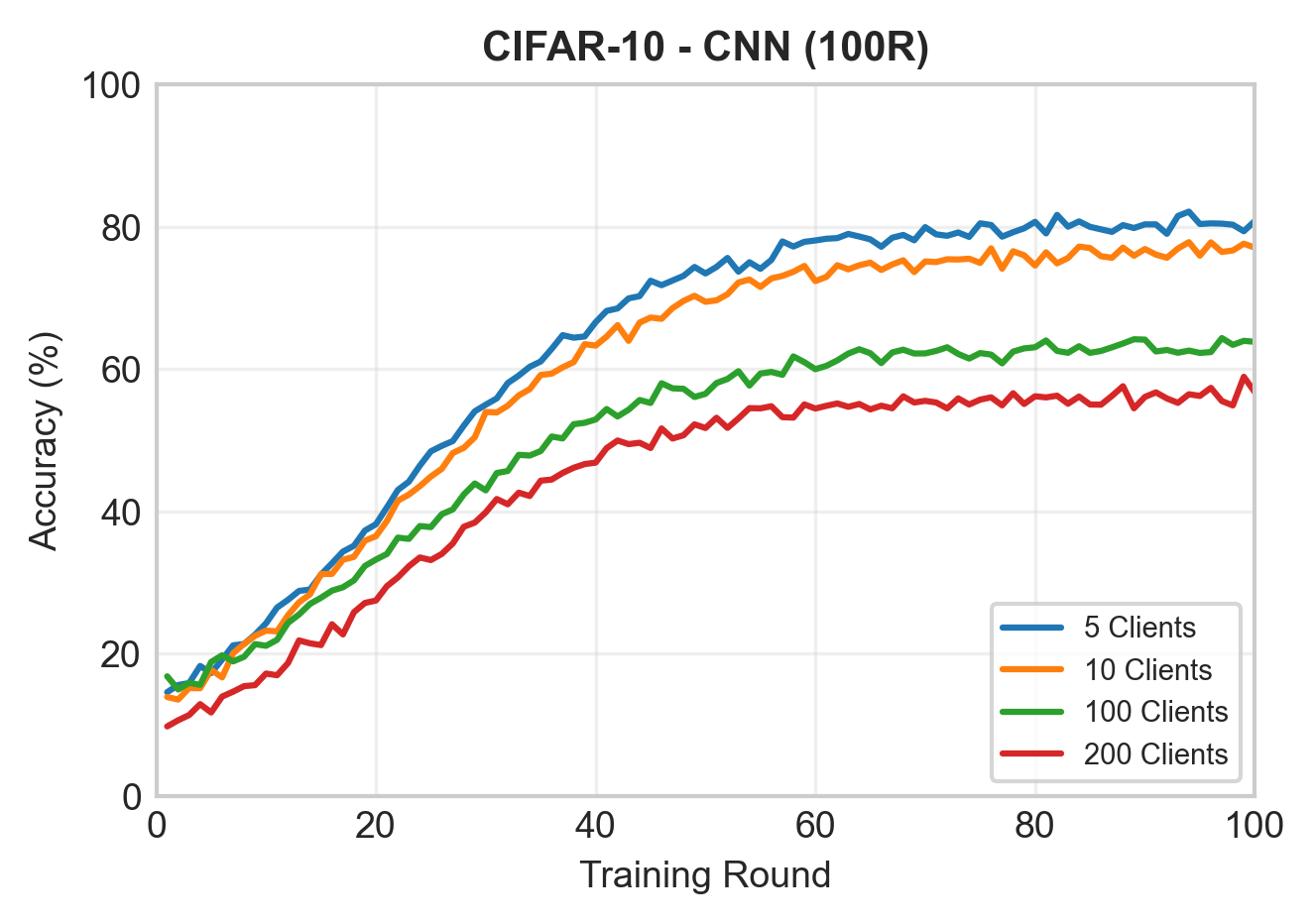}\caption{CIFAR10}\end{subfigure}
\begin{subfigure}{0.48\columnwidth}\includegraphics[width=\textwidth]{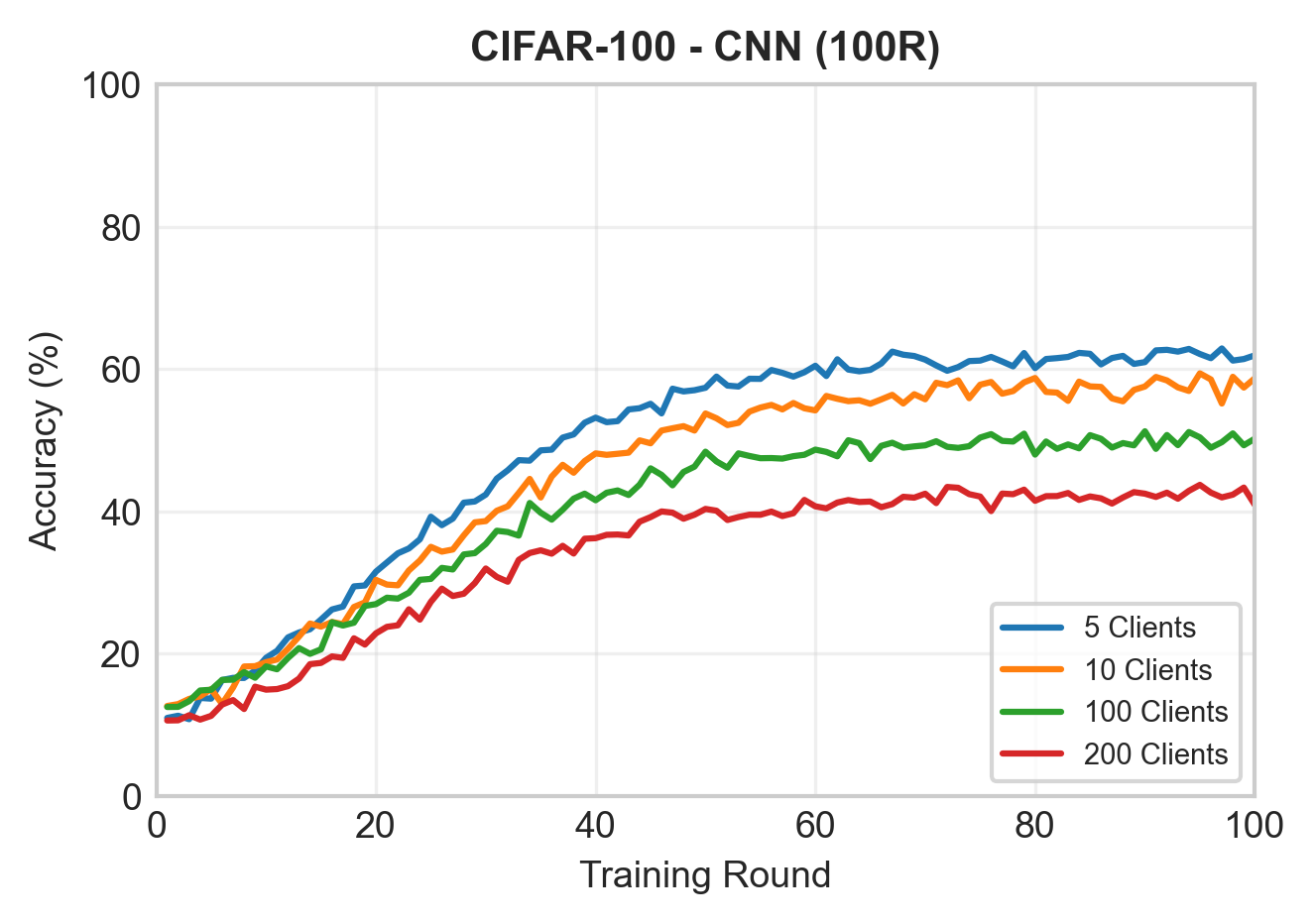}\caption{CIFAR100}\end{subfigure}
\caption{\textbf{CNN Accuracy Convergence (100 Rounds).} Stable convergence on simple tasks contrasts with limited performance on CIFAR-100 due to shallow depth.}
\label{fig:cnn_acc_convergence}
\end{figure}

Figure~\ref{fig:cnn_acc_convergence} shows accuracy convergence for the 10-layer CNN. While it converges stably on simple tasks (MNIST $>$99\%), its limited capacity significantly effects the performance on complex datasets like CIFAR-100 (around 50\%), which underscores the necessity for deeper architectures in complex scenarios.

\subsubsection{Accuracy Convergence}
\begin{figure*}[!htbp]
\centering
\begin{minipage}{\textwidth}
\centering
\begin{subfigure}{0.23\textwidth}\includegraphics[width=\textwidth]{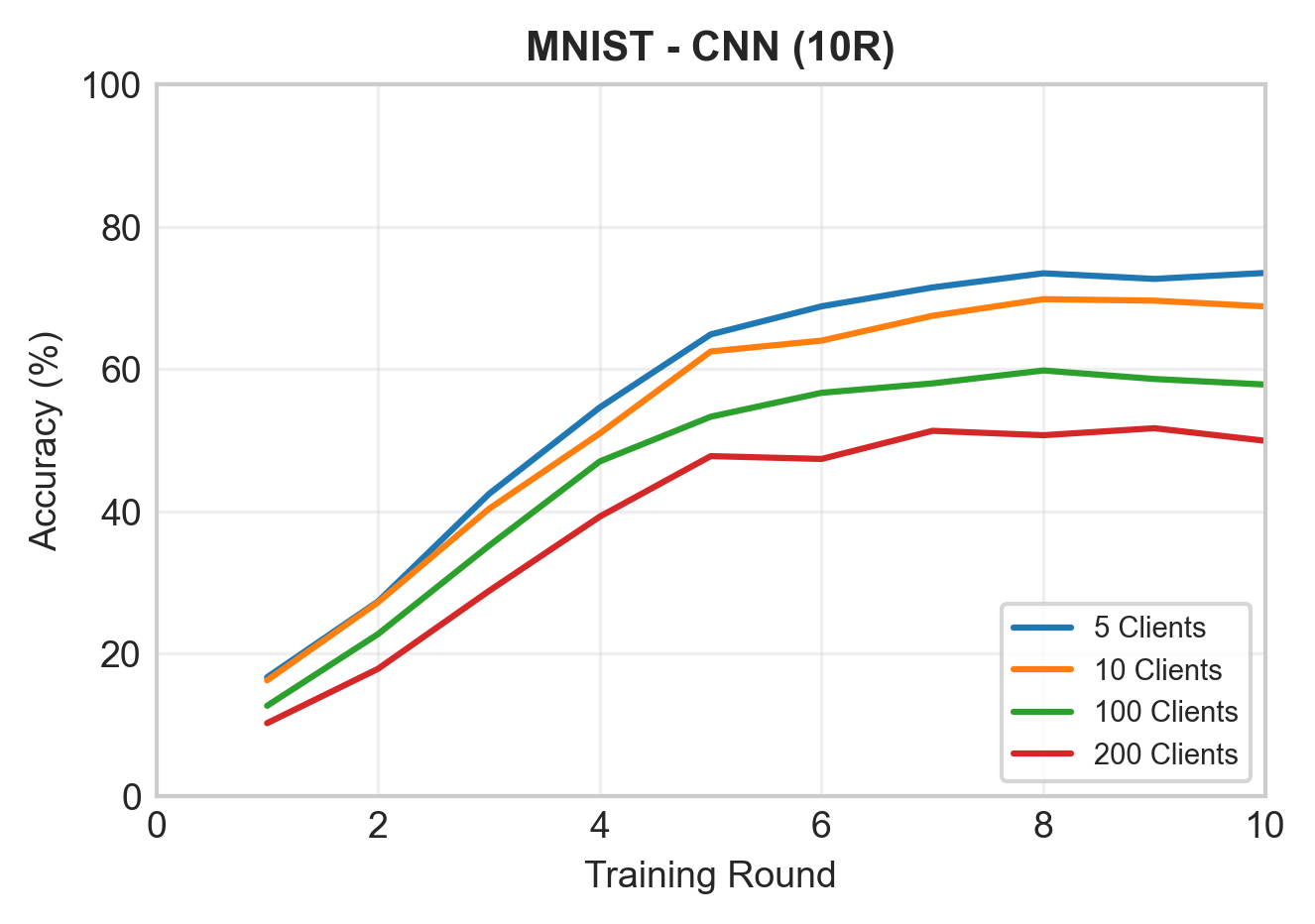}\end{subfigure}
\begin{subfigure}{0.23\textwidth}\includegraphics[width=\textwidth]{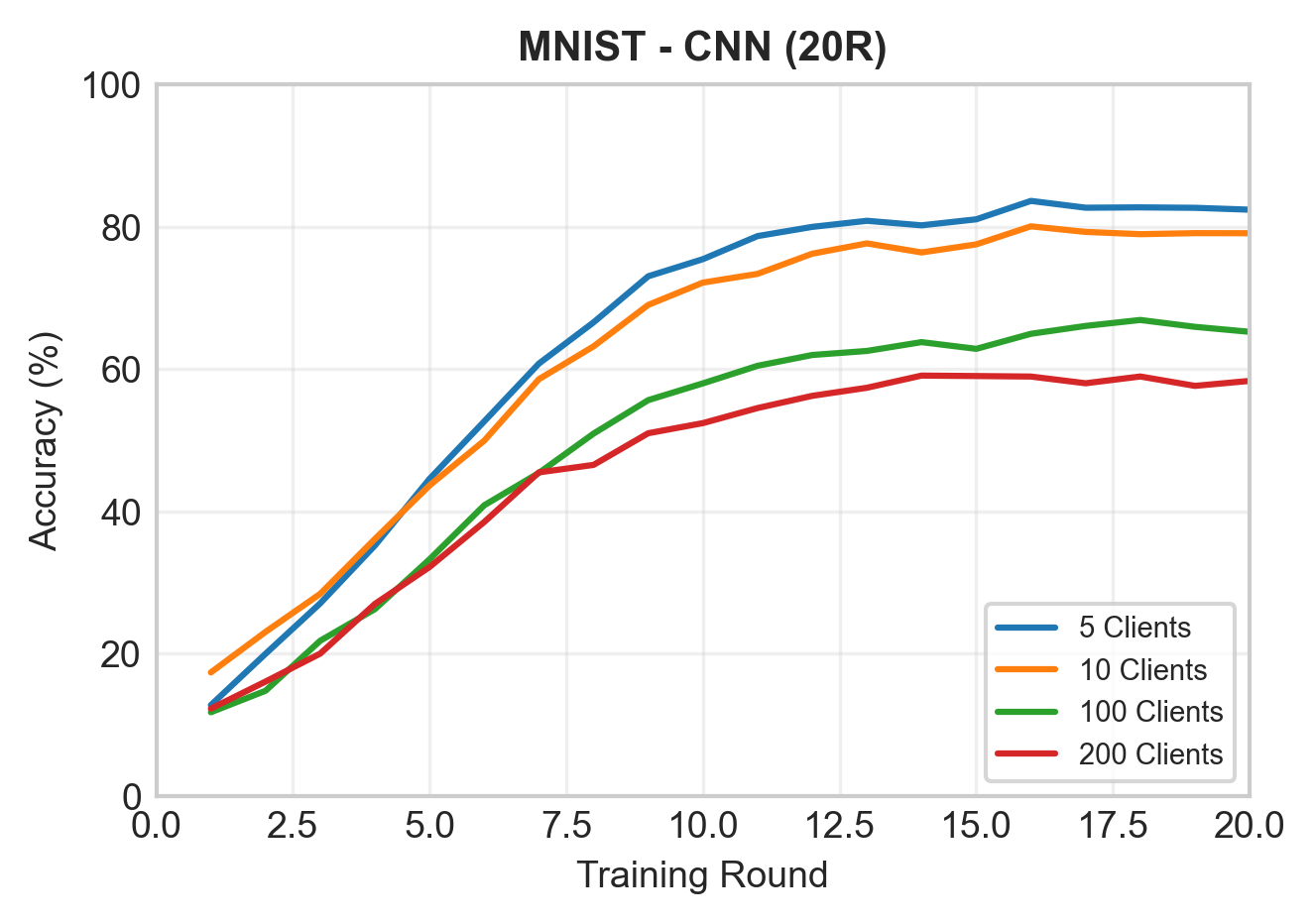}\end{subfigure}
\begin{subfigure}{0.23\textwidth}\includegraphics[width=\textwidth]{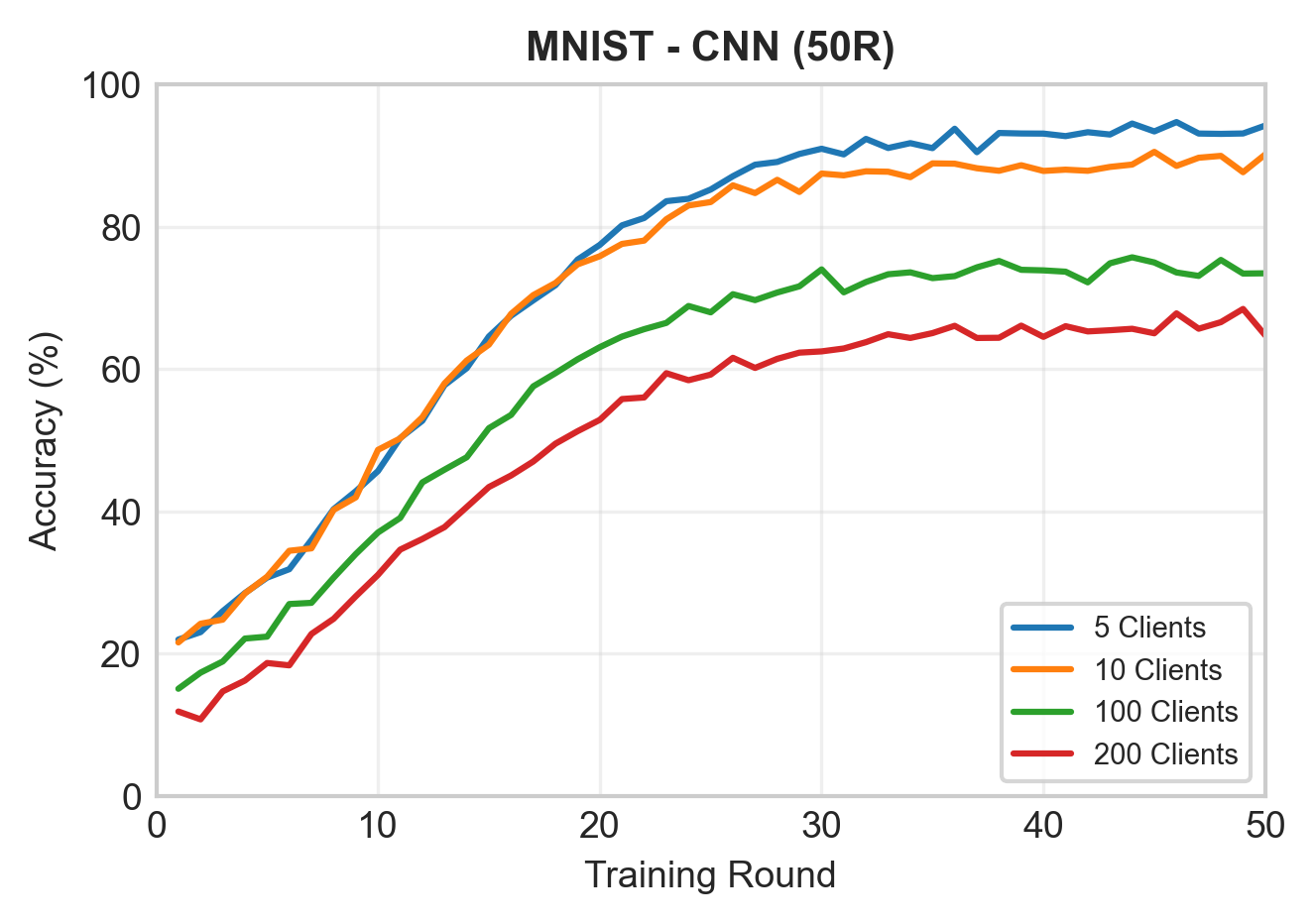}\end{subfigure}
\begin{subfigure}{0.23\textwidth}\includegraphics[width=\textwidth]{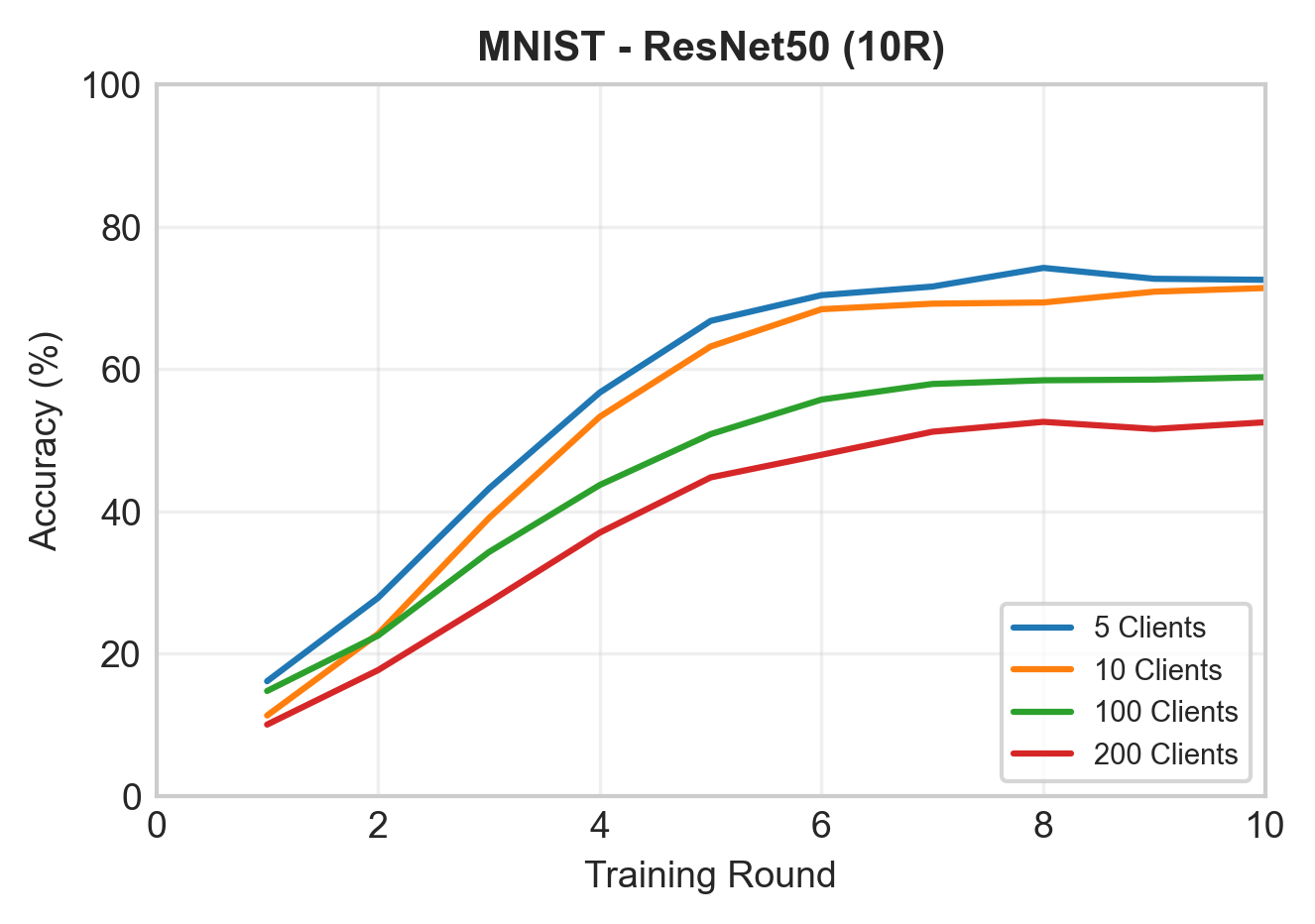}\end{subfigure}
\begin{subfigure}{0.23\textwidth}\includegraphics[width=\textwidth]{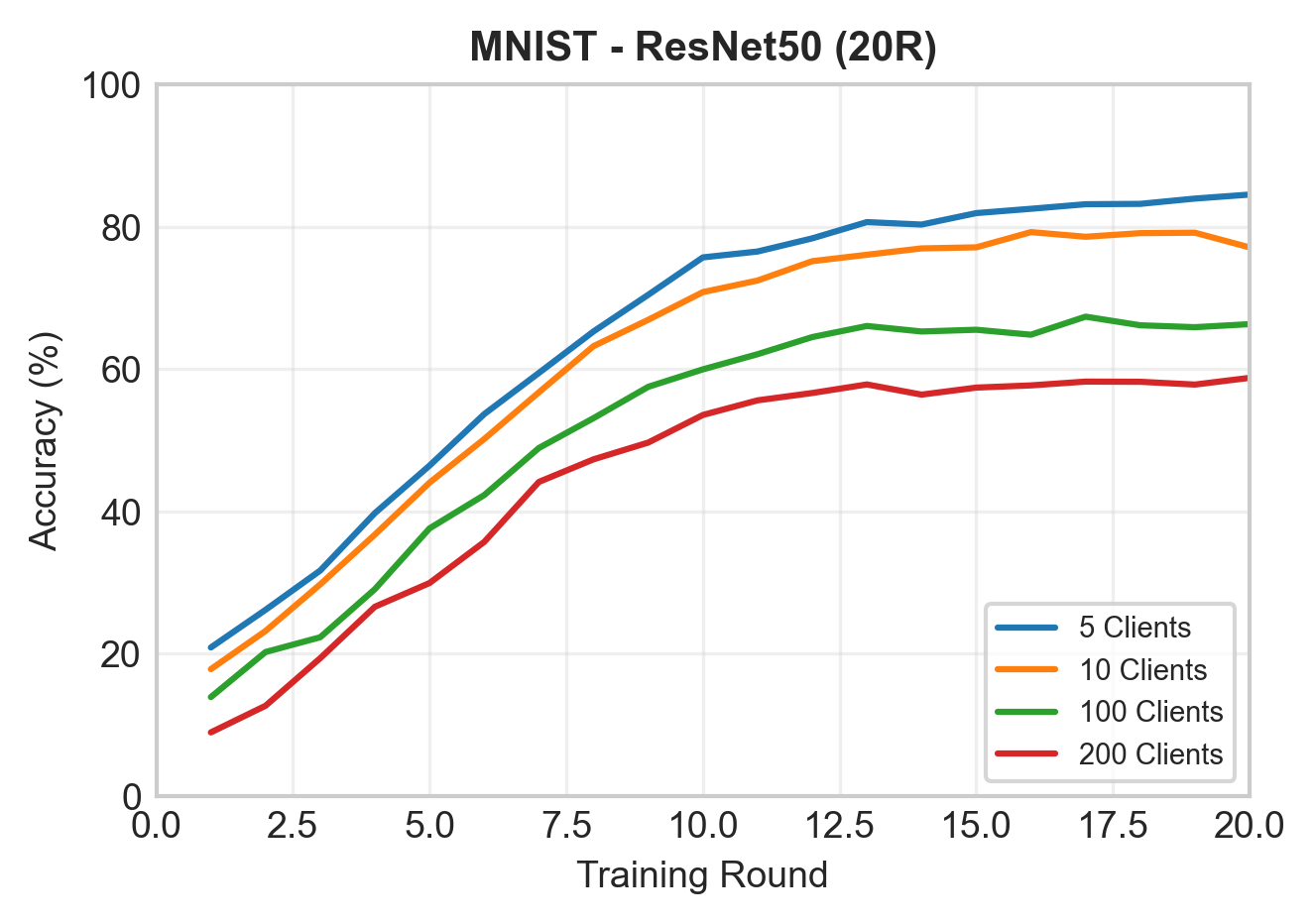}\end{subfigure}
\begin{subfigure}{0.23\textwidth}\includegraphics[width=\textwidth]{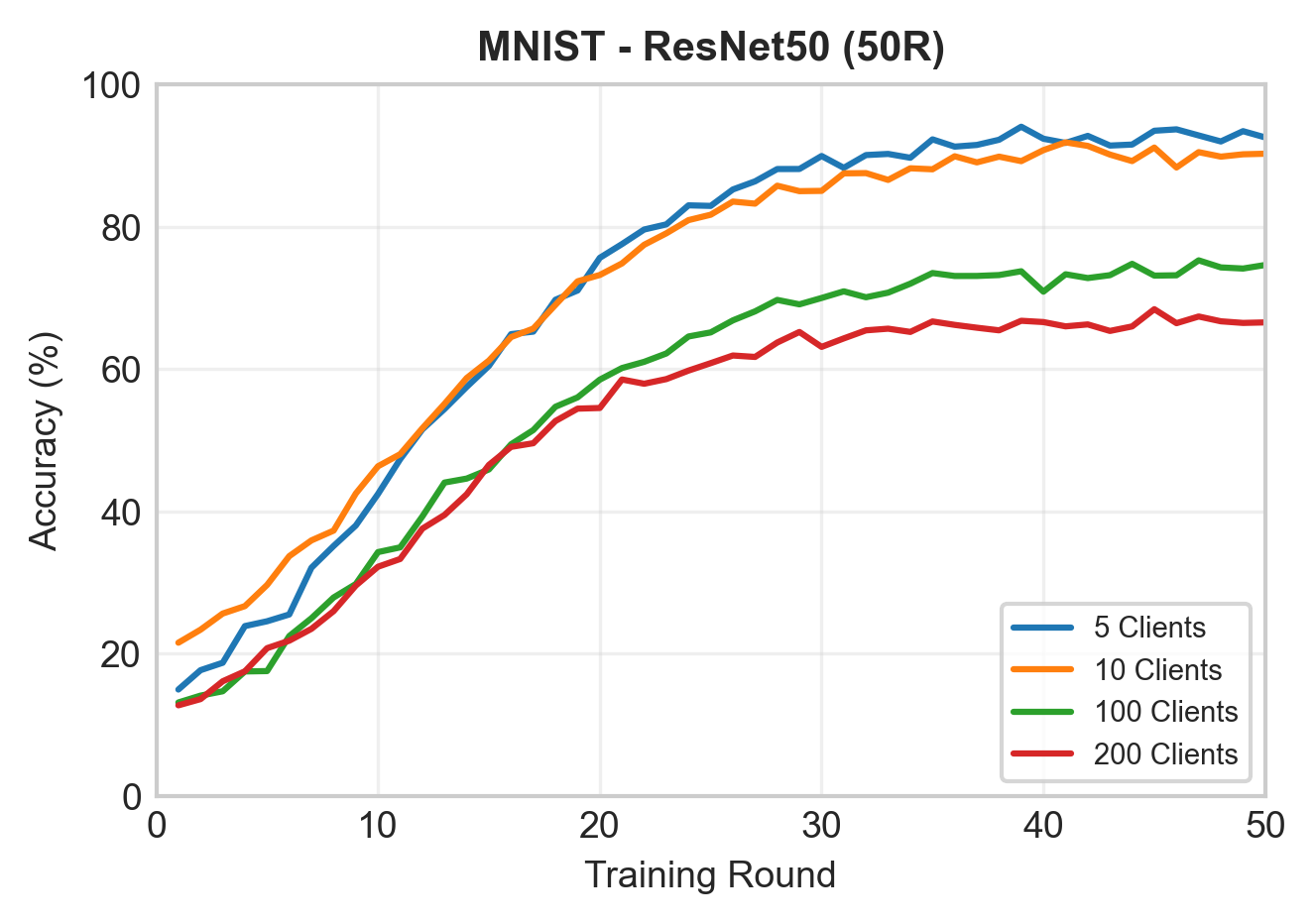}\end{subfigure}
\begin{subfigure}{0.23\textwidth}\includegraphics[width=\textwidth]{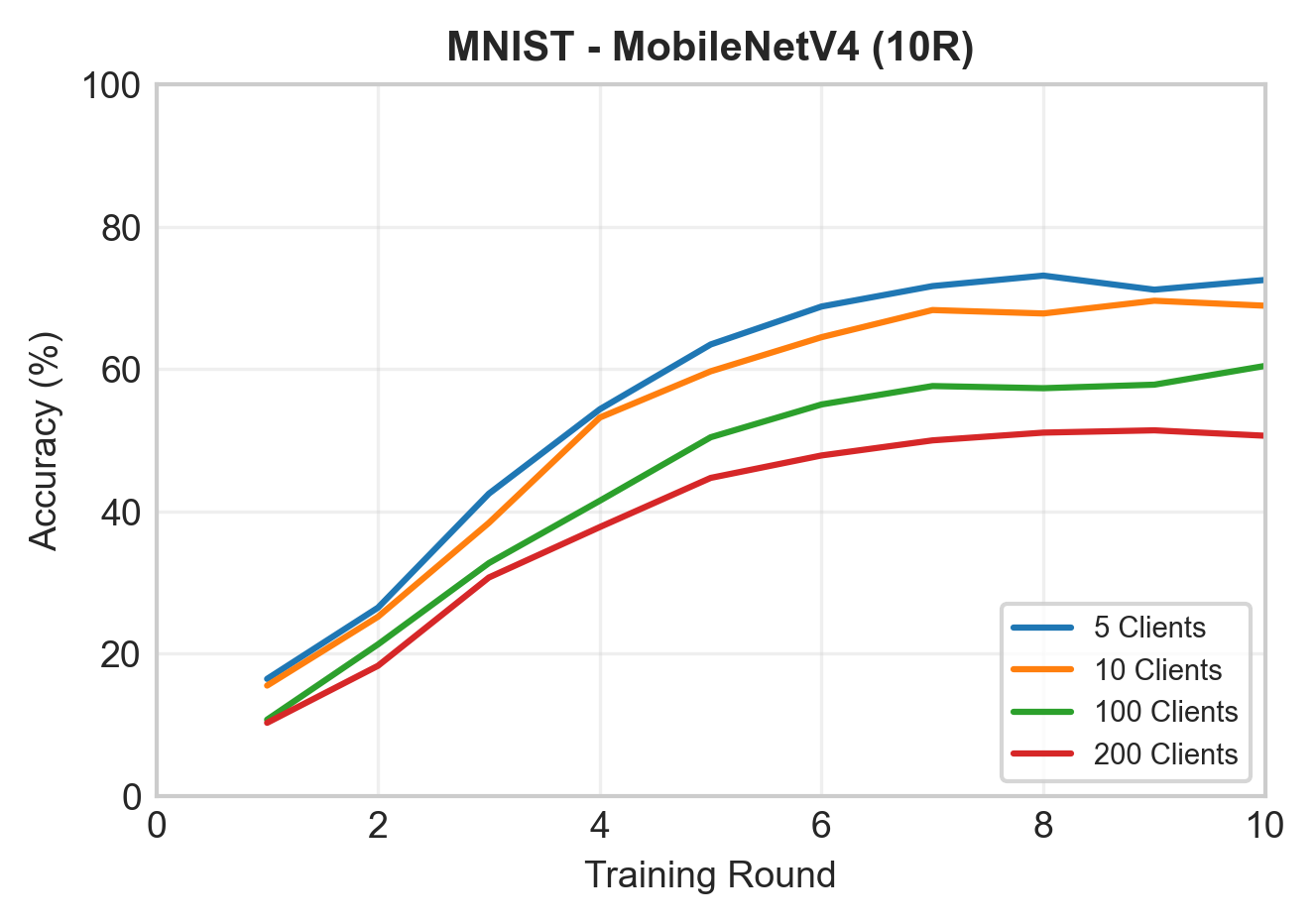}\end{subfigure}
\begin{subfigure}{0.23\textwidth}\includegraphics[width=\textwidth]{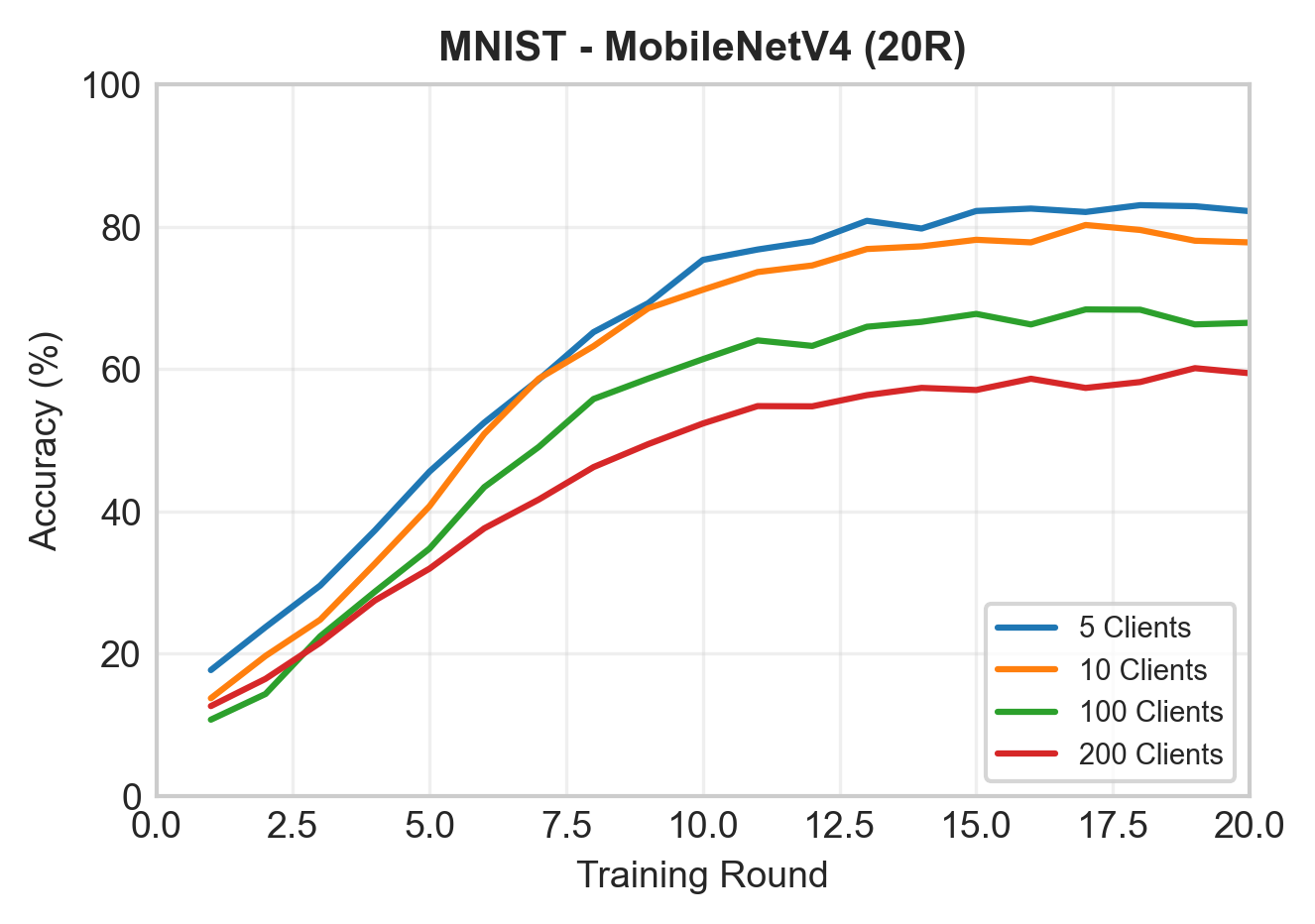}\end{subfigure}
\begin{subfigure}{0.23\textwidth}\includegraphics[width=\textwidth]{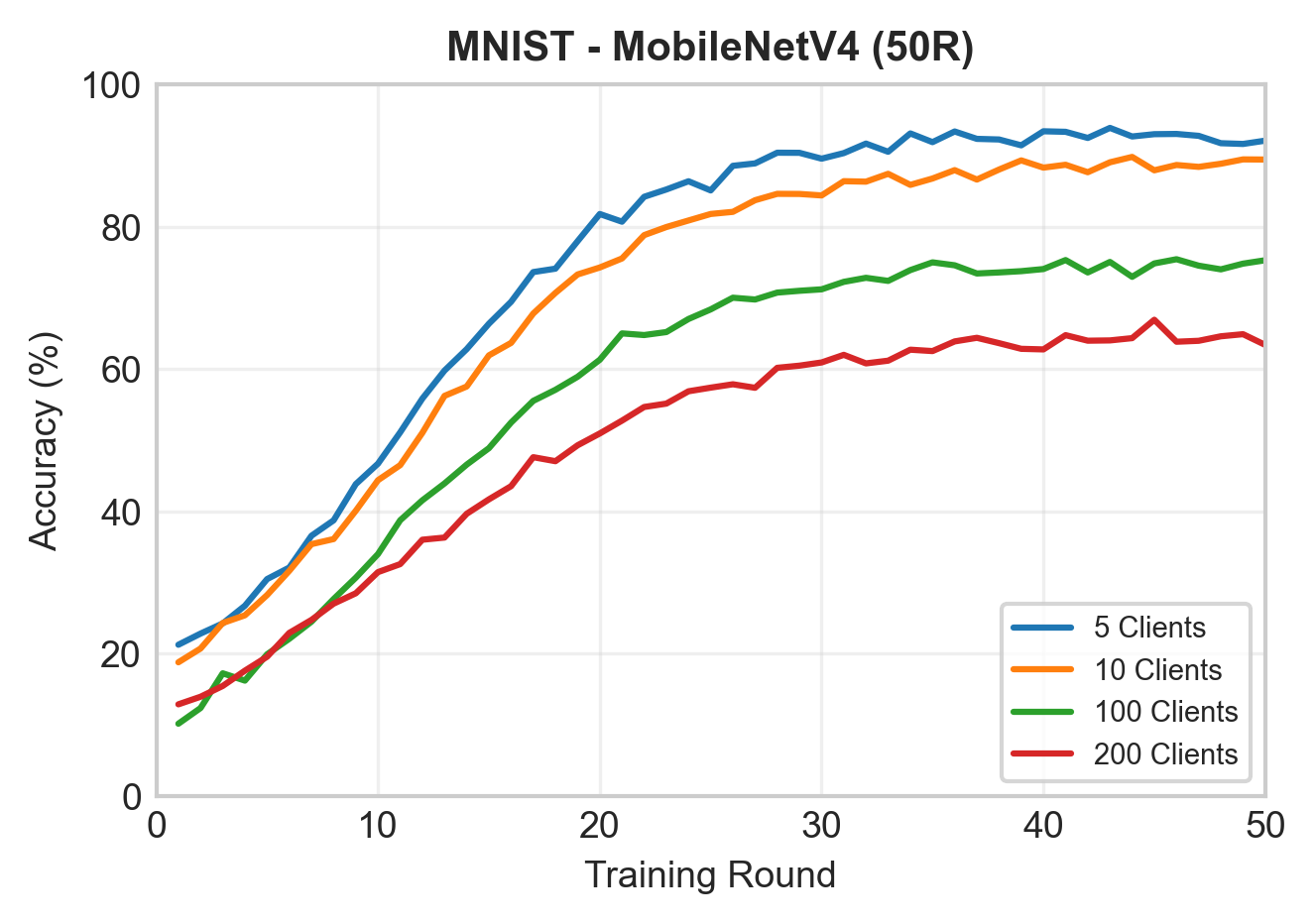}\end{subfigure}
\begin{subfigure}{0.23\textwidth}\includegraphics[width=\textwidth]{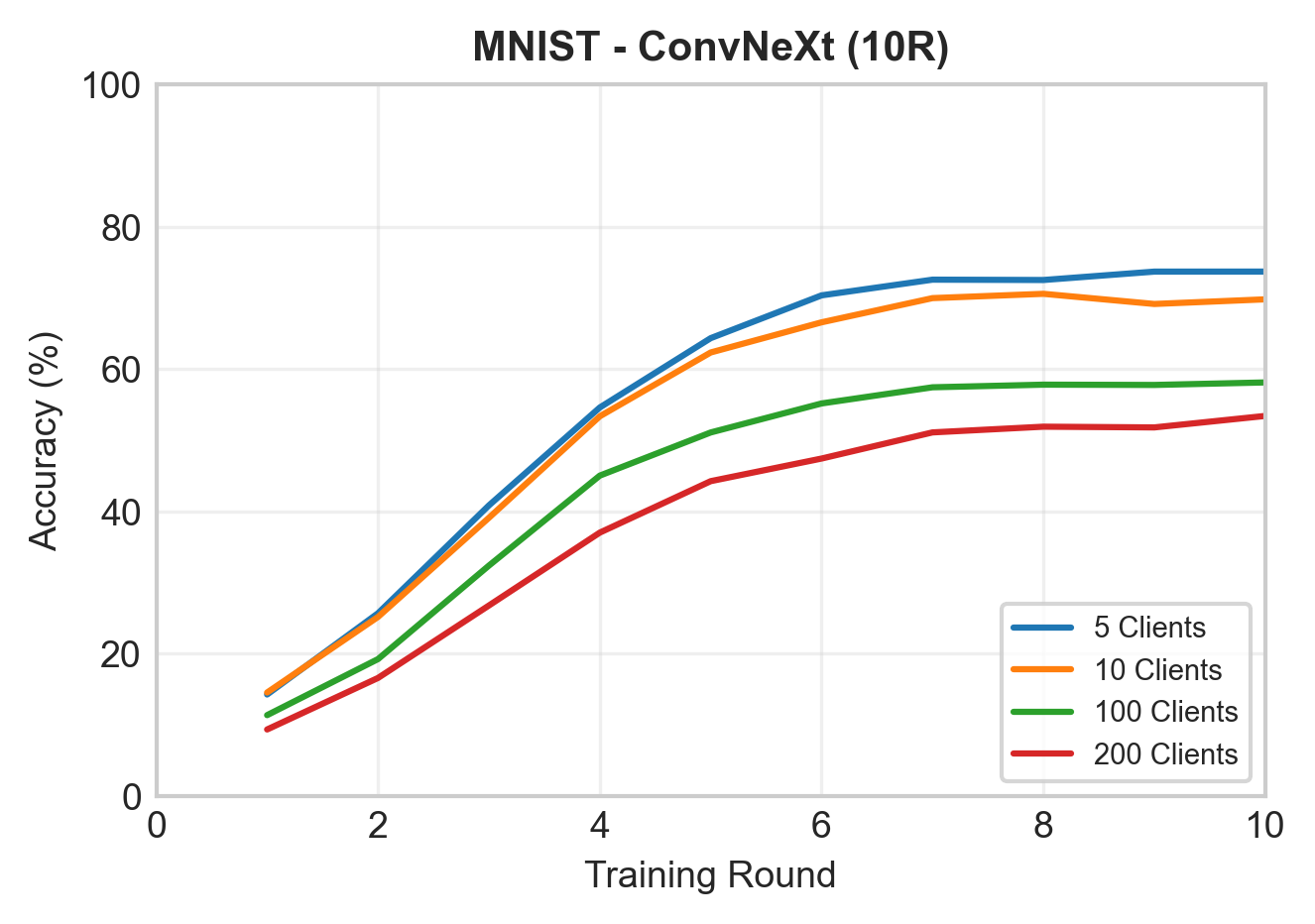}\end{subfigure}
\begin{subfigure}{0.23\textwidth}\includegraphics[width=\textwidth]{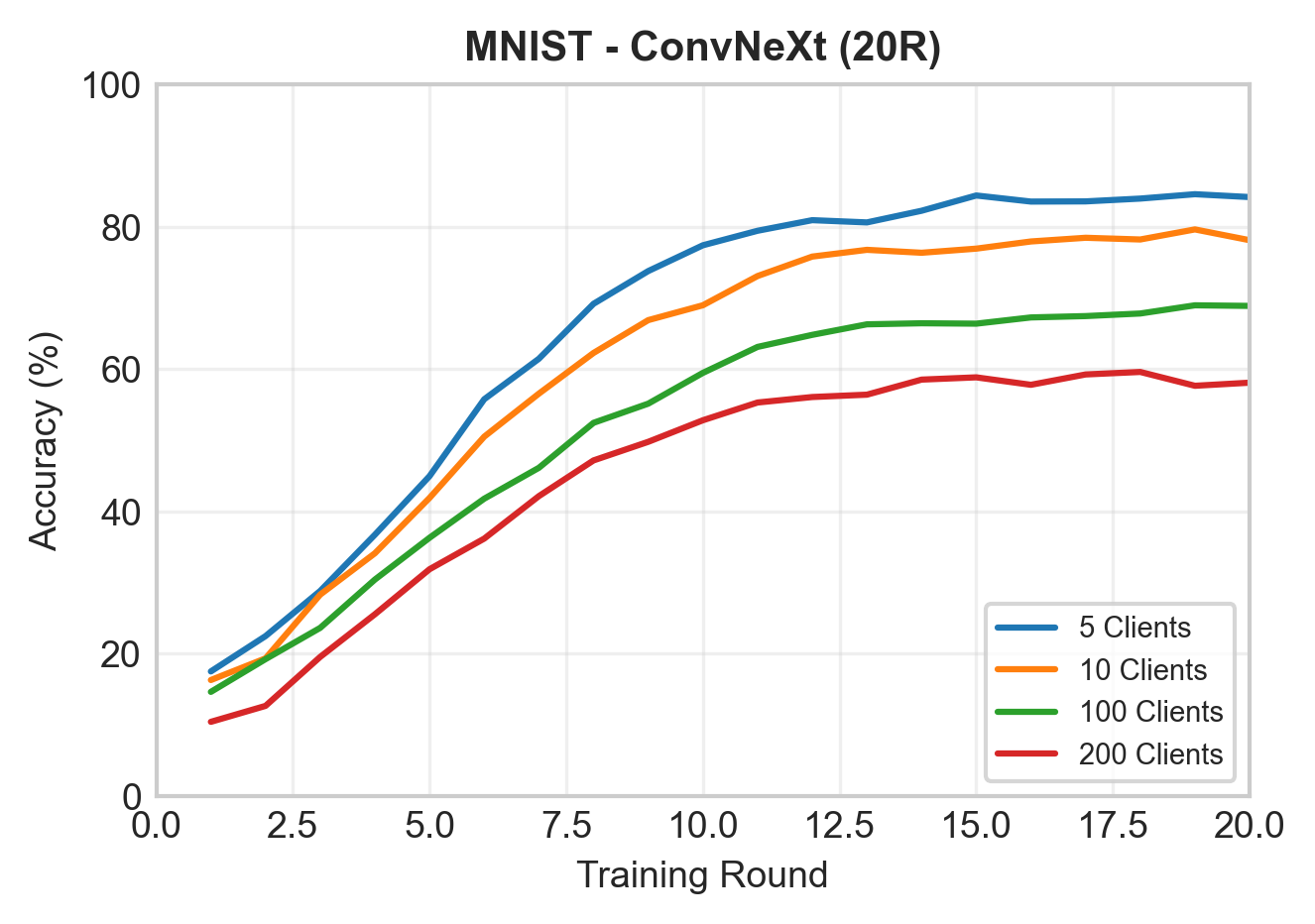}\end{subfigure}
\begin{subfigure}{0.23\textwidth}\includegraphics[width=\textwidth]{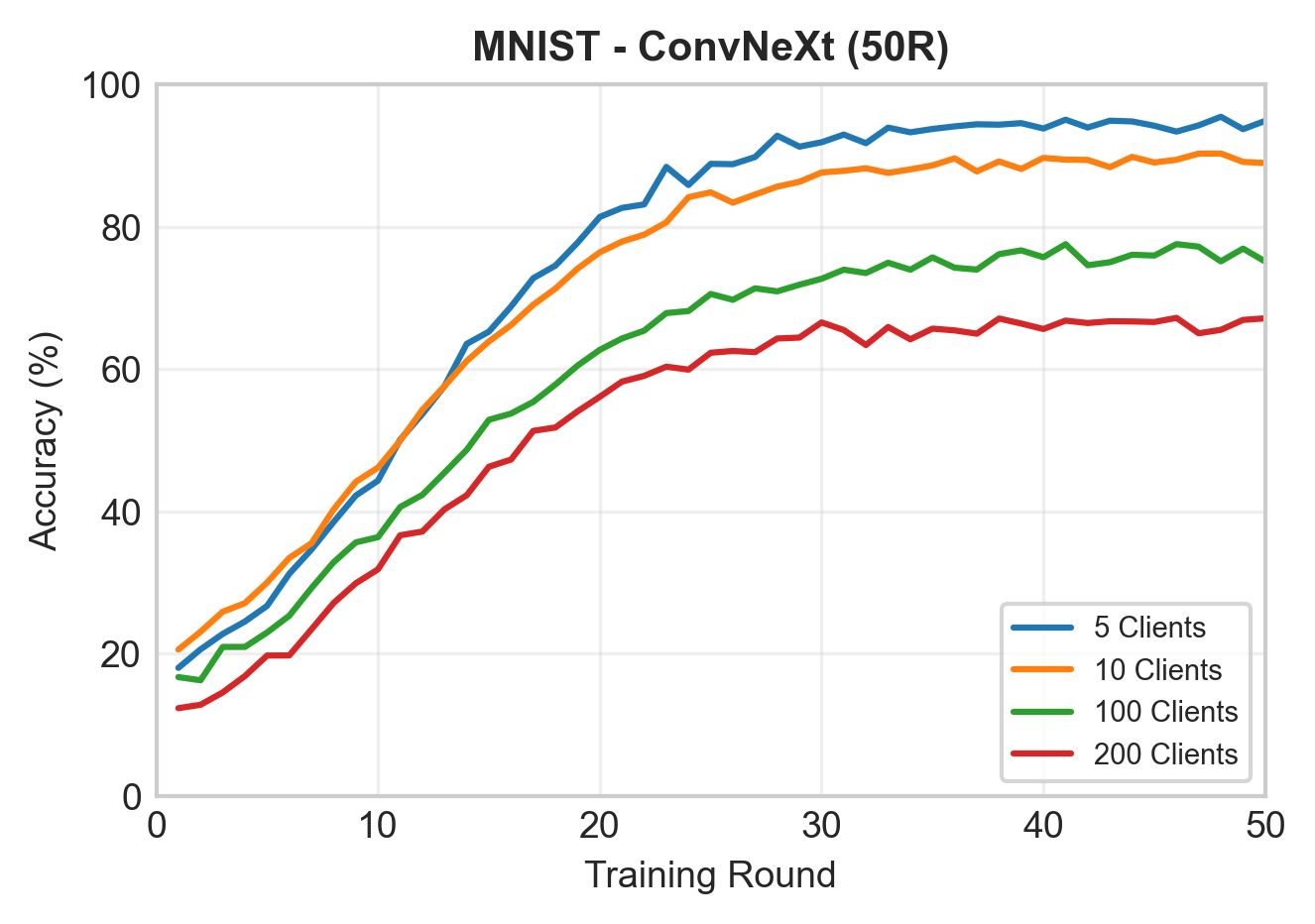}\end{subfigure}
\caption{\textbf{MNIST Dataset Accuracy Convergence (10, 20, 50 Rounds).} Comprehensive accuracy convergence analysis across four model architectures (CNN, ResNet50, MobileNetV4, ConvNeXt) and varying training durations. Each row corresponds to one architecture, with columns showing 10, 20, and 50 training rounds. All architectures achieve rapid convergence on this baseline dataset, reaching $>$95\% accuracy within 10 rounds and approaching 99\% by 50 rounds across all client configurations (5, 10, 100, 200 clients). The consistent performance demonstrates QSplitFL's effectiveness even on simple classification tasks, with minimal performance gap between architectures due to the dataset's low complexity.}
\label{fig:mnist_acc_additional}
\end{minipage}

\begin{minipage}{\textwidth}
\centering
\begin{subfigure}{0.23\textwidth}\includegraphics[width=\textwidth]{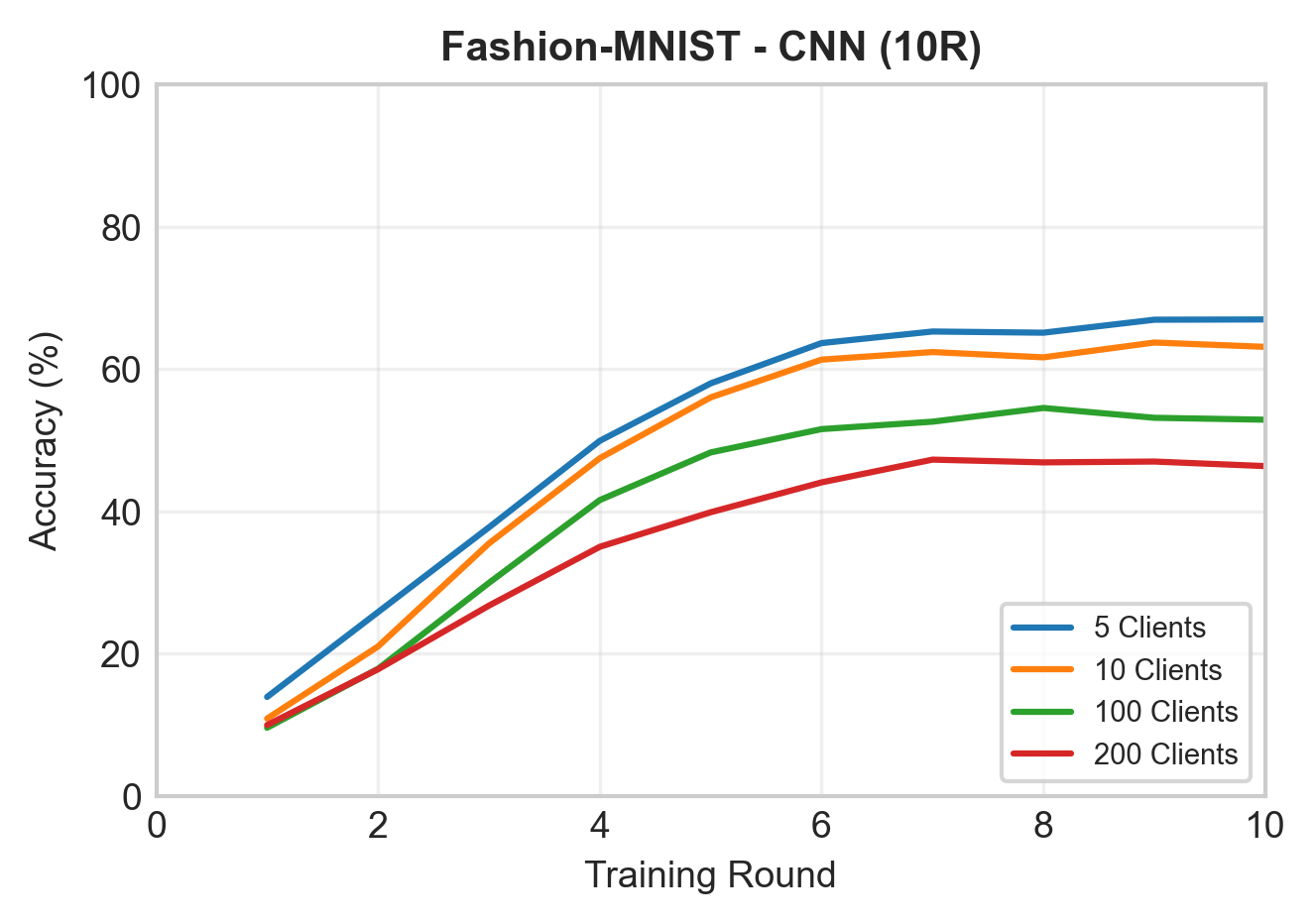}\end{subfigure}
\begin{subfigure}{0.23\textwidth}\includegraphics[width=\textwidth]{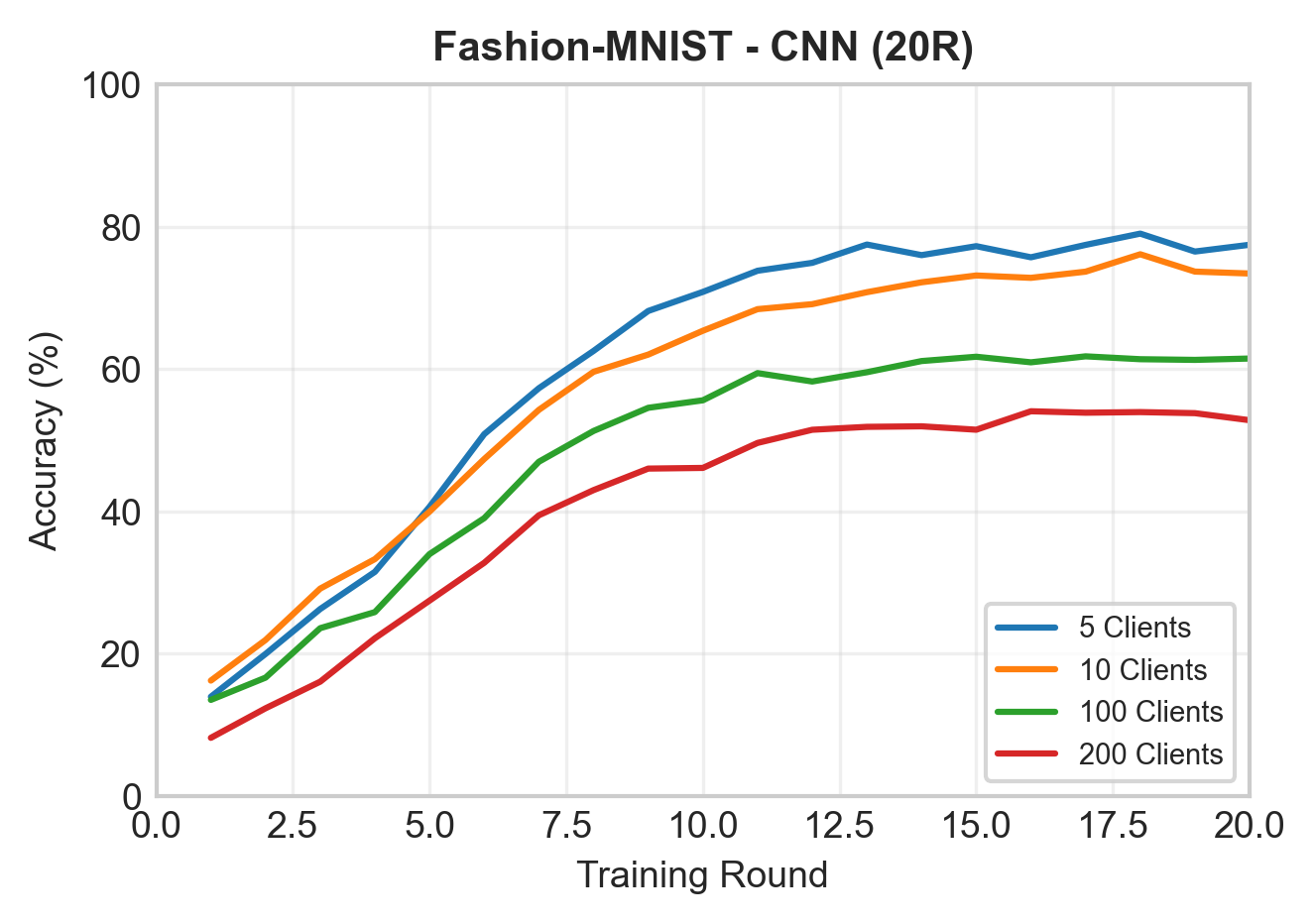}\end{subfigure}
\begin{subfigure}{0.23\textwidth}\includegraphics[width=\textwidth]{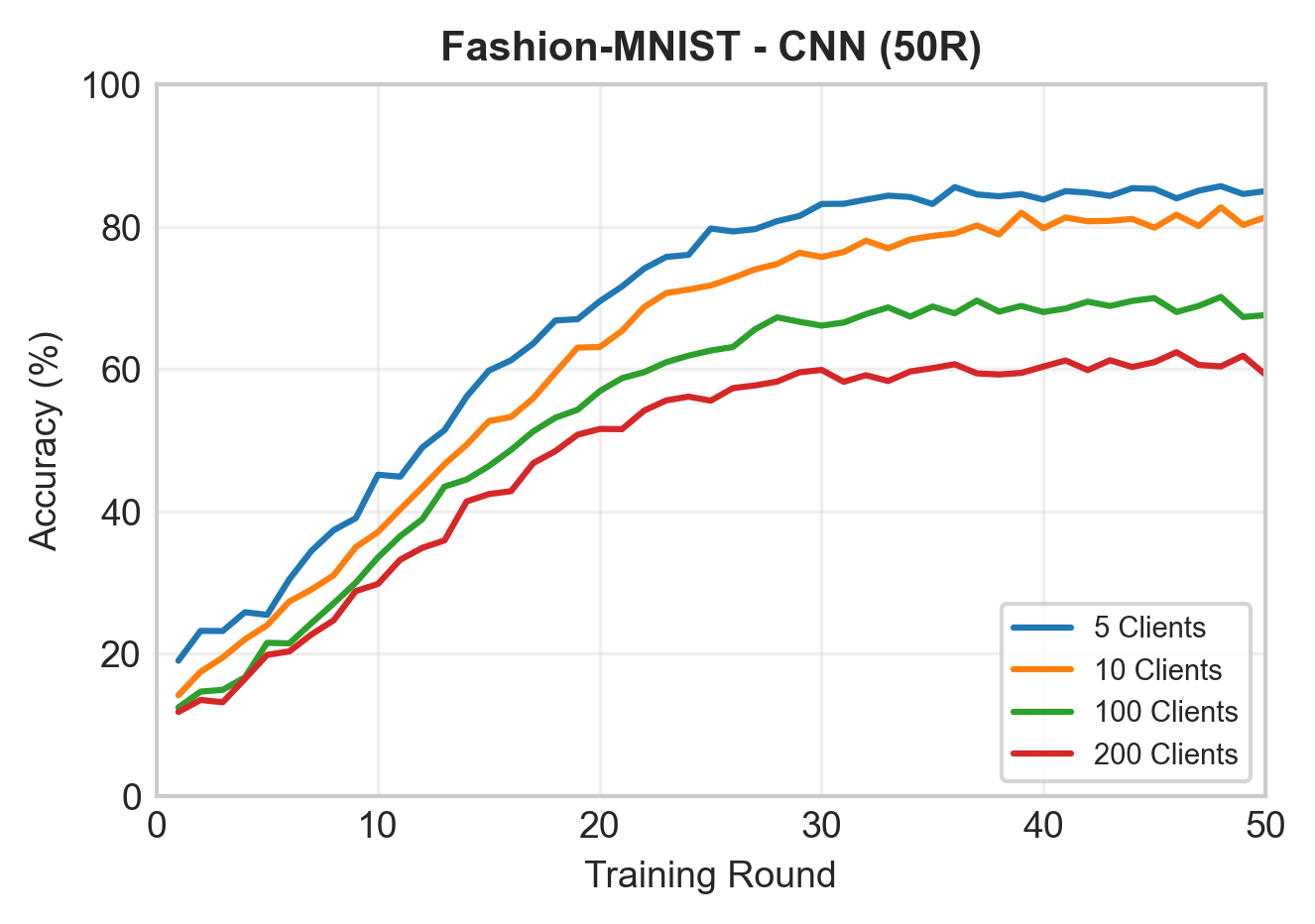}\end{subfigure}
\begin{subfigure}{0.23\textwidth}\includegraphics[width=\textwidth]{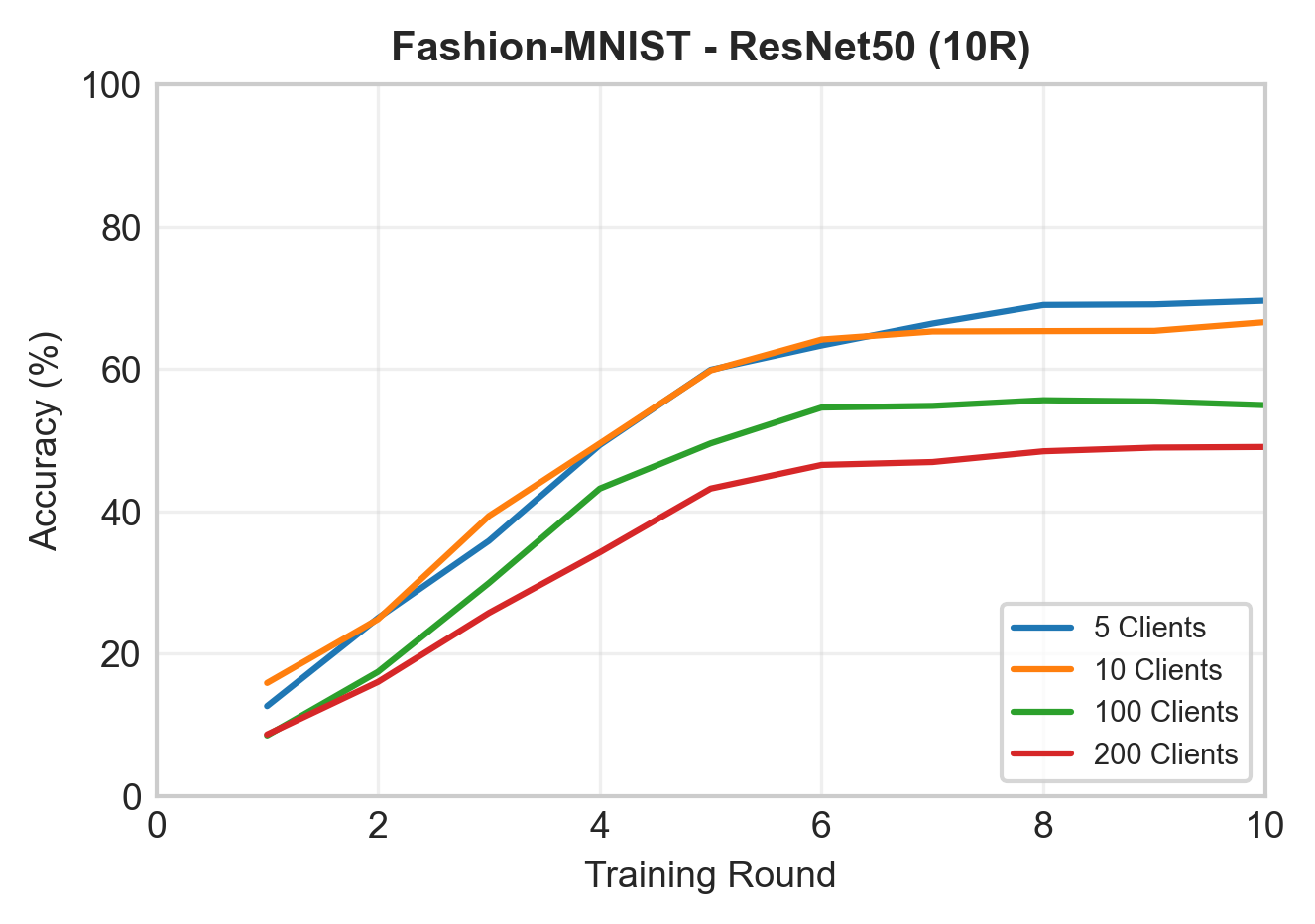}\end{subfigure}
\begin{subfigure}{0.23\textwidth}\includegraphics[width=\textwidth]{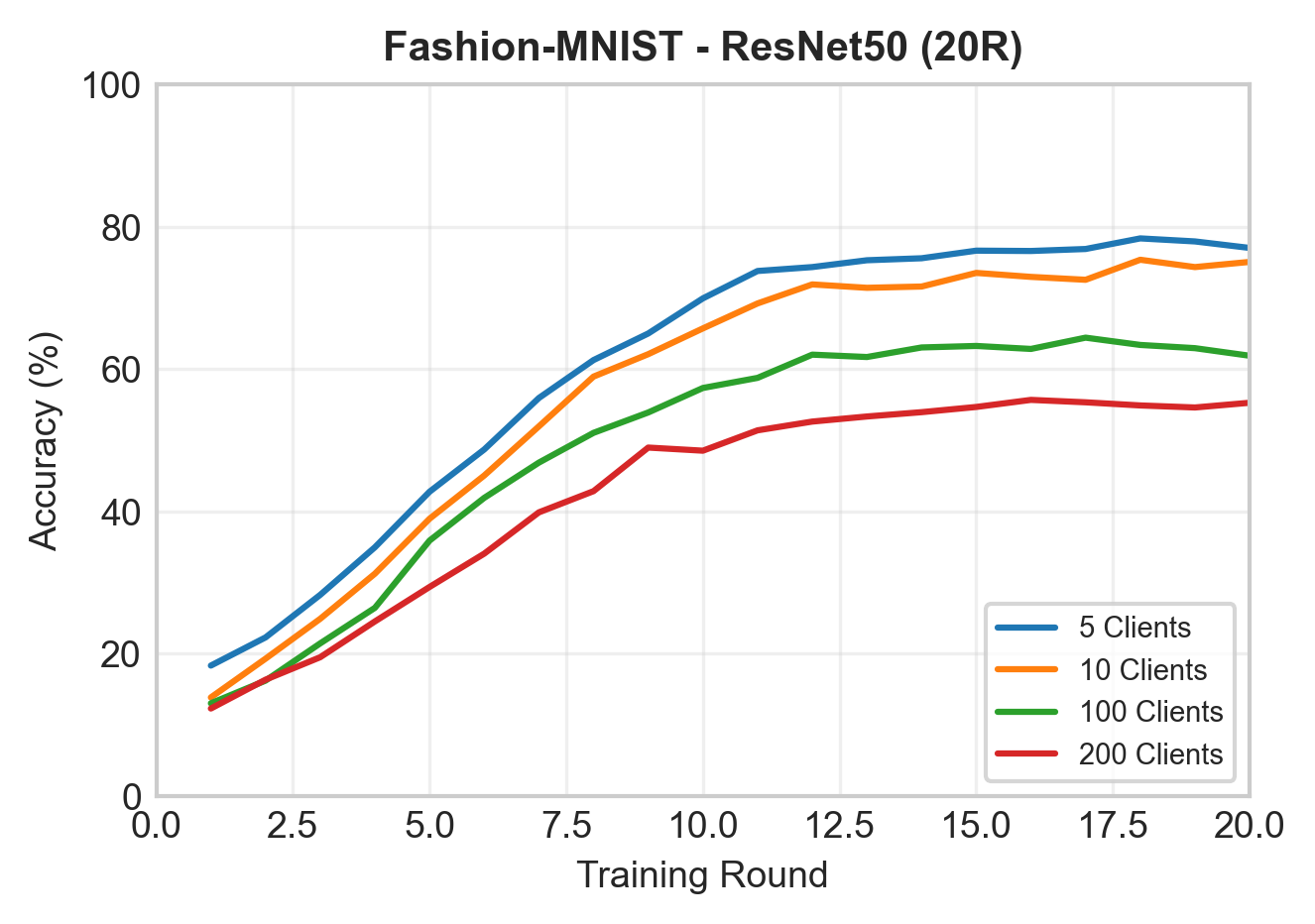}\end{subfigure}
\begin{subfigure}{0.23\textwidth}\includegraphics[width=\textwidth]{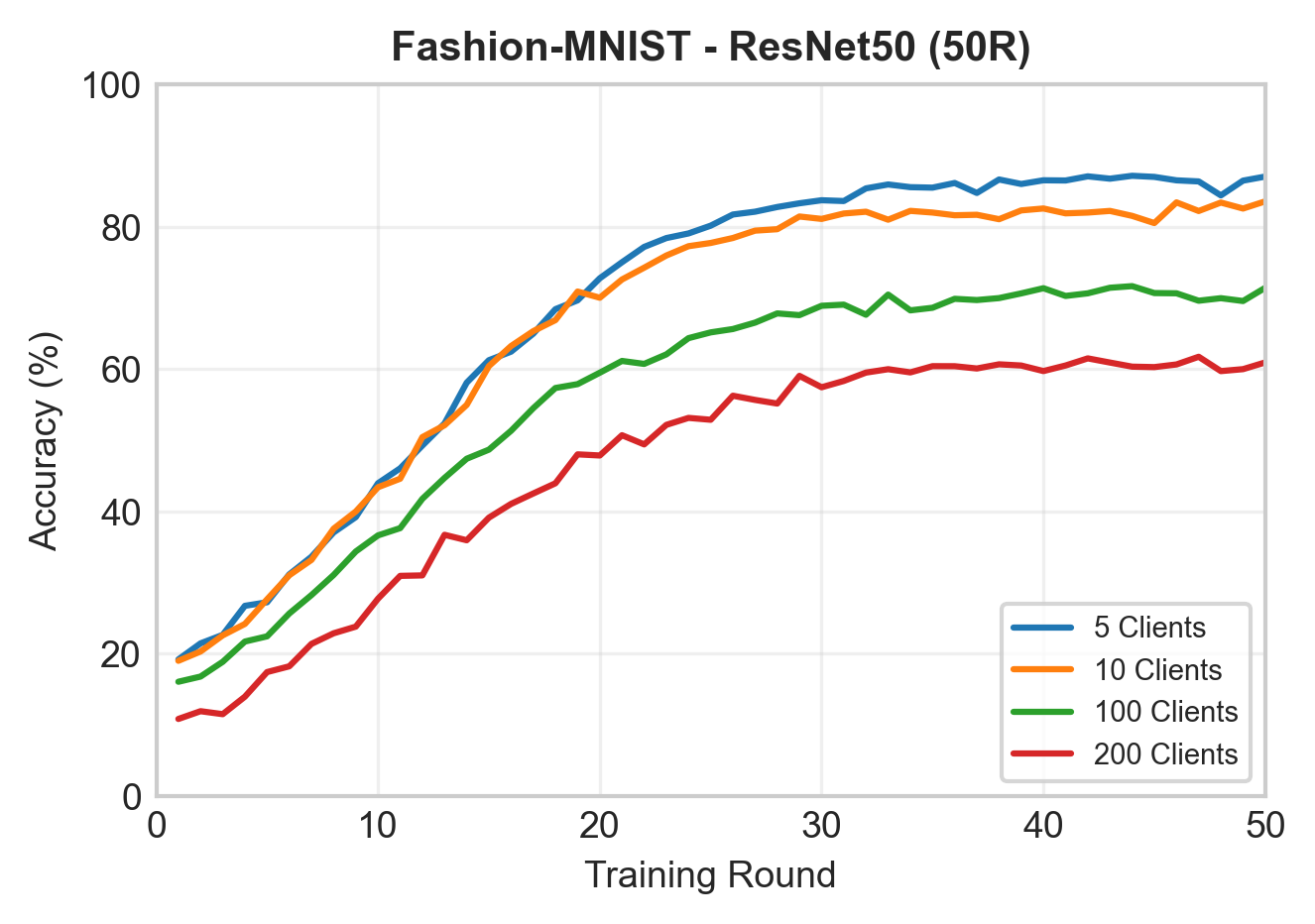}\end{subfigure}
\begin{subfigure}{0.23\textwidth}\includegraphics[width=\textwidth]{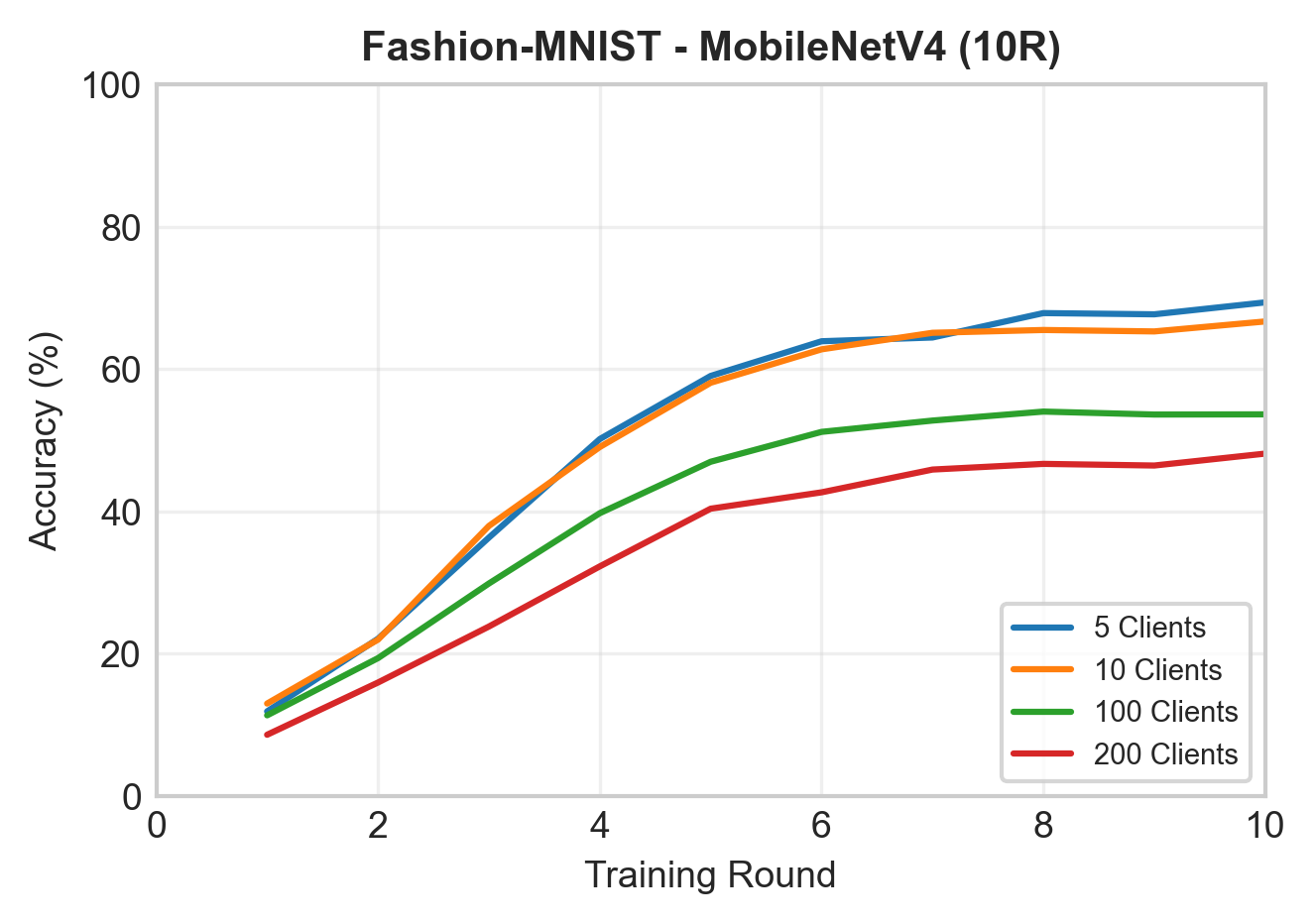}\end{subfigure}
\begin{subfigure}{0.23\textwidth}\includegraphics[width=\textwidth]{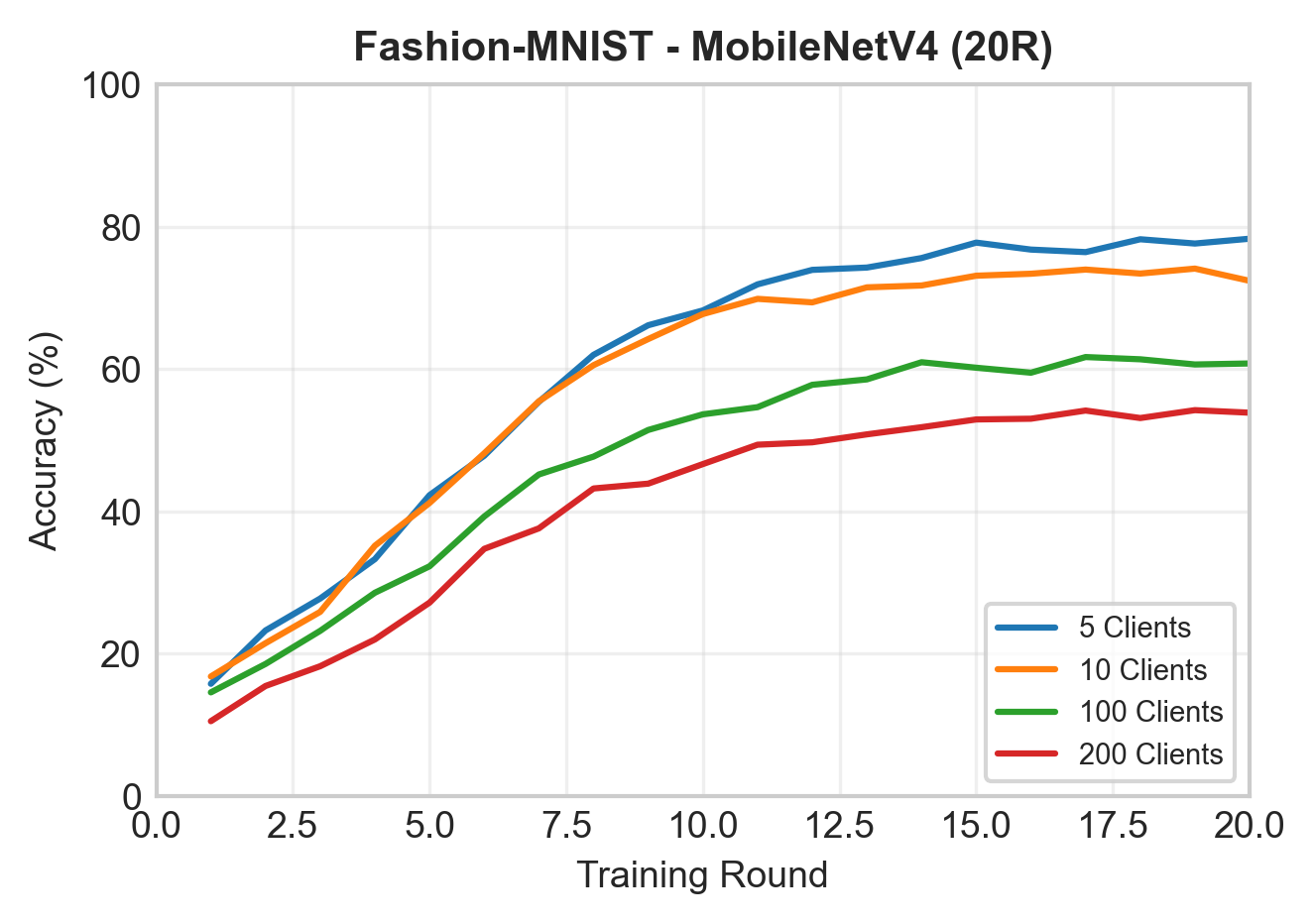}\end{subfigure}
\begin{subfigure}{0.23\textwidth}\includegraphics[width=\textwidth]{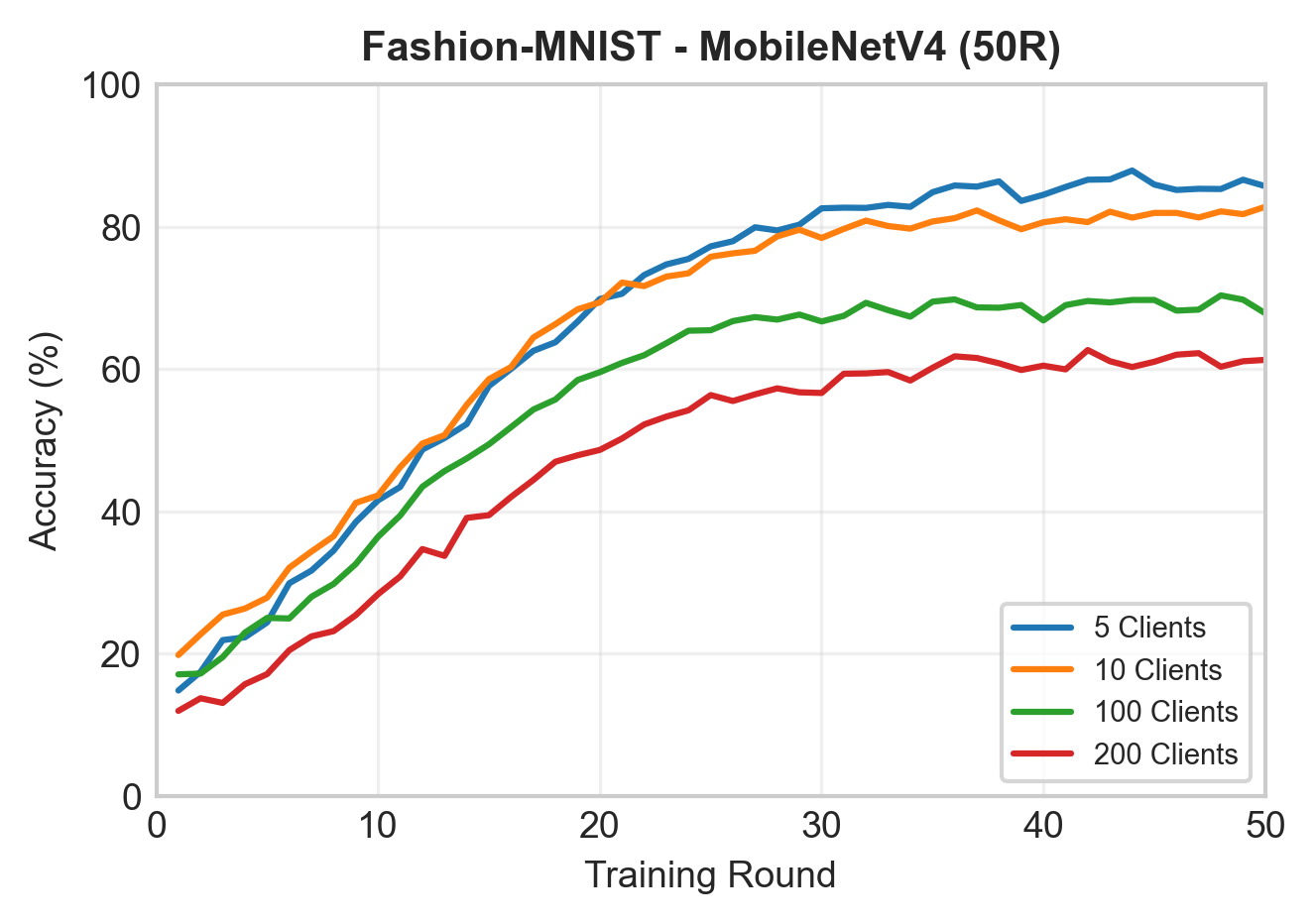}\end{subfigure}
\begin{subfigure}{0.23\textwidth}\includegraphics[width=\textwidth]{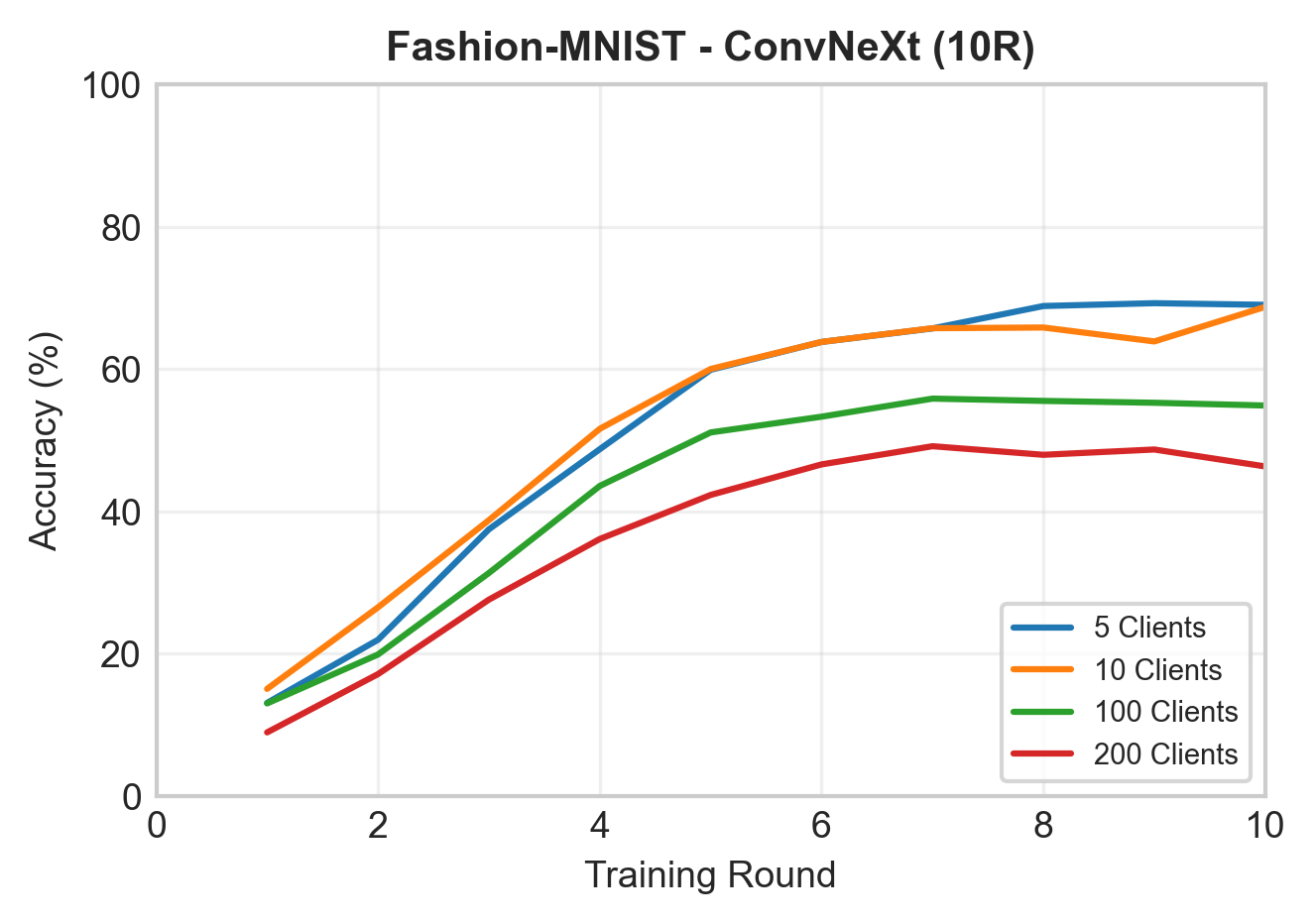}\end{subfigure}
\begin{subfigure}{0.23\textwidth}\includegraphics[width=\textwidth]{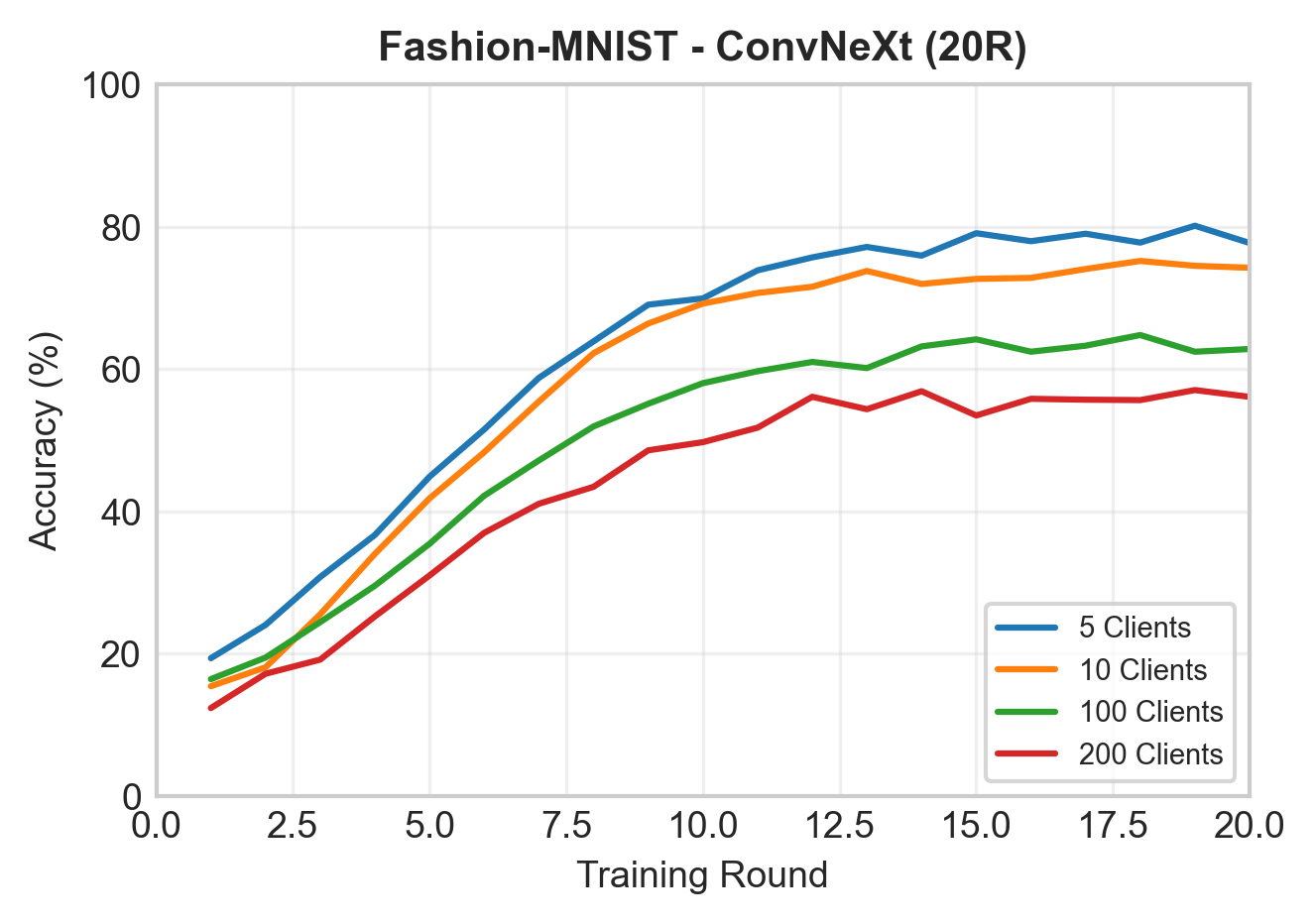}\end{subfigure}
\begin{subfigure}{0.23\textwidth}\includegraphics[width=\textwidth]{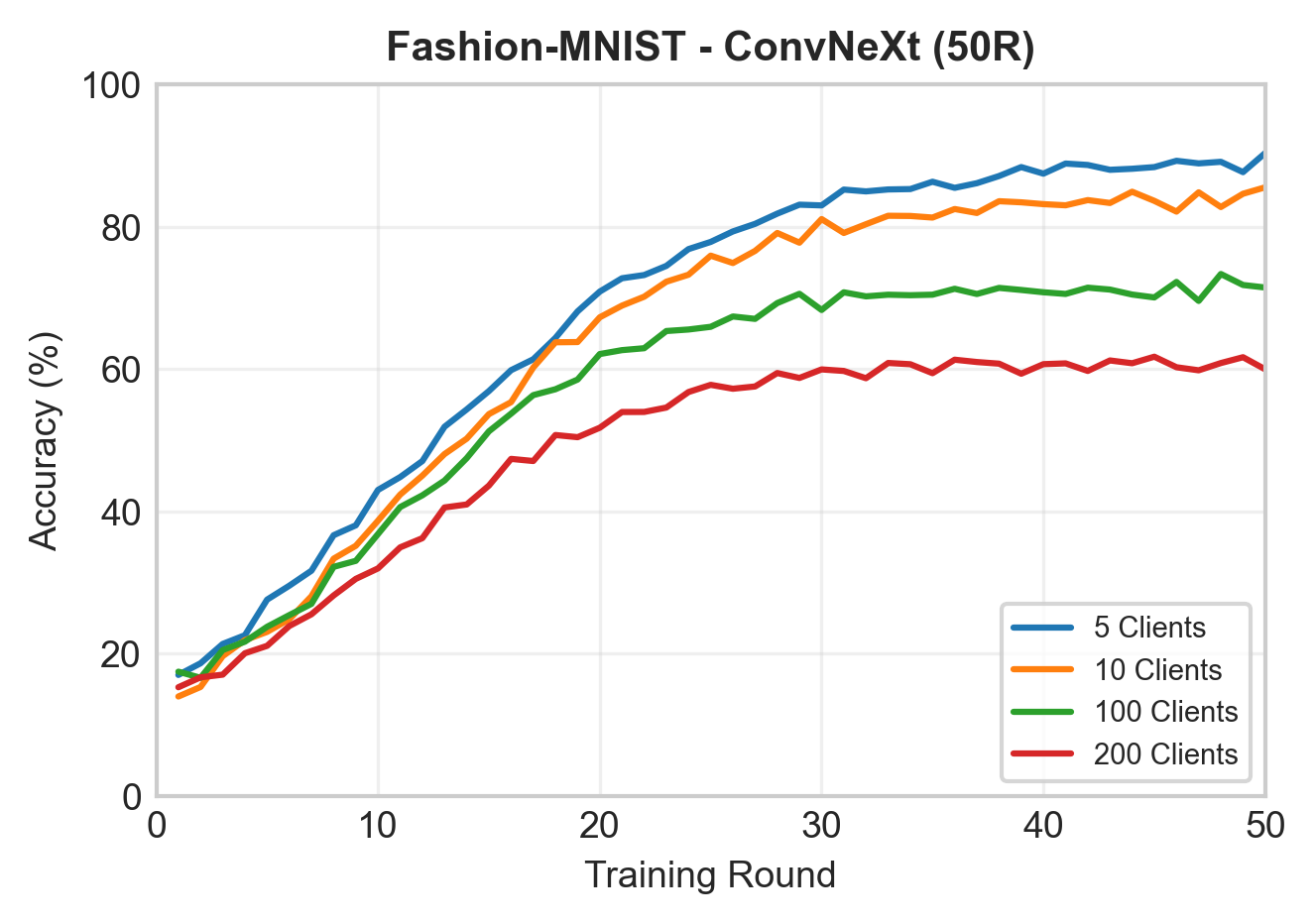}\end{subfigure}
\caption{\textbf{Fashion-MNIST Accuracy Convergence (10, 20, 50 Rounds).} Accuracy convergence comparison across architectures and training rounds for the Fashion-MNIST dataset. The figure shows CNN (row 1), ResNet50 (row 2), MobileNetV4 (row 3), and ConvNeXt (row 4) performance at 10, 20, and 50 rounds. This dataset is more structurally complex than MNIST: ConvNeXt consistently achieves the highest accuracy ($\sim$94\% at 50 rounds), followed by ResNet50 and MobileNetV4, while CNN plateaus around 90\%. The performance gap widens with extended training, demonstrating the benefit of deeper architectures for moderately complex visual classification tasks. All models show stable convergence across diverse client counts.}
\label{fig:fmnist_acc_additional}
\end{minipage}
\end{figure*}

\begin{figure*}[!htbp]
\centering
\begin{minipage}{\textwidth}
\centering
\begin{subfigure}{0.23\textwidth}\includegraphics[width=\textwidth]{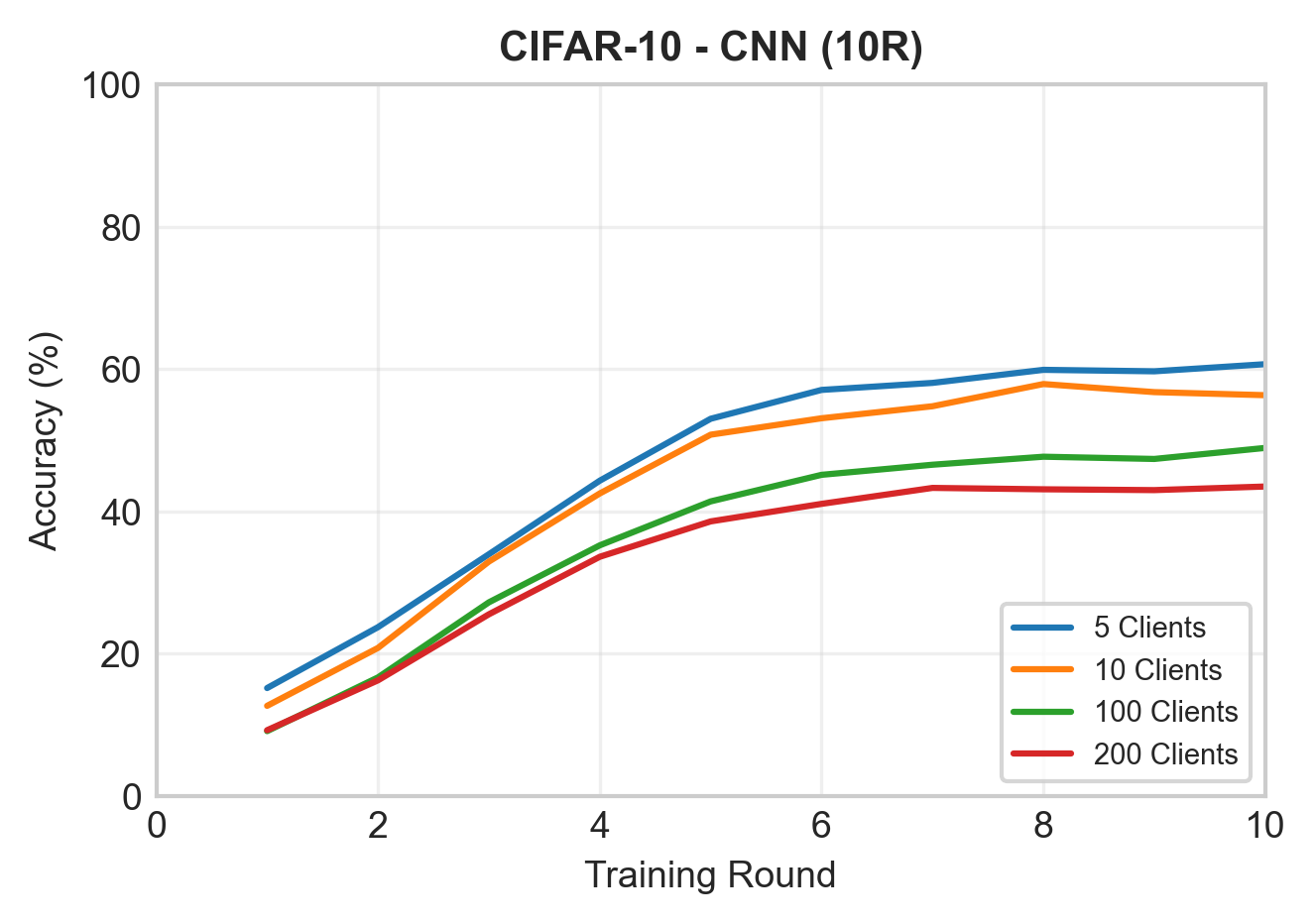}\end{subfigure}
\begin{subfigure}{0.23\textwidth}\includegraphics[width=\textwidth]{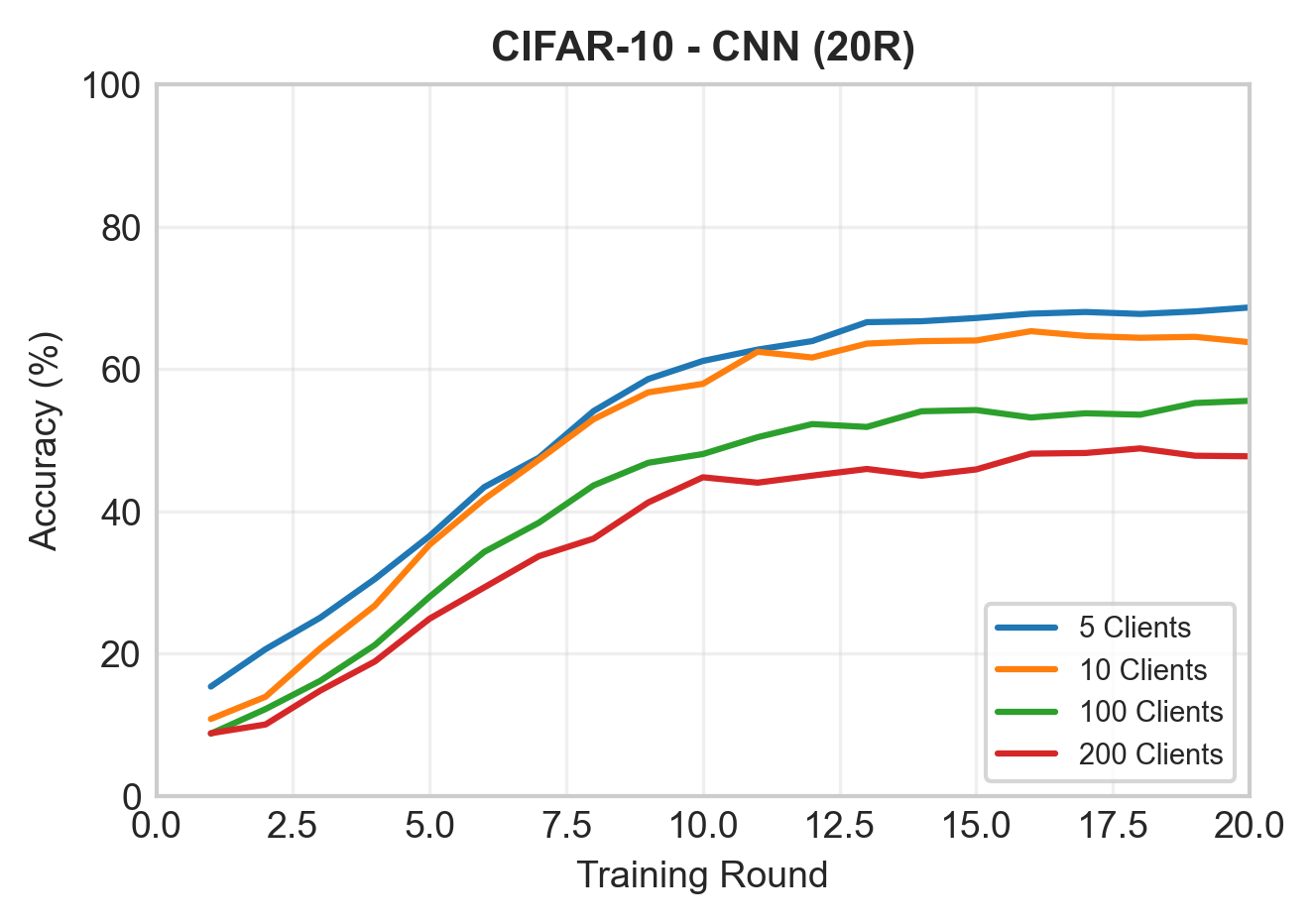}\end{subfigure}
\begin{subfigure}{0.23\textwidth}\includegraphics[width=\textwidth]{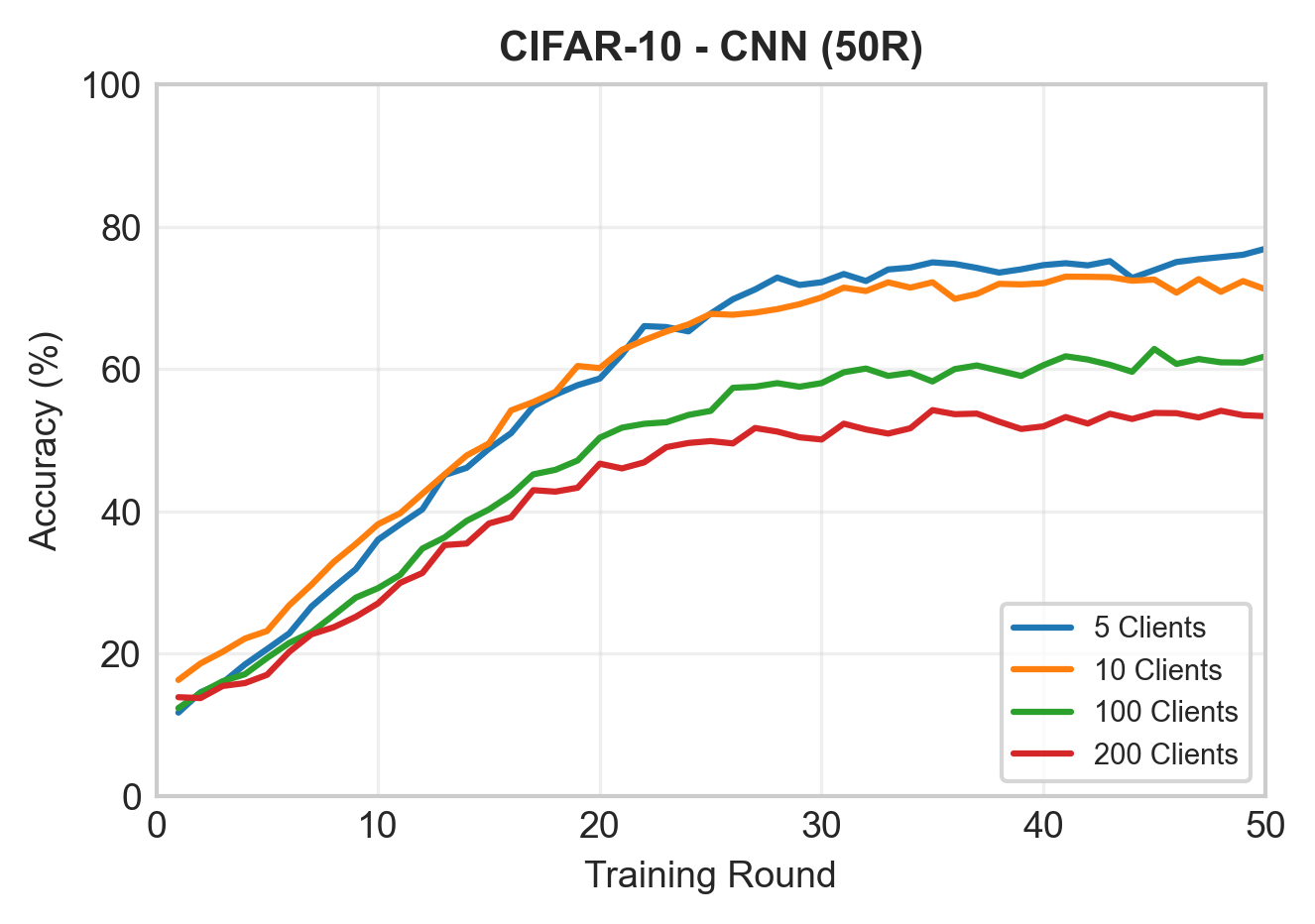}\end{subfigure}
\begin{subfigure}{0.23\textwidth}\includegraphics[width=\textwidth]{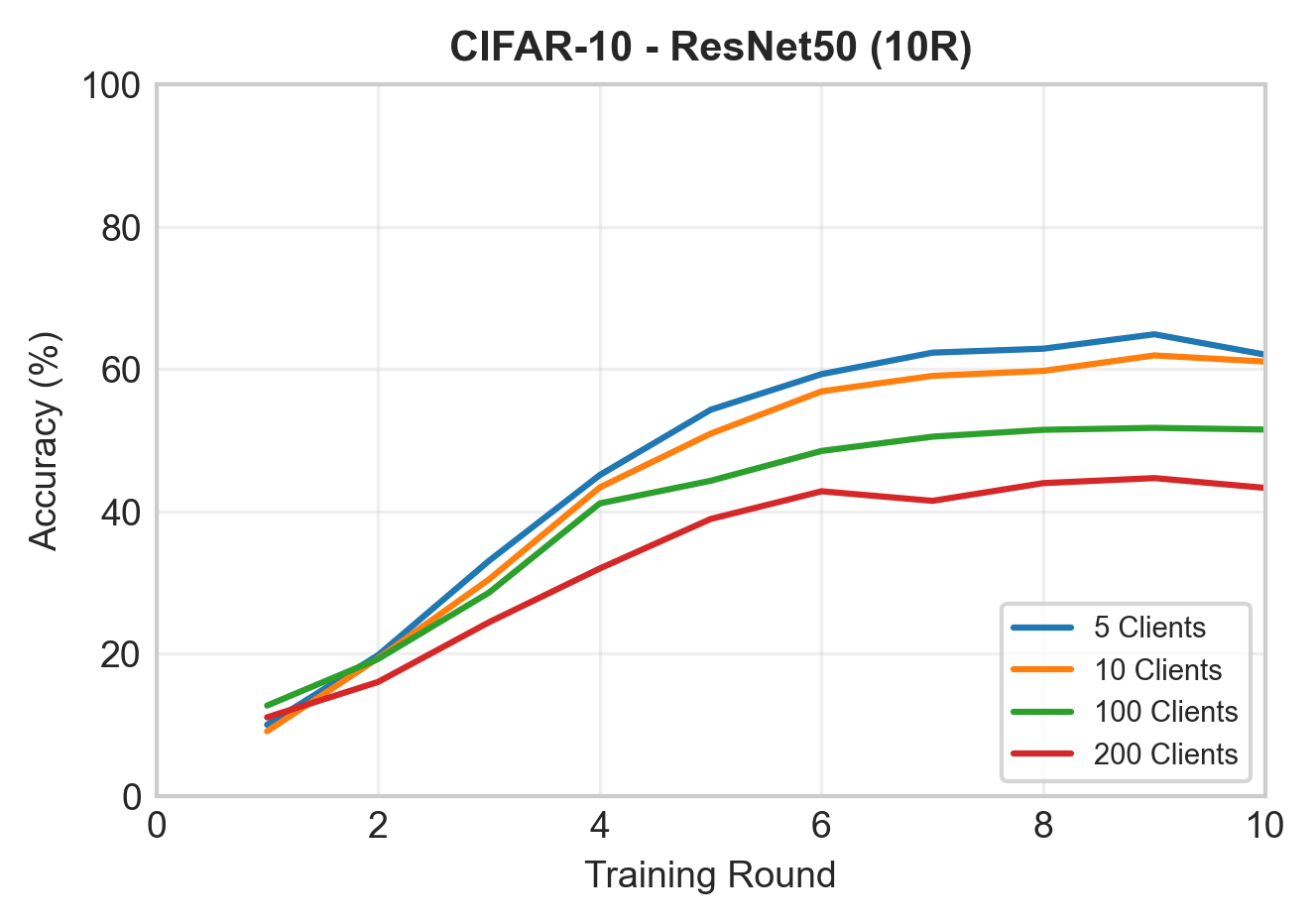}\end{subfigure}
\begin{subfigure}{0.23\textwidth}\includegraphics[width=\textwidth]{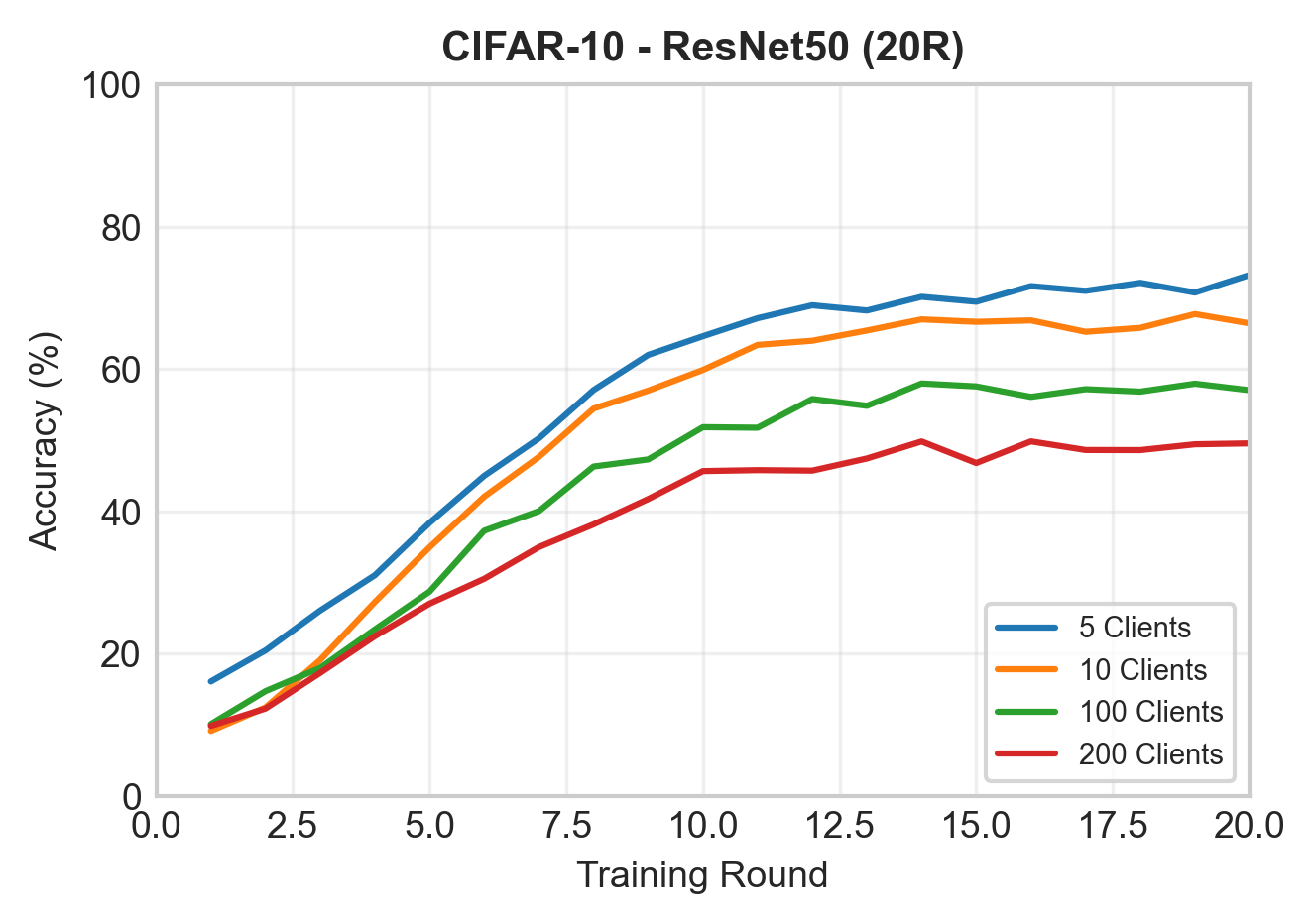}\end{subfigure}
\begin{subfigure}{0.23\textwidth}\includegraphics[width=\textwidth]{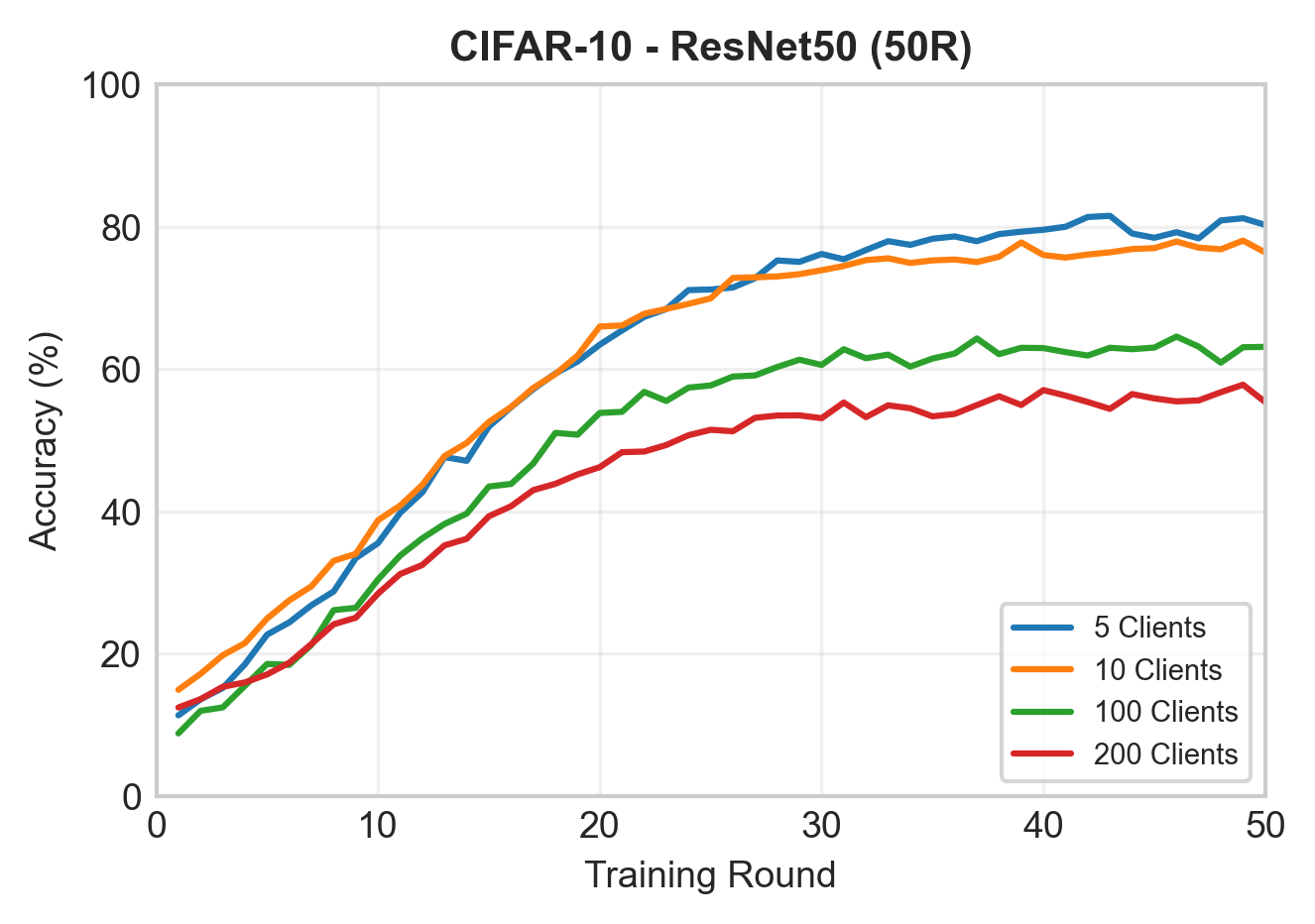}\end{subfigure}
\begin{subfigure}{0.23\textwidth}\includegraphics[width=\textwidth]{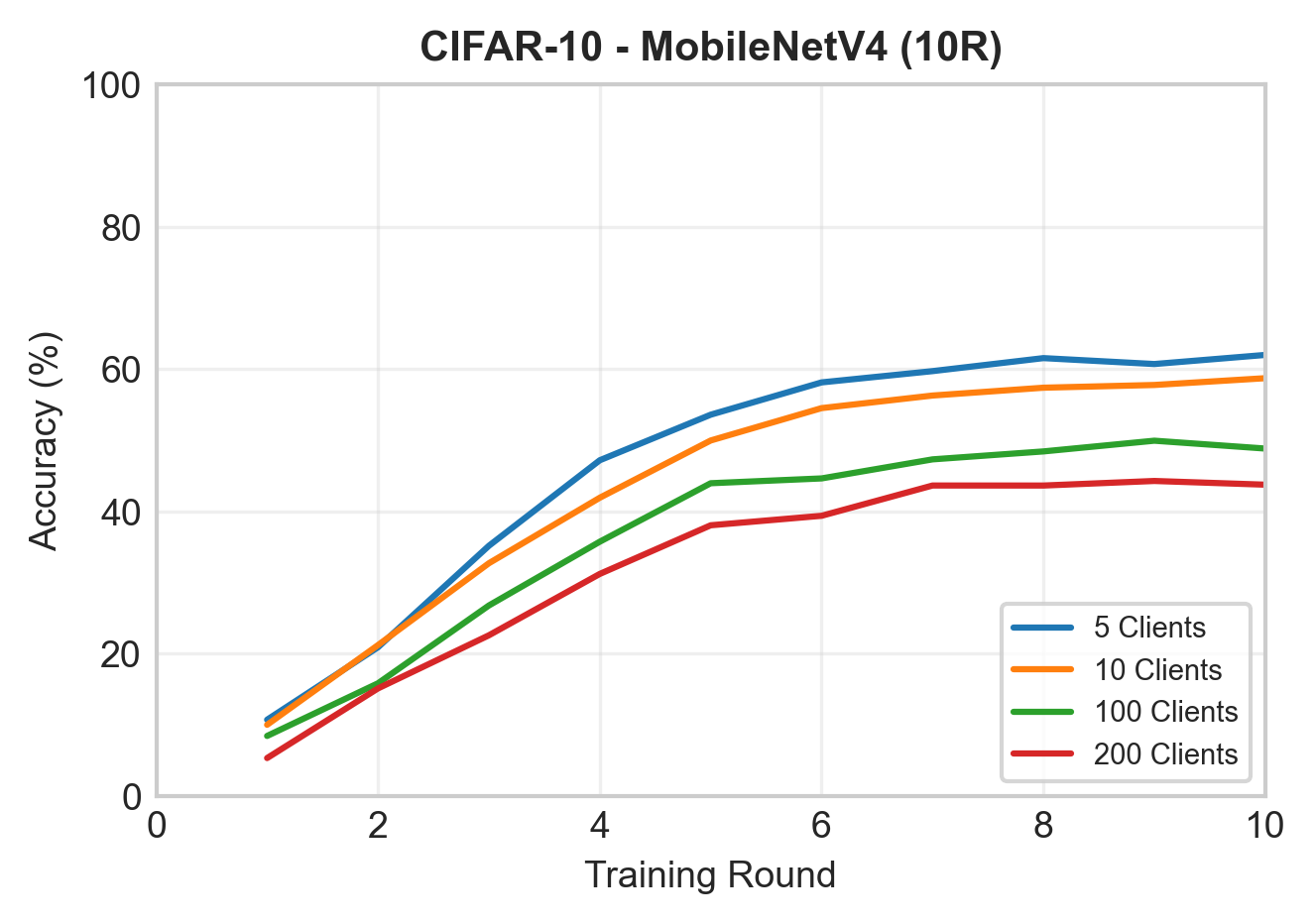}\end{subfigure}
\begin{subfigure}{0.23\textwidth}\includegraphics[width=\textwidth]{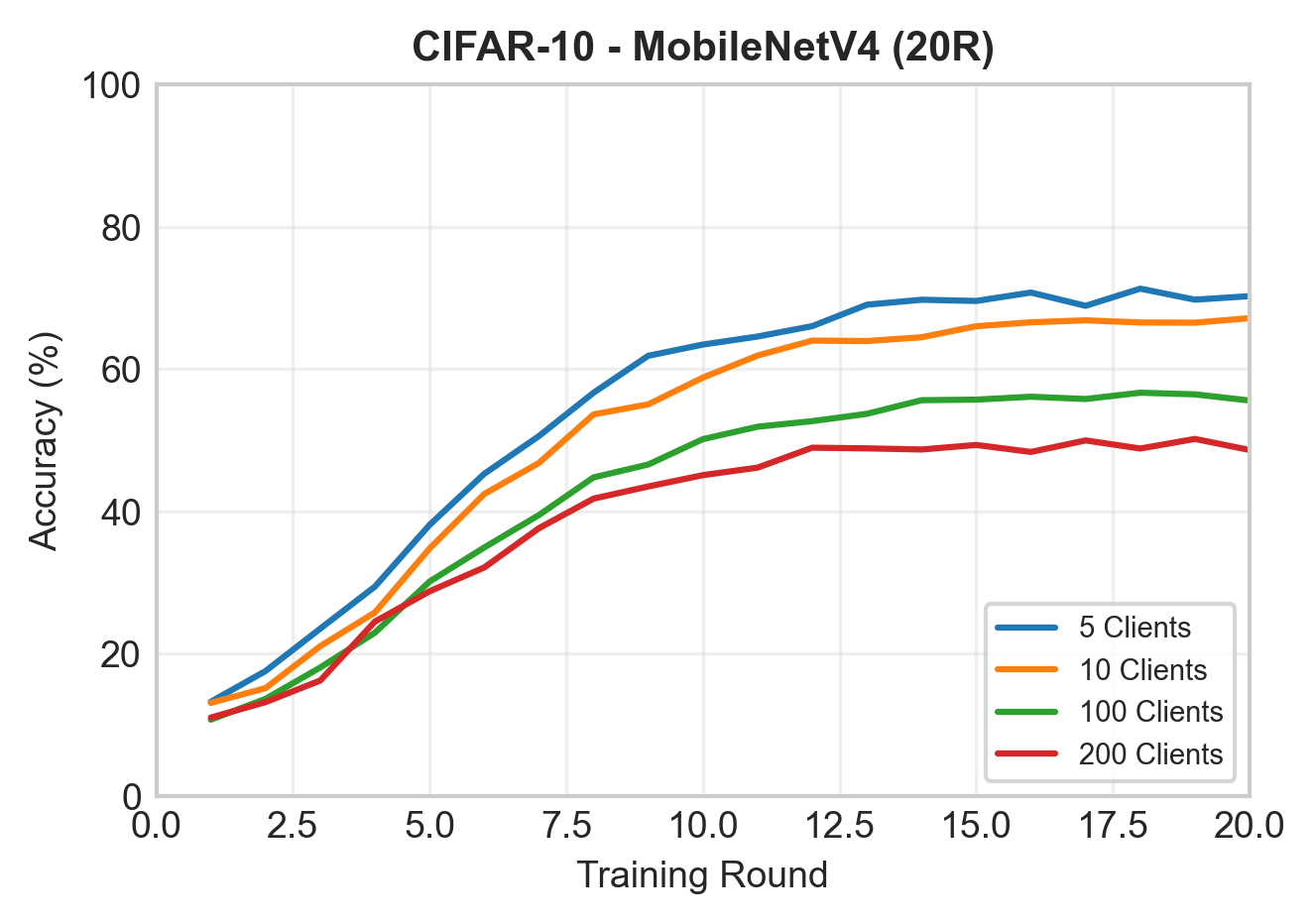}\end{subfigure}
\begin{subfigure}{0.23\textwidth}\includegraphics[width=\textwidth]{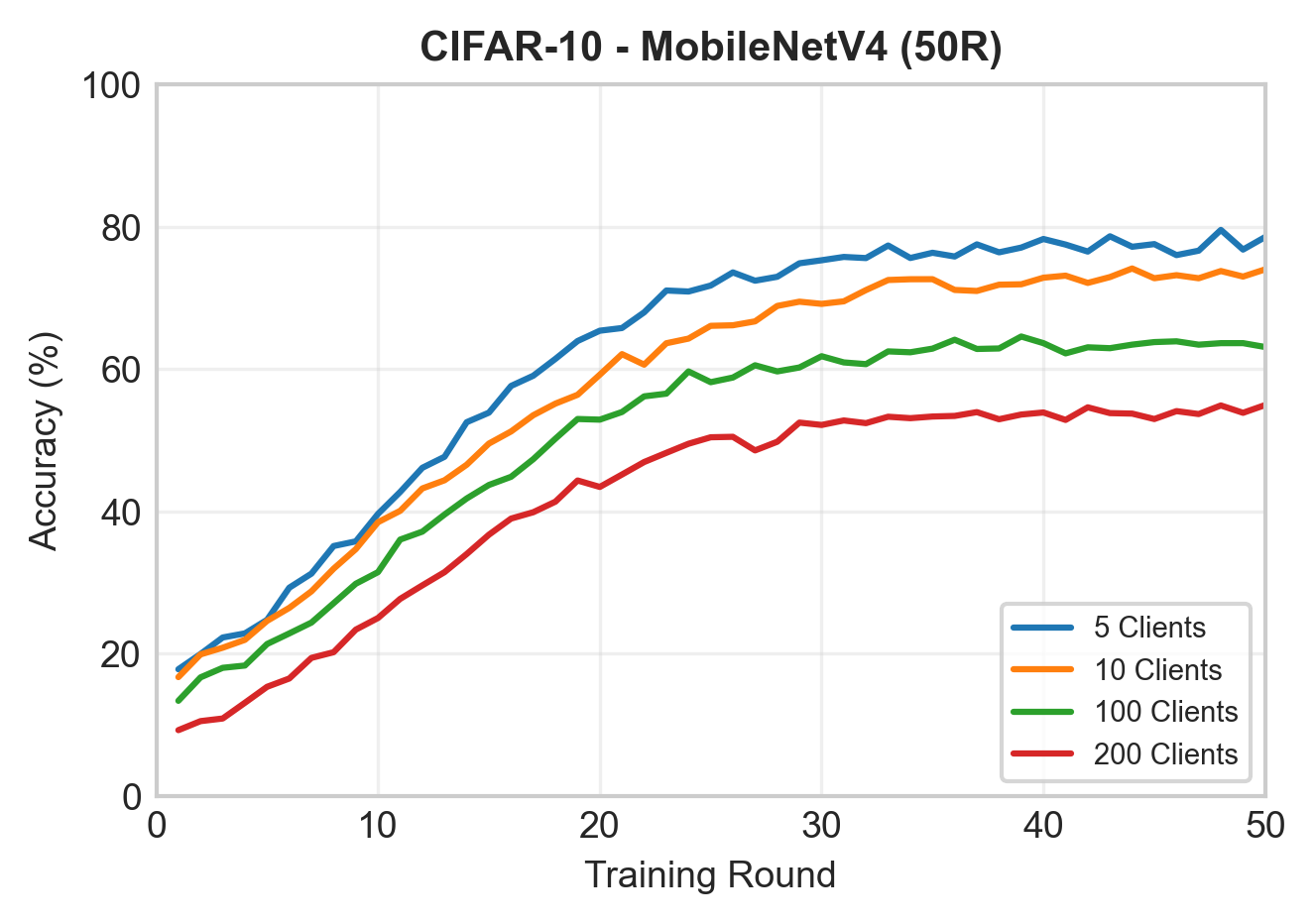}\end{subfigure}
\begin{subfigure}{0.23\textwidth}\includegraphics[width=\textwidth]{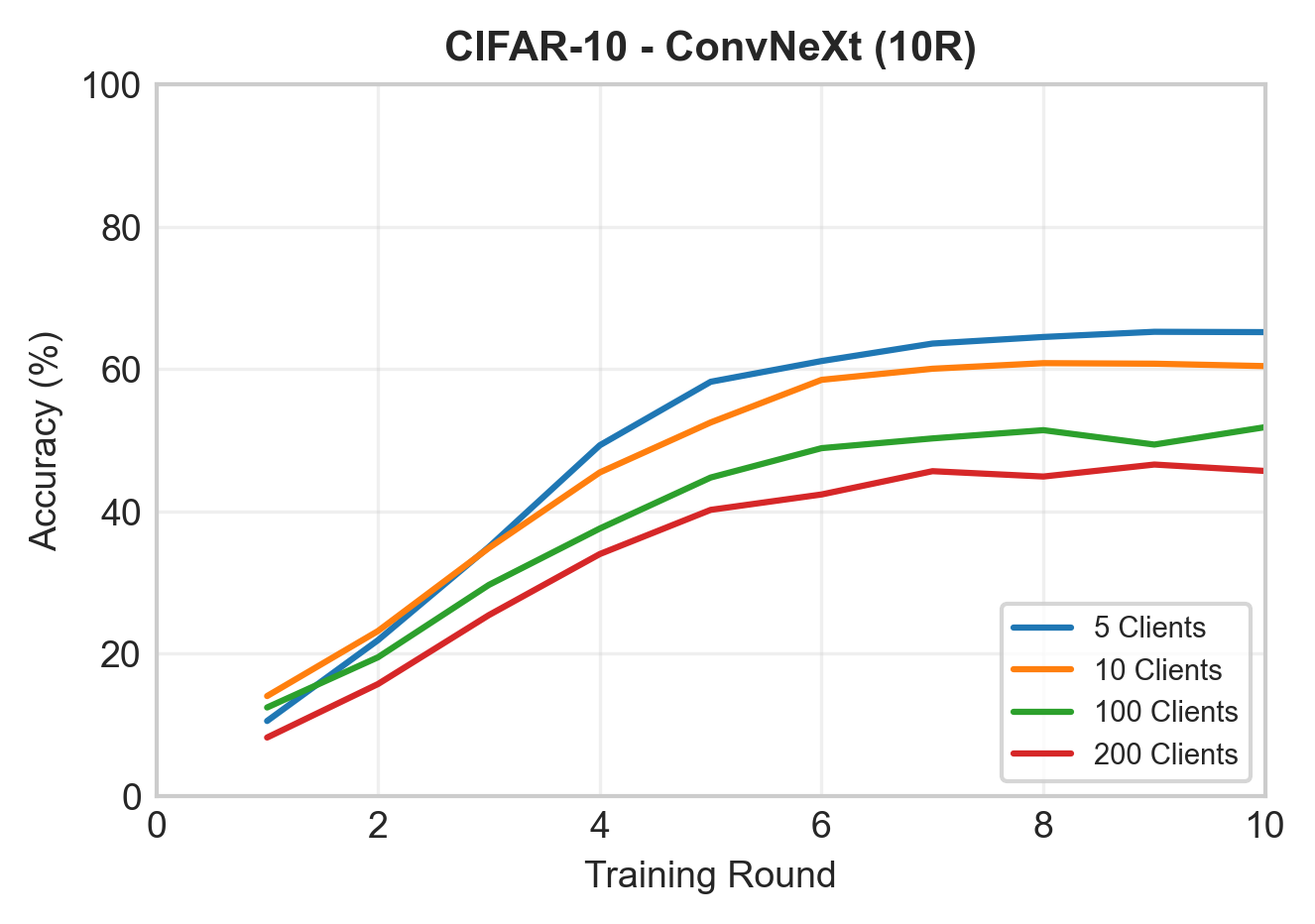}\end{subfigure}
\begin{subfigure}{0.23\textwidth}\includegraphics[width=\textwidth]{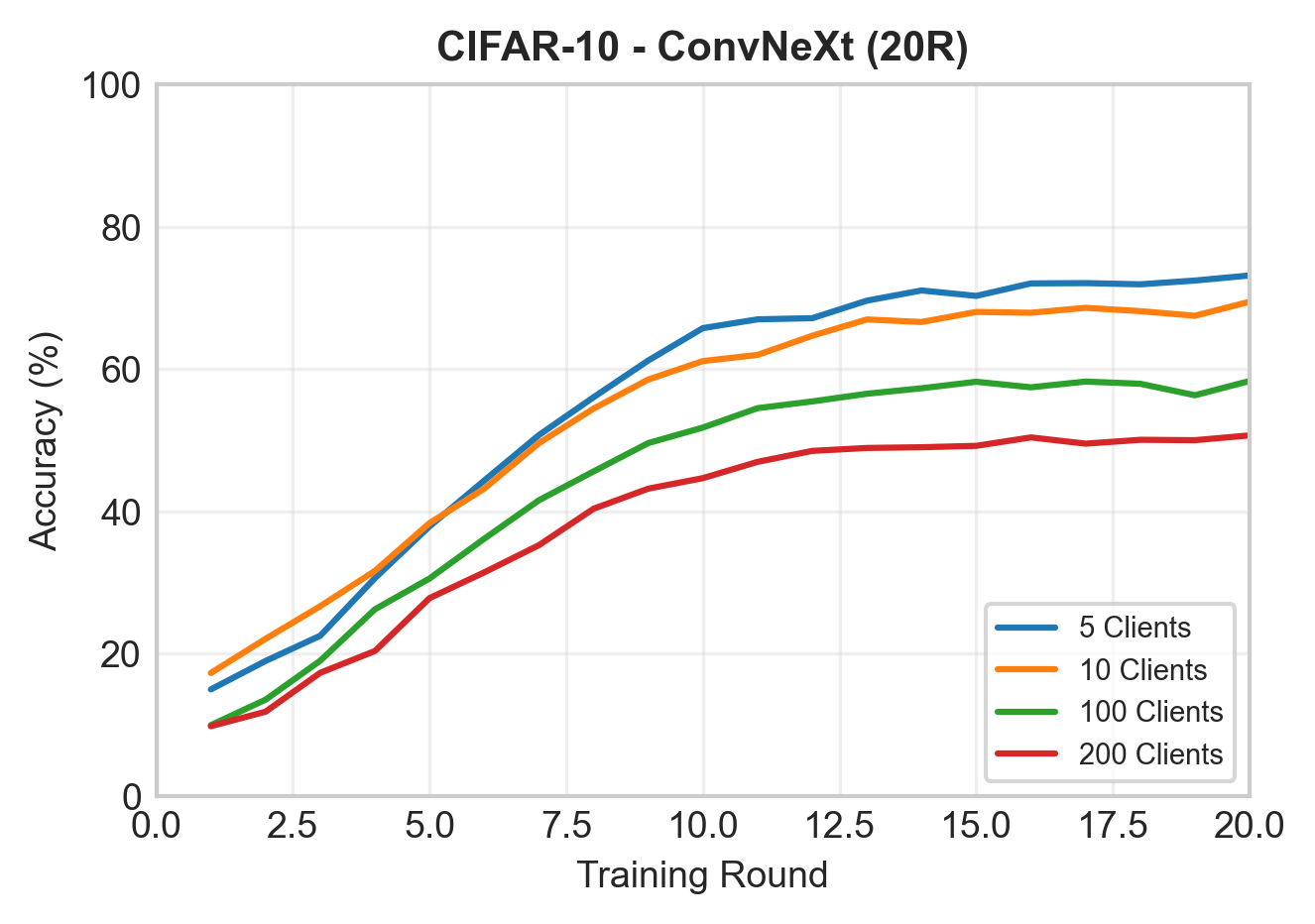}\end{subfigure}
\begin{subfigure}{0.23\textwidth}\includegraphics[width=\textwidth]{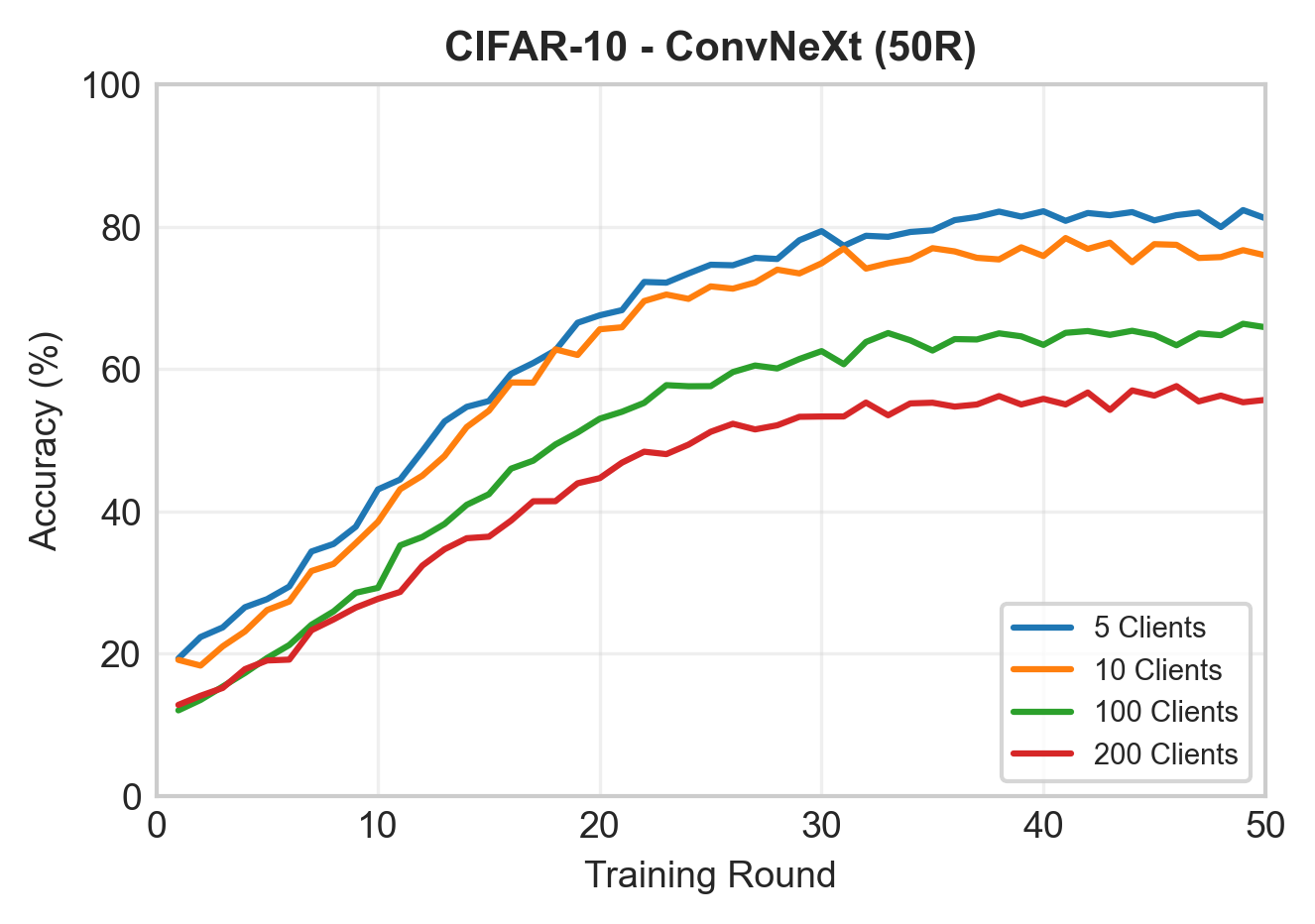}\end{subfigure}
\caption{\textbf{CIFAR-10 Accuracy Convergence (10, 20, 50 Rounds).} Convergence trends for CIFAR-10 dataset shows performance differentiation among different architectures. The figure illustrates that 10-round training is insufficient for effective feature learning, particularly for shallow CNN. Extended training to 50 rounds enables deep architectures (ResNet50, ConvNeXt) to achieve substantially higher accuracy ($\sim$86\%) compared to lighter models. Varying rounds (10, 20, 50 rounds) clearly demonstrates the importance of adequate training duration for complex datasets, while the neural network architectures (CNN, ResNet50, MobileNetV4, ConvNeXt) emphasizes the necessity of capability aware split point selection to leverage deep model capacity for RGB data processing.}
\label{fig:cifar10_acc_additional}
\end{minipage}

\begin{minipage}{\textwidth}
\centering
\begin{subfigure}{0.23\textwidth}\includegraphics[width=\textwidth]{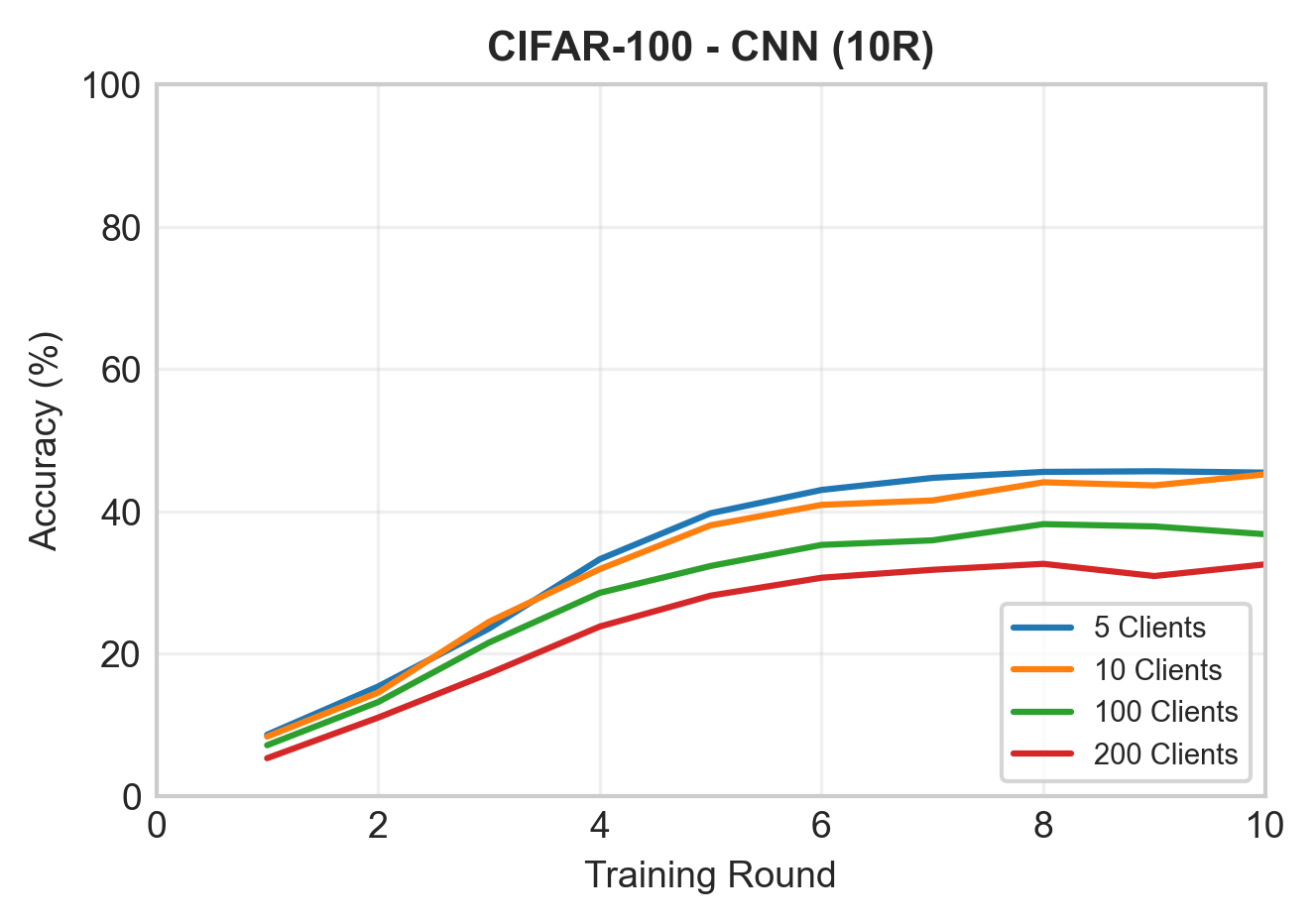}\end{subfigure}
\begin{subfigure}{0.23\textwidth}\includegraphics[width=\textwidth]{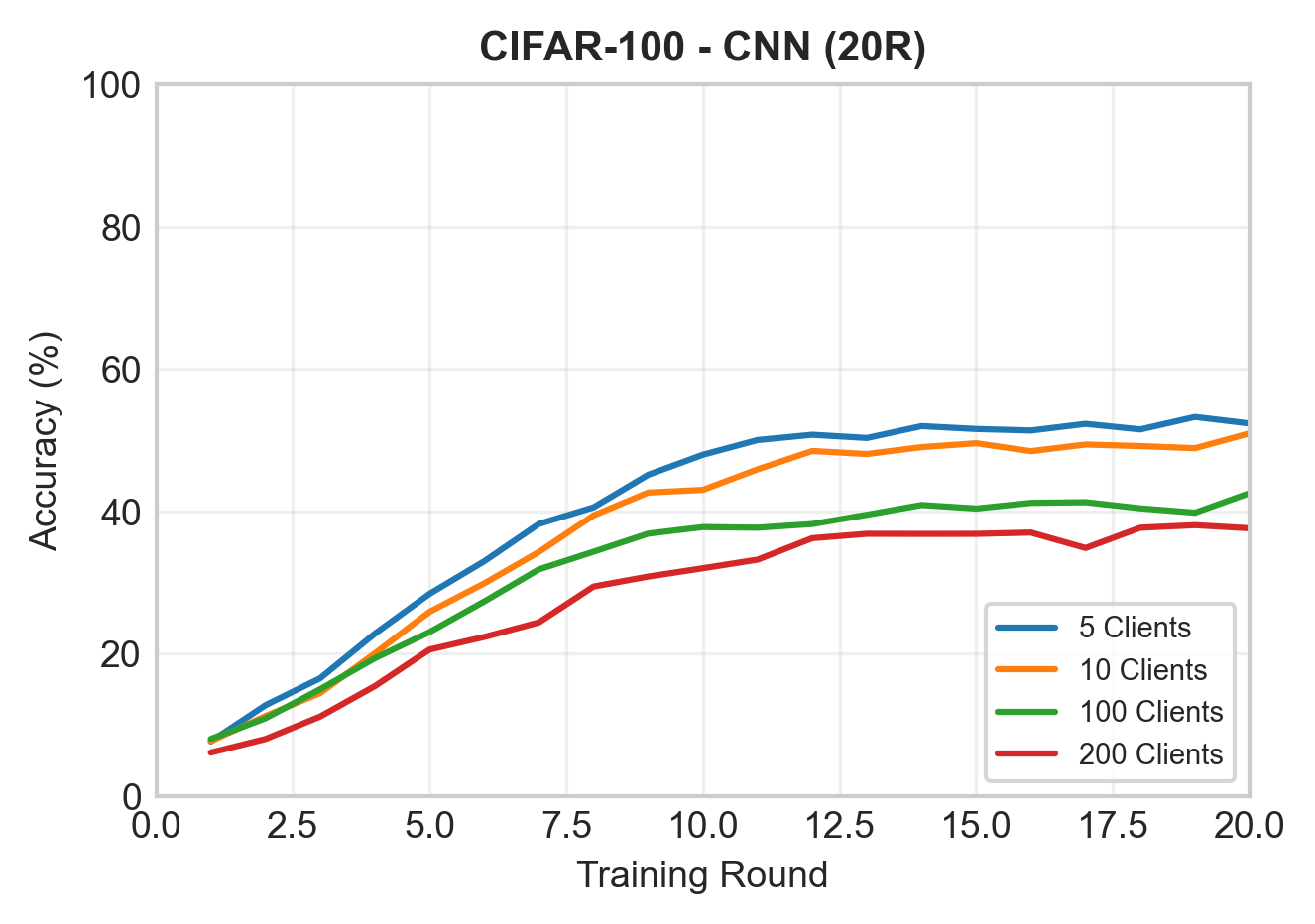}\end{subfigure}
\begin{subfigure}{0.23\textwidth}\includegraphics[width=\textwidth]{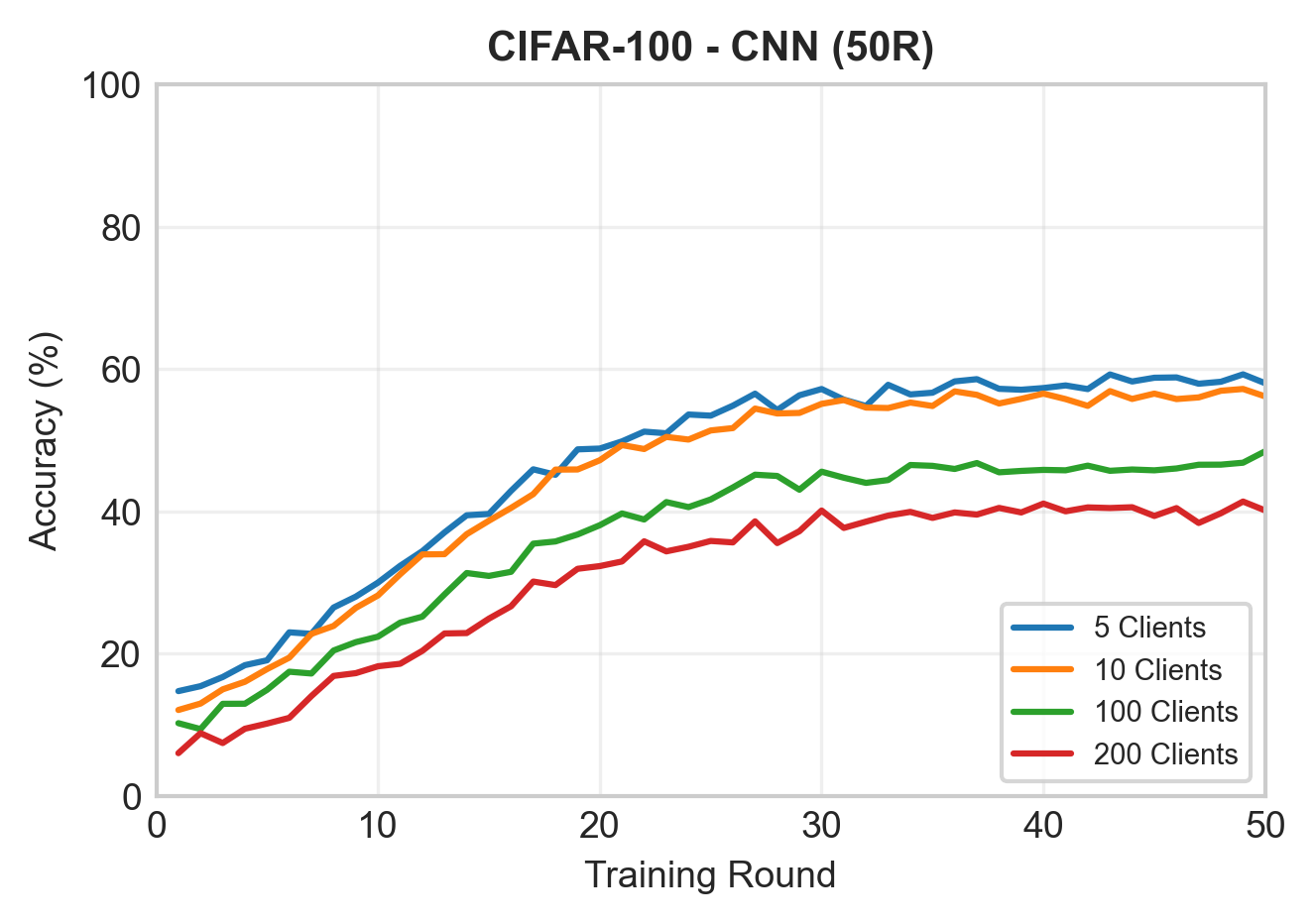}\end{subfigure}
\begin{subfigure}{0.23\textwidth}\includegraphics[width=\textwidth]{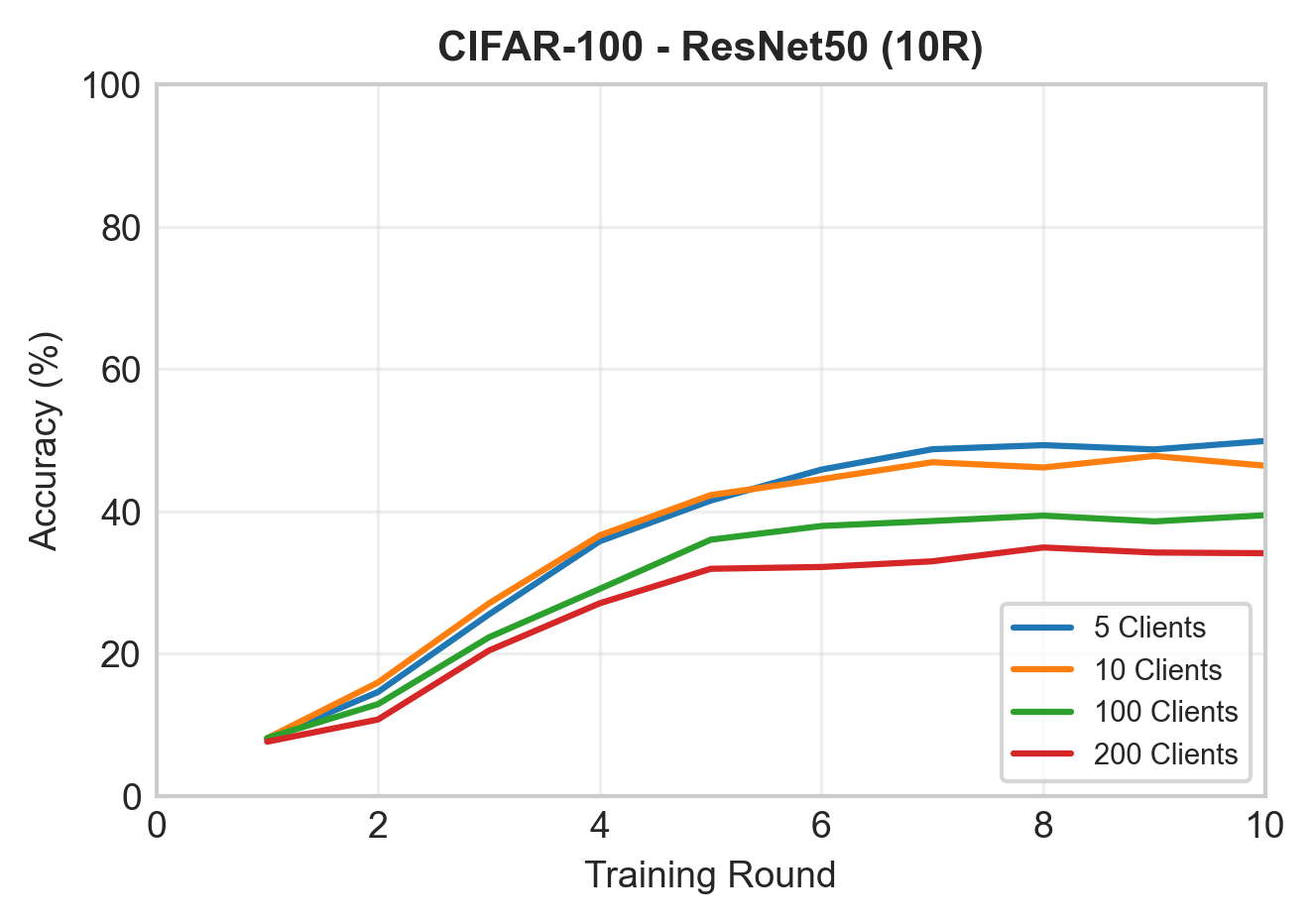}\end{subfigure}
\begin{subfigure}{0.23\textwidth}\includegraphics[width=\textwidth]{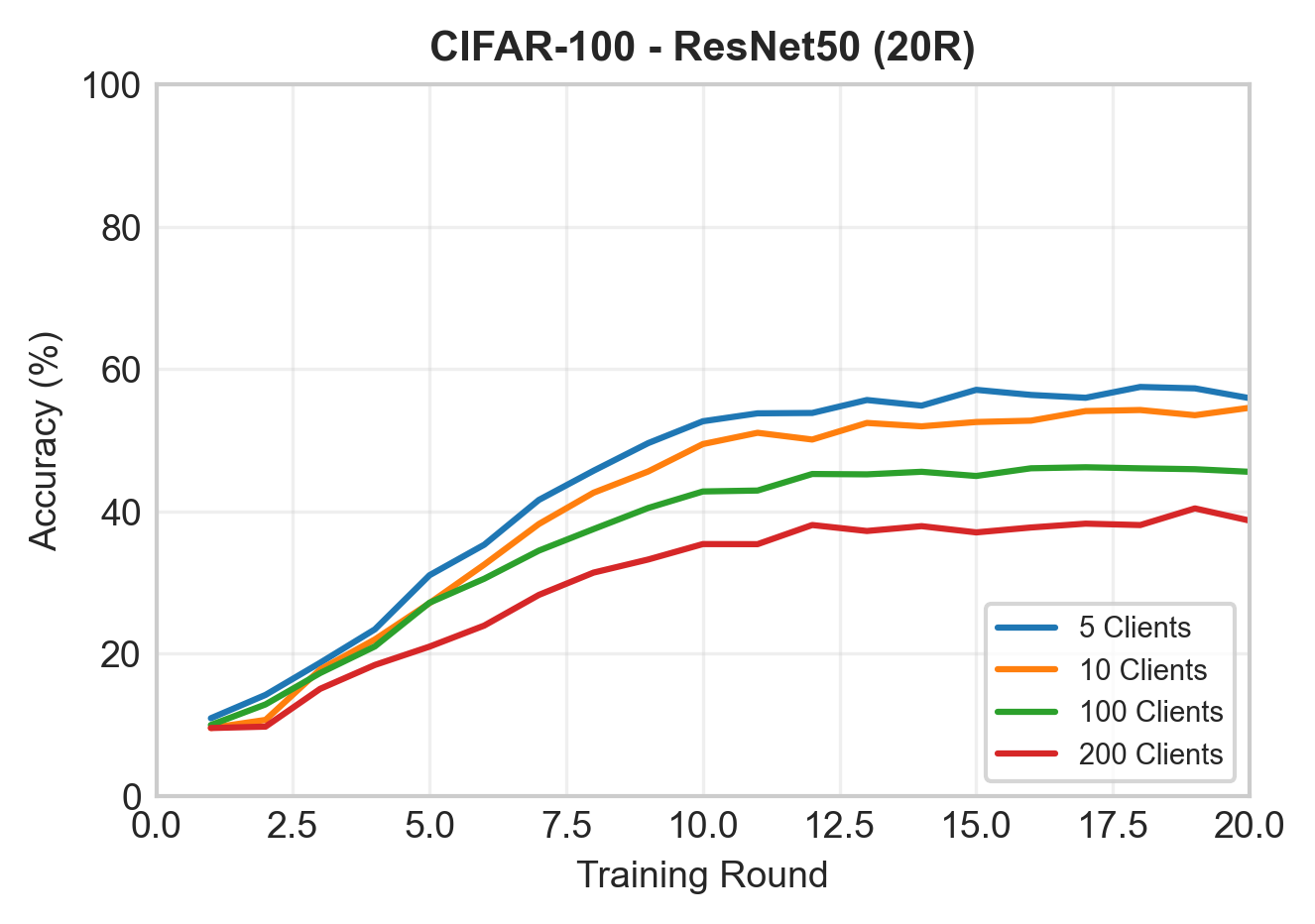}\end{subfigure}
\begin{subfigure}{0.23\textwidth}\includegraphics[width=\textwidth]{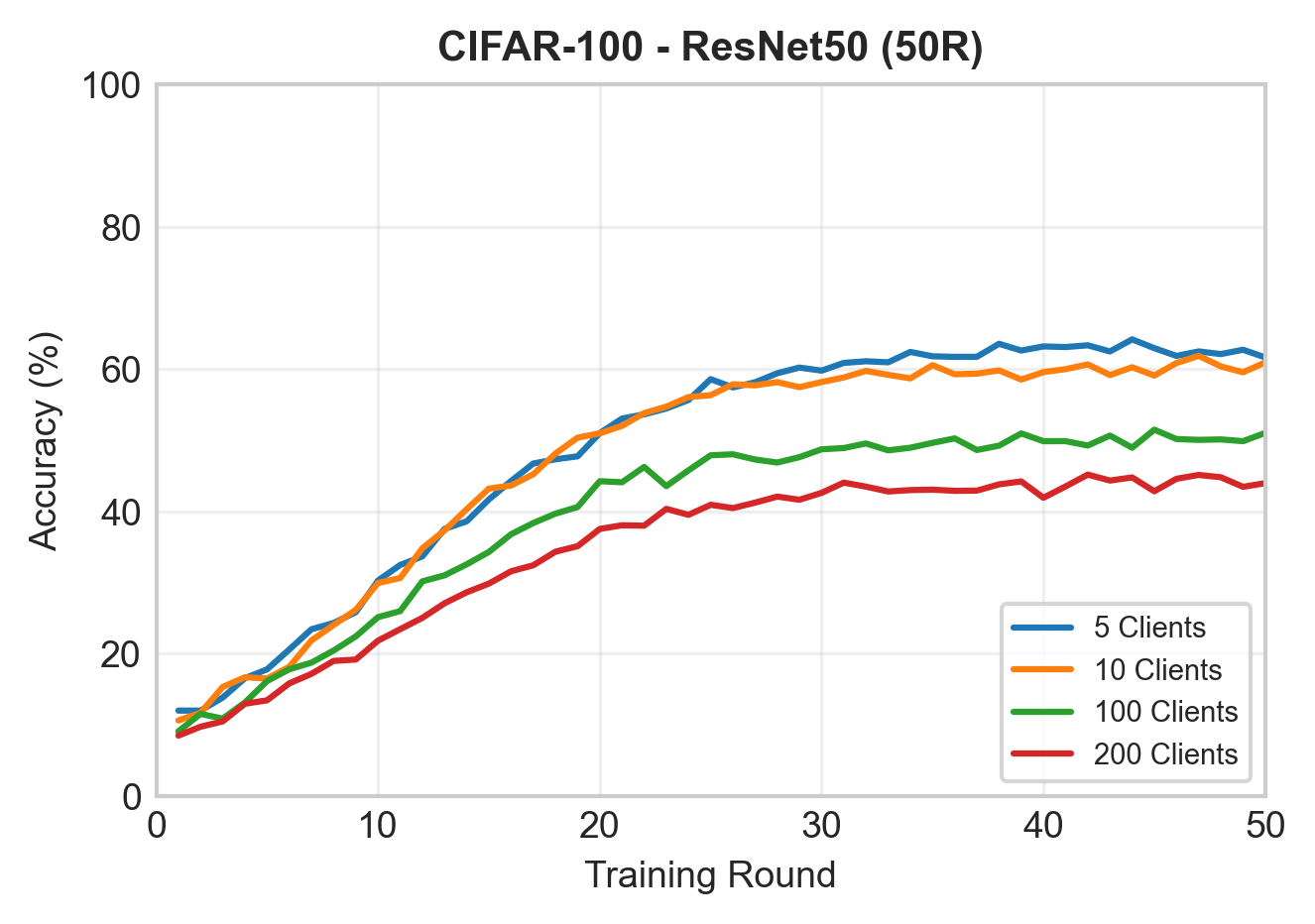}\end{subfigure}
\begin{subfigure}{0.23\textwidth}\includegraphics[width=\textwidth]{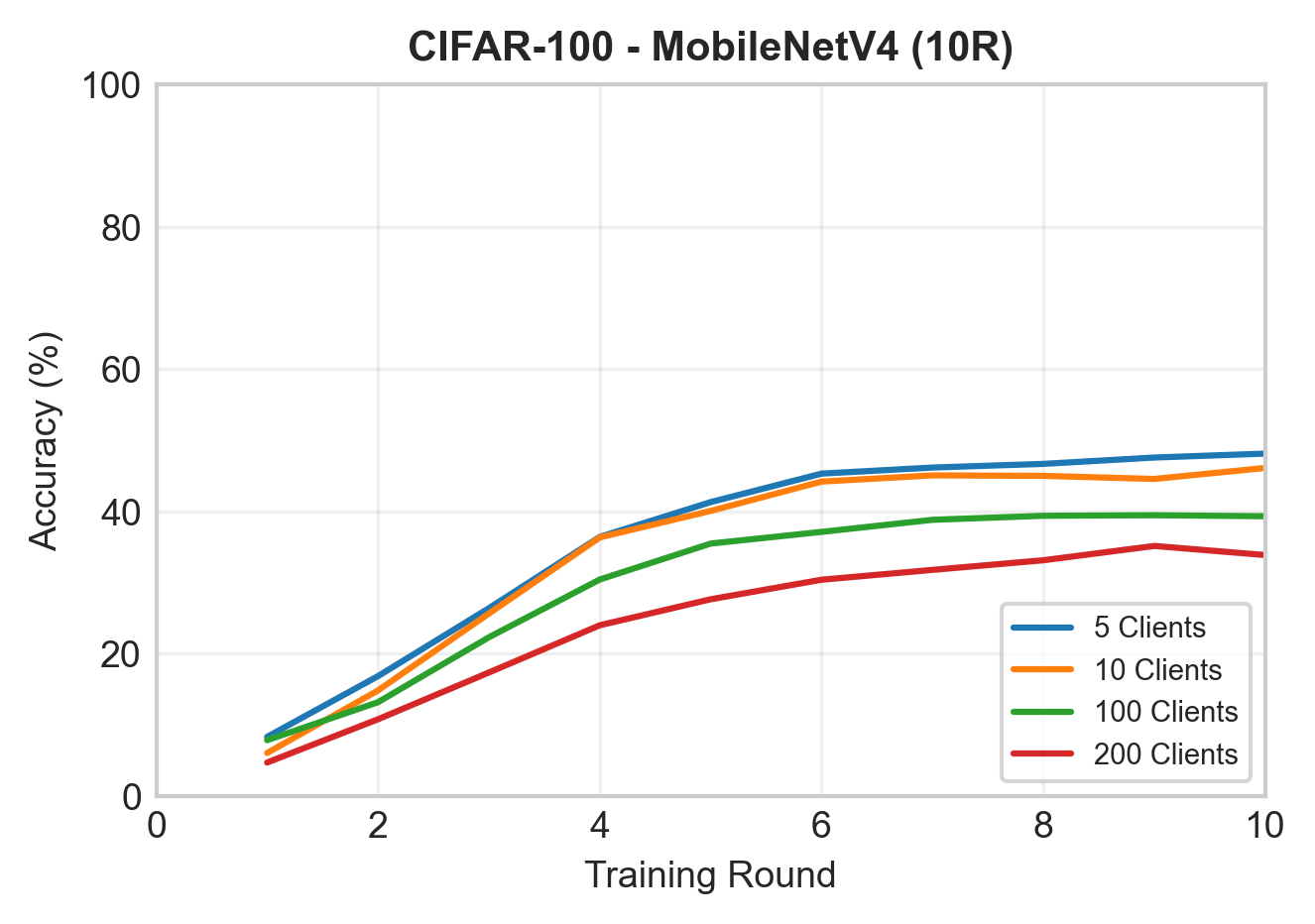}\end{subfigure}
\begin{subfigure}{0.23\textwidth}\includegraphics[width=\textwidth]{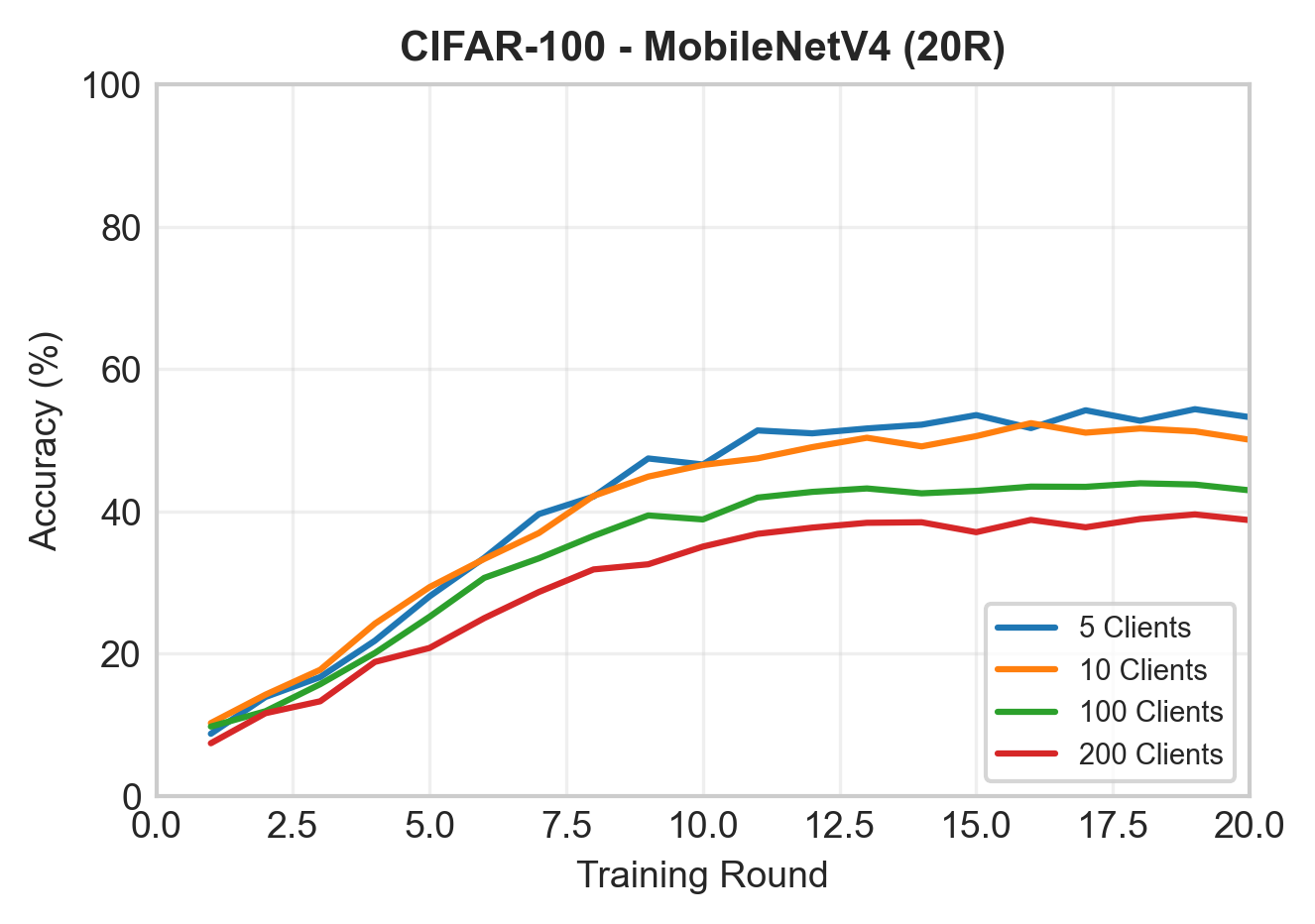}\end{subfigure}
\begin{subfigure}{0.23\textwidth}\includegraphics[width=\textwidth]{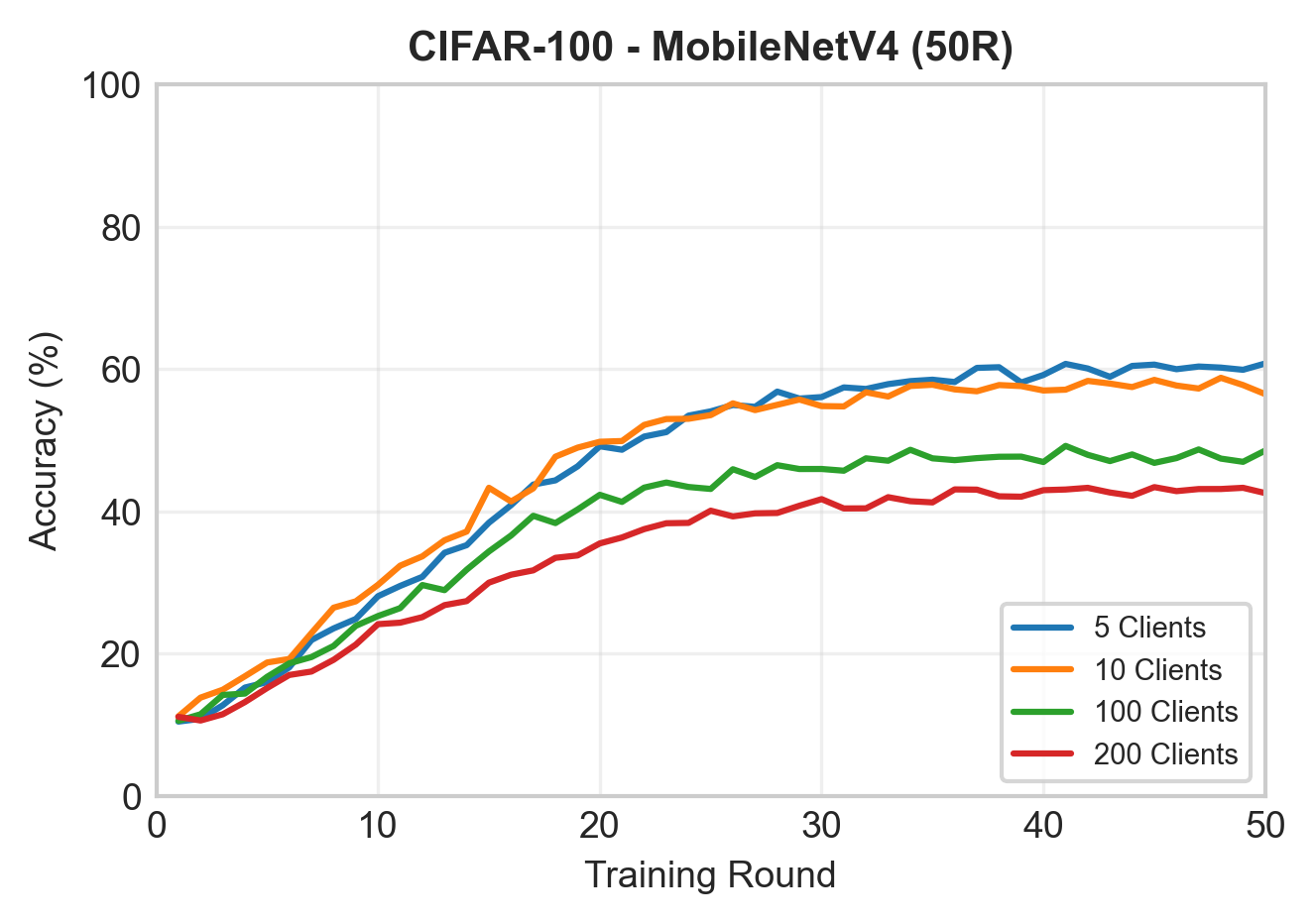}\end{subfigure}
\begin{subfigure}{0.23\textwidth}\includegraphics[width=\textwidth]{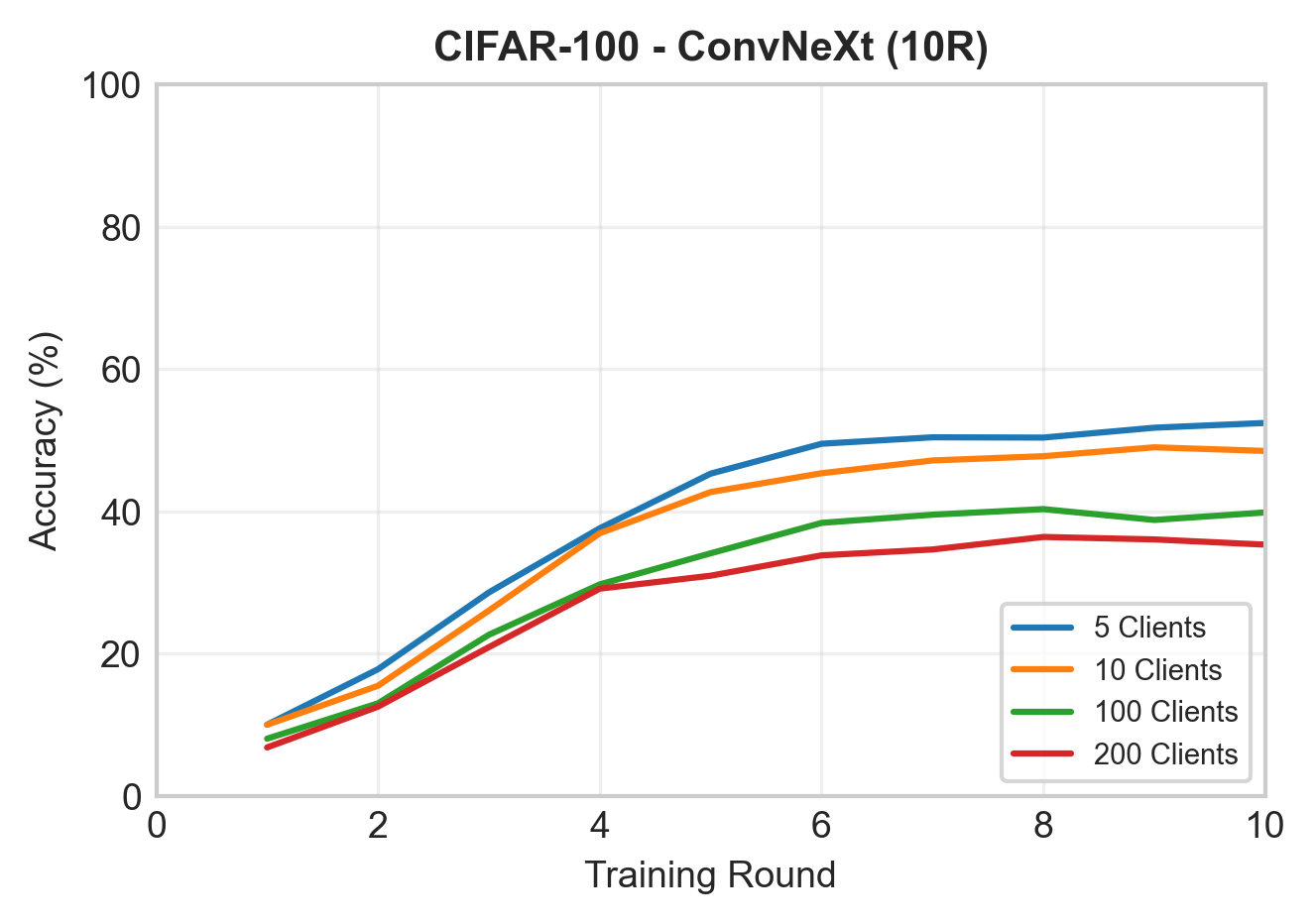}\end{subfigure}
\begin{subfigure}{0.23\textwidth}\includegraphics[width=\textwidth]{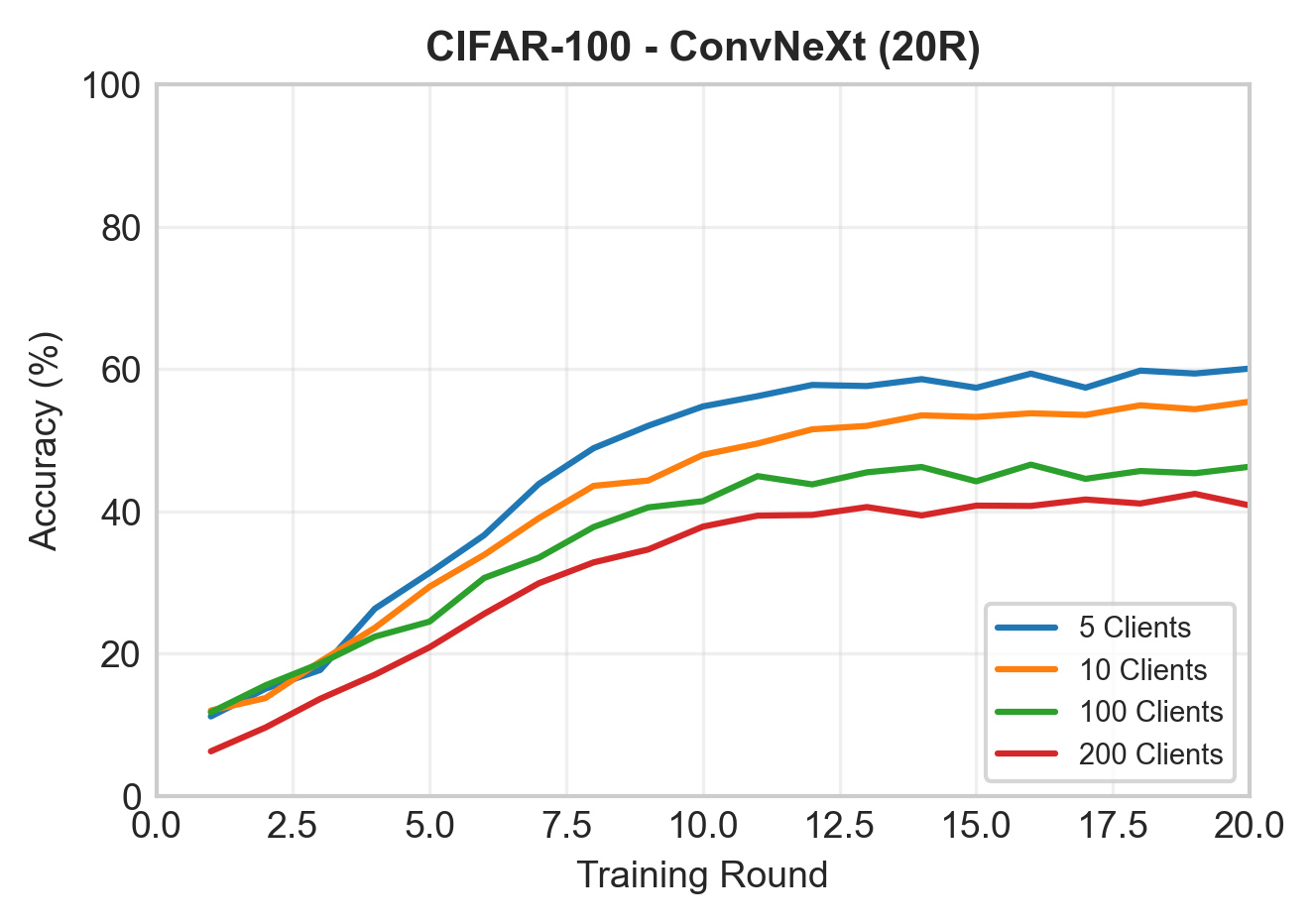}\end{subfigure}
\begin{subfigure}{0.23\textwidth}\includegraphics[width=\textwidth]{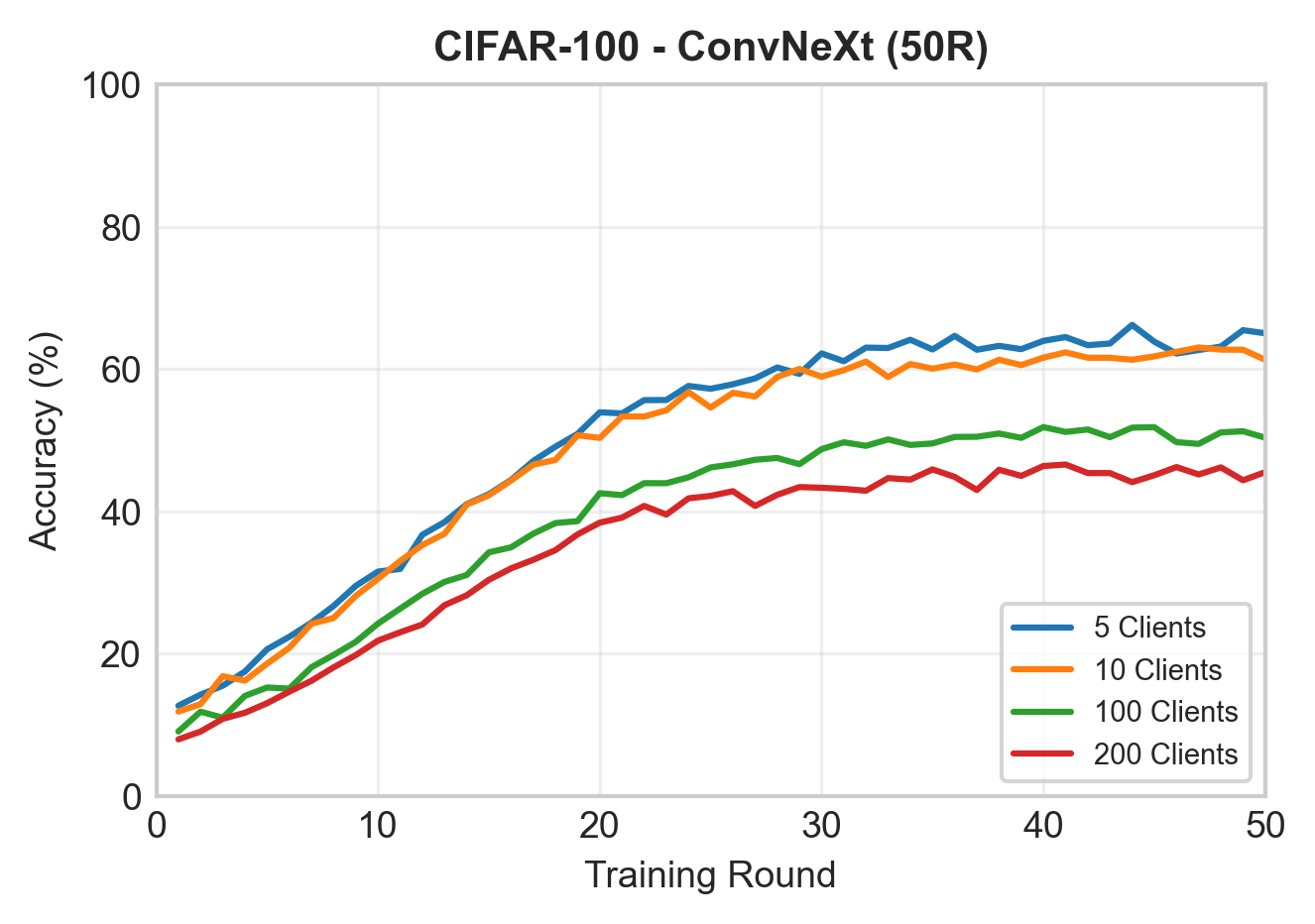}\end{subfigure}
\caption{\textbf{CIFAR-100 Accuracy Convergence (10, 20, 50 Rounds).} Performance analysis on the most challenging fine grained classification task with 100 classes. The 4$\times$3 grid (architectures $\times$ rounds) reveals architectural impact: shallow CNN struggles to capture fine grained features even at 50 rounds ($\sim$62\% accuracy), while ConvNeXt reaches $\sim$68\%. The progressive improvement from 10 to 50 rounds demonstrates that QSplitFL's split point adaptation successfully enables resource constrained edge devices to train deep models by offloading computational burden to the server. This capability is critical for complex real world tasks where shallow networks fundamentally lack sufficient capacity, which highlights QSplitFL's value proposition for heterogeneous federated learning environments.}
\label{fig:cifar100_acc_additional}
\end{minipage}
\end{figure*}

Figures~\ref{fig:mnist_acc_additional}--\ref{fig:cifar100_acc_additional} provide detailed convergence plots. All models rapidly converge on MNIST/Fashion-MNIST ($>$95\% in 10 rounds). On CIFAR-10/100, deep architectures (ConvNeXt, ResNet50) show clear better performance, with ConvNeXt outperforming CNN by 5-8\% on CIFAR-100, which shows the benefit of adaptive split selection.

\subsection{Ablation Study}
\label{appendix:ablation}
Table~\ref{tab:ablation} presents an ablation study evaluating the contribution of each design component in QSplitFL. All configurations are evaluated on CIFAR-10 with ResNet50 using 10 clients over 100 rounds. Each row isolates a single design choice while keeping all other components identical to the full model. The results show that the committee-based DQN ($M=3$), the decayed reward function, and the full capability-aware state representation each contribute meaningfully to convergence speed and final accuracy.

\begin{table}[h]
\centering
\caption{Ablation Study on QSplitFL Design Choices 
(CIFAR-10, ResNet50, 10 Clients, 100 Rounds)}
\label{tab:ablation}
\resizebox{\textwidth}{!}{%
\begin{tabular}{|l|c|c|c|}
\hline
\textbf{Configuration} & \textbf{Accuracy (\%)} & 
\textbf{Avg. Split Layer} & \textbf{Rounds to 80\% Acc.} \\
\hline
Full QSplitFL ($M$=3, decayed reward, capability state) & \textbf{83.73} & \textbf{37.4} & \textbf{28} \\
\hline
Single-head DQN ($M$=1, all else same)                  & 71.3           & 27.0 (fixed) & 53          \\
\hline
Committee $M$=5                                          & 83.9           & 37.6         & 31          \\
\hline
No decay ($\lambda$=0, flat reward)                      & 78.1           & 35.2         & 41          \\
\hline
High decay ($\lambda$=0.1)                               & 80.4           & 36.1         & 30          \\
\hline
Equal capability weights ($w_i$=0.25 $\forall i$)        & 82.5           & 37.0         & 29          \\
\hline
CPU-only state (no memory/battery/network)               & 76.8           & 33.5         & 38          \\
\hline
Random split (no RL, fixed mid-layer)                    & 69.2           & 25.0 (fixed) & 67          \\
\hline
\end{tabular}%
}
\end{table}

\noindent\textbf{Why does Committee $M$=5 require more rounds to reach 80\% accuracy than $M$=3?}

Here in this case, $M$=5 needs 3 votes to pick a split; but on the other hand, $M$=3 only needs 2. Early on, the five MLP heads are still noisy where the replay buffer barely has data, so Q-value estimates are scattered. For this reason more ties happen, and breaking them with mean Q-values adds variance to which split gets chosen, which also stalls the steady loss drops that drive early accuracy. The shared encoder takes longer to settle too, since five heads sending conflicting gradients are harder to reconcile than three. $M$=5 does edge ahead by round 100 (83.9\% vs. 83.73\%), but that early friction pushes its 80\% crossing from round 28 back to round 31.

\subsection{Extended Discussion}
\label{appendix:extended_discussion}
This subsection expands on the design questions raised in the main paper, covering the split depth lower bound and its privacy implications, the system-level cost of the framework, the choice of capability features, the position of QSplitFL relative to prior reinforcement learning work, and its limitations.

\subsubsection{Split Depth Lower Bound and Privacy.}
The lower bound $\ell_{\min} = \lceil L/2 \rceil$ is a configurable prior rather than a fixed rule. We set it to half the network depth by default for two reasons that were introduced in the action space definition: it forces clients to perform meaningful local feature extraction, and it prevents raw or near-raw inputs from leaving the device. We agree that a shallower split can be more compute-efficient when clients are powerful, so the bound is exposed as a deployment parameter: in a cluster of capable devices where on-device computation is cheap, an operator can lower $\ell_{\min}$ so that the agent is free to explore shallower splits and shift more work to the client, whereas a cluster of weak devices benefits from keeping $\ell_{\min}$ high. The default value encodes a conservative trade-off that favors privacy and bounded communication when client capability is unknown. Split depth is also a privacy control, which makes it a decisive factor in partitioning, because a deeper split keeps more layers on the device and transmits higher-level activations that are abstracted further from the input, and such representations are generally harder to invert than the shallow activations produced near the input layer~\cite{vepakomma2018split,li2021label}. The bound therefore acts as a minimum privacy floor: operators who require stronger guarantees can raise $\ell_{\min}$, add differentially private noise to the smashed data~\cite{wu2023split,abadi2016deep}, or, as a direction for future work, incorporate an explicit privacy term into the reward so that the agent jointly optimizes accuracy, communication, and leakage risk.

\subsubsection{System-Level Cost Considerations.}
Because the motivation of QSplitFL is capability-aware optimization in resource-constrained settings, we clarify how the framework relates to system-level cost. First, the controller itself is inexpensive: the capability-aware state requires only $\mathcal{O}(|\mathcal{K}|)$ aggregation, in contrast to weight-based methods that collect parameters and run PCA, and the committee of small MLPs is negligible next to the DNN being trained. Second, communication overhead is governed by the split layer, since the volume of smashed data per round scales with the activation size at layer $\ell$, and deeper splits transmit smaller tensors. The capability-aware state captures this through the network term $C_{\text{Net}}^{(k)}(t) = 1 - \text{Latency}^{(k)}(t)/\text{Latency}_{\max}$, so poorly connected clients receive lower scores and the agent favors communication-efficient deeper splits, as reported in the summary of findings. Third, client computation and energy are handled in the same implicit manner: the CPU, memory, and battery terms penalize weak or low-battery clients, so the agent offloads more layers to the server for them. We acknowledge that wall-clock latency and on-device energy are reflected here only through these proxies, and that direct hardware-level profiling of energy and end-to-end latency on a physical edge testbed would strengthen the validation. We leave such measurement, together with explicit communication, latency, and energy terms in the reward for multi-objective optimization, to future work.

\subsubsection{Capability Feature Selection.}
The state uses four capability metrics, namely CPU availability, memory availability, battery level, and network latency. These were selected because they are the dominant and directly observable bottlenecks for on-device DNN training: compute throughput, working-set memory, energy budget, and activation transfer time. They are also cheap to read from standard operating-system counters with negligible overhead, and each maps directly to the computation and communication trade-off that the split point controls. We deliberately avoid weight-derived features, which require costly collection and PCA projection. The ablation in Appendix~\ref{appendix:ablation} quantifies the effect of this choice: reducing the state to CPU only lowers accuracy from $83.73\%$ to $76.8\%$ and slows convergence from $28$ to $38$ rounds to reach $80\%$ accuracy, while using equal importance weights ($w_i = 0.25$) costs little ($82.5\%$). This indicates that the four metrics are complementary and that the policy is robust to the exact weighting but sensitive to dropping a metric entirely. The tunable weights $w_i$ let operators emphasize whichever resource is most constrained in their setting, such as battery for mobile clients or latency for rural network links.

\subsubsection{Positioning and Novelty.}
QSplitFL integrates established reinforcement learning components, including DQN with experience replay, a target network, reward shaping, and ensemble-style committee voting. Its contribution is not a new learning rule but a problem-specific reformulation for SFL. The first element is a lightweight, interpretable state that replaces high-dimensional weight representations and removes the PCA step, which to our knowledge yields the first DQN-based adaptive split-point selector for SFL. The second is a decayed loss-drop reward tailored to the early-convergence structure of SFL training. The third repurposes committee voting, which is related to Bootstrapped DQN~\cite{osband2016deep} and ensemble DQN~\cite{nixon2020bootstrapped}, specifically as a defense against reward hacking under this reward design, and we validate its effect empirically in the ablation. We therefore position the value of the work at the systems level, as a practical, interpretable, and low-overhead controller, rather than as a new reinforcement learning primitive.

\subsubsection{Limitations and Future Work.}
Our evaluation uses standard vision benchmarks and four convolutional architectures. Because the controller only needs a layer-index range, the framework is architecture-agnostic, and extending it to less conventional models such as vision transformers, sequence models, and graph networks, as well as to non-vision modalities such as medical imaging, time-series IoMT data, and language tasks, is a natural next step. Other directions follow from the discussion above: validation on a real edge testbed with measured energy and latency, explicit privacy and communication terms in the reward, and adaptive per-tier relaxation of the split depth bound $\ell_{\min}$.

\end{document}